\documentclass{article}
\PassOptionsToPackage{square,numbers}{natbib}
\usepackage[final]{neurips_2020}
\usepackage[utf8]{inputenc} 
\usepackage[T1]{fontenc}    
\usepackage{hyperref}       
\usepackage{url}            
\usepackage{booktabs}       
\usepackage{amsfonts}       
\usepackage{nicefrac}       
\usepackage{microtype}      
\usepackage{epsfig}
\usepackage{graphicx}
\usepackage{subfig}
\usepackage{multirow}
\usepackage{verbatim}
\usepackage{algorithm}
\usepackage{algorithmic}
\usepackage{appendix}

\newcommand{\figref}[1]{Fig.~\ref{#1}}
\newcommand{\tabref}[1]{Table~\ref{#1}}
\newcommand{\secref}[1]{Sec.~\ref{#1}}
\newcommand{\eqnref}[1]{Eqn.~\ref{#1}} 
\newcommand{\algref}[1]{Alg.~\ref{#1}}
\usepackage{color}

\author{%
    Alex Wong \\
    UCLA Vision Lab \\
    \texttt{alexw@cs.ucla.edu} \\
    \And
    Safa Cicek \\
    UCLA Vision Lab \\
    \texttt{safacicek@ucla.edu} \\
    \And
    Stefano Soatto \\
    UCLA Vision Lab \\
    \texttt{soatto@cs.ucla.edu} \\
}

\title{Targeted Adversarial Perturbations \\ for Monocular Depth Prediction}
\begin{document}
\maketitle
\begin{abstract}
We study the effect of adversarial perturbations on the task of monocular depth prediction. Specifically, we explore the ability of small, imperceptible additive perturbations to selectively alter the perceived geometry of the scene. We show that such perturbations can not only globally re-scale the predicted distances from the camera, but also alter the prediction to match a different target scene. We also show that, when given semantic or instance information, perturbations can fool the network to alter the depth of specific categories or instances in the scene, and even remove them while preserving the rest of the scene. To understand the effect of targeted perturbations, we conduct experiments on state-of-the-art monocular depth prediction methods. Our experiments reveal vulnerabilities in monocular depth prediction networks, and shed light on the biases and context learned by them.  
\end{abstract}

\begin{figure}[h]
    \centering
    \includegraphics[width=1.00\textwidth]{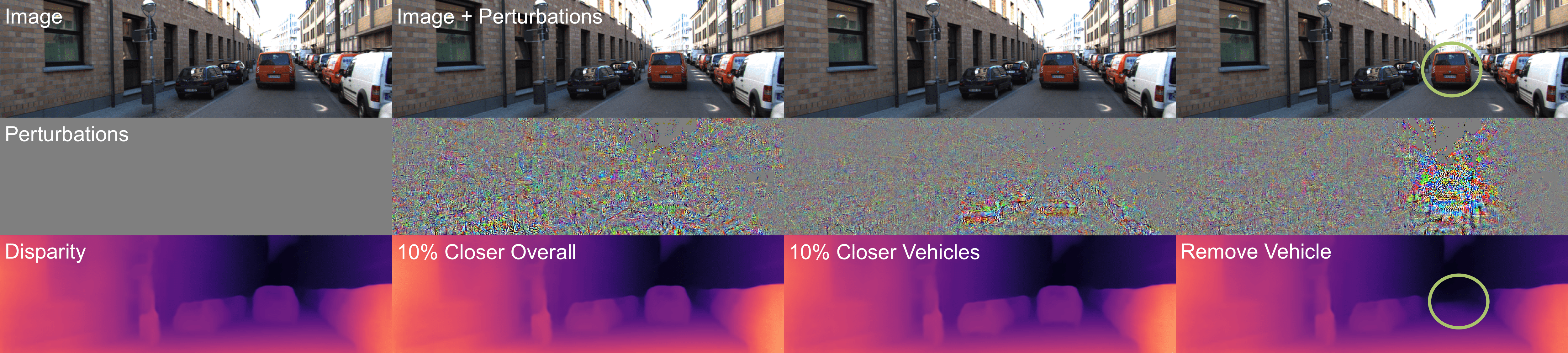}
    \caption{\textbf{Altering the predicted scene with adversarial perturbations.} Top to bottom: input image; adversarial perturbations with upper norm of $2 \times 10^{-2}$; predicted scene visualized as disparity. Left to right: original image and predicted scene; overall scene altered to be 10\% closer; all vehicles altered to be 10\% closer; vehicle in the center of the road is removed by perturbations.}
    \vspace{-1.2em}
    \label{fig:teaser}
\end{figure}

\vspace{-0.3em}
\section{Introduction}
\label{sec:introduction}
\vspace{-0.5em}
Consider the image shown in the top-left of \figref{fig:teaser}, captured from a moving car.  The corresponding depth of the scene, inferred by a deep neural network and visualized as disparity, is shown underneath. Can adding a small perturbation cause the perceived vehicle in front of us disappear? Indeed, this is shown on the rightmost panel of the same figure: The perturbed image, shown on the top-right, is indistinguishable from the original. Yet, the perturbation, amplified and shown in the center row, causes the depth map to be altered in a way that makes the car in front of us disappear.

Adversarial perturbations are small signals that, when added to images, are imperceptible yet can cause the output of a deep neural network to change catastrophically \cite{szegedy2013intriguing}. We know that they can fool a network to mistake a tree for a peacock \cite{moosavi2016deepfool}. But, as autonomous vehicles are increasingly employing learned perception modules, mistaking a stop sign for a speed limit \cite{eykholt2018robust} or causing obstacles to disappear is not just an interesting academic exercise. We explore the possibility that small perturbations can alter not just the class label associated to an image, but the inferred depth map, for instance to make the entire scene appear closer or farther, or portions of the scene, like specific objects, to become invisible or be perceived as being elsewhere in the scene.

When semantic segmentation is available, perturbations can target a specific category in the predicted scene. Some categories (e.g. traffic lights, humans) are harder to attack than others (e.g. roads, nature). When instance segmentation is available, perturbations can manipulate individual objects, for instance make a car disappear or move it to another location. We call these phenomena collectively as \textit{stereopagnosia}, as the solid geometric analogue of prosopagnosia \cite{damasio1982prosopagnosia}.   

Stereopagnosia sheds light on the role of context in the representation of geometry with deep networks. When attacking a specific category or instance, while most of the perturbations are localized, some are distributed throughout the scene, far from the object of interest. Even when the target effect is localized (e.g., make a car disappear), the perturbations are non-local, indicating that the network exploits non-local context, which represents a vulnerability. Could one perturb regions in the image, for instance displaying billboards, thus making cars seemingly disappear?

We note that, although the adversarial perturbations we consider are not universal, that is, they are tailored to a specific scene and its corresponding image, they are somewhat robust. Blurring the image after applying the perturbations reduces, but does not eliminate, stereopagnosia. To understand generalizability of adversarial perturbations, we examine the transferability of the perturbations between monocular depth prediction models with different architectures and losses.
\vspace{-1em}
\section{Related Work}
\vspace{-0.8em}
Adversarial perturbations have been studied extensively for classification (\secref{sec:adversarial_perturbations}). We focus on regression, where there exists some initial work. However, we study the \textit{targeted} case where the entire scene, a particular object class, or even an instance is manipulated by the choice of perturbation.


\vspace{-0.7em}
\subsection{Adversarial Perturbations}
\vspace{-0.5em}
\label{sec:adversarial_perturbations}
The early works \cite{goodfellow2014explaining,szegedy2013intriguing} showed the existence of small, imperceptible additive noises that can alter the predictions of deep learning based classification networks. Since then many more advanced attacks \cite{moosavi2016deepfool} have been proposed. \cite{moosavi2017universal} showed the existence of universal perturbations i.e. constant additive perturbations to degrade the accuracy over the entire dataset.

More recently, \cite{naseer2019cross} studied transferability of attacks across datasets and models. \cite{peck2017lower} derived lower bounds on the magnitudes of perturbations. \cite{najafi2019robustness} studied the attacks in semi-supervised learning setting. \cite{qin2019adversarial,tramer2019adversarial} proposed methods to enhance robustness to adversarial attacks. \cite{laidlaw2019functional} extended adversarial attacks beyond small additive perturbations. \cite{ilyas2019adversarial} showed that the existence of adversarial attacks makes deep networks more predictive.

Despite the exponentially growing literature on adversarial attacks for the classification task, there only have been a few works extending analysis of adversarial perturbations to dense-pixel prediction tasks. \cite{xie2017adversarial} studied adversarial perturbations for detection and segmentation. \cite{hendrik2017universal} demonstrated targeted universal attacks for semantic segmentation. \cite{wong2020stereopagnosia} studied non-targeted perturbations for stereo, and \cite{ranjan2019attacking} used patch attacks for optical flow. \cite{mopuri2018generalizable} examined universal perturbations in a data-free setting for segmentation and depth prediction to alter predictions in arbitrary directions. Unlike them, we study \textit{targeted} attacks where network is fooled to predict a specific target.

Our goal is to analyze the robustness of the monocular depth prediction networks to different targeted attacks to explore possible explanations of what is learned by these models. With a similar motivation, \cite{dijk2019neural} inserted various objects into the images and \cite{hu2019visualization} identified a small set of pixels from which a network can predict depth with small error. Unlike them, we analyze the monocular depth networks by studying their robustness against \textit{targeted adversarial} attacks.

\vspace{-0.7em}
\subsection{Monocular Depth Prediction}
\vspace{-0.5em}
\cite{eigen2015predicting,eigen2014depth, laina2016deeper, liu2015deep, liu2016learning, yin2019enforcing} trained deep networks with ground-truth annotations to predict depth from a single image. However, high quality depth maps are often unavailable and, when available, are expensive to acquire. Hence, trends shifted to weaker supervision from crowd-sourced data \cite{chen2016single}, and ordinal relationships amongst depth measurements \cite{fu2018deep,zoran2015learning}. 

Recently, supervisory trends shifted to unsupervised (self-supervised) learning, which relies on stereo-pairs or video sequences during training, and provides supervision in the form of image reconstruction. While depth from video-based methods is up to an \textit{unknown} scale, stereo-based methods can predict depth in \textit{metric} scale because the pose (baseline) between the cameras is known. 

To learn depth from stereo-pairs, \cite{garg2016unsupervised} predicted disparity by reconstructing one image from its stereo-counterpart. Monodepth \cite{godard2017unsupervised} predicted both left and right disparities from a single image and laid the foundation for \cite{pillai2019superdepth,poggi2018learning, wong2019bilateral}. To learn depth from videos, \cite{mahjourian2018unsupervised, zhou2017unsupervised} also jointly learned pose between temporally adjacent frames to enable image reconstruction by reprojection. \cite{wang2018learning, yang2018deep} leveraged visual odometry, \cite{fei2018geo} used gravity, \cite{casser2019depth, luo2018every} considered motion segmentation, and \cite{yin2018geonet} jointly learned depth, pose and optical flow. Monodepth2 \cite{godard2019digging} explored both stereo and video-based methods and proposed a reprojection loss to discard potential occlusions. PackNet \cite{guizilini20203d} used 3D convolutions.

To study the effect of adversarial perturbations, we examine the robustness of unsupervised methods Monodepth \cite{godard2017unsupervised}, and the state-of-the-art Monodepth2 \cite{godard2019digging}, PackNet \cite{guizilini20203d} for outdoor scenes, and supervised method VNL \cite{yin2019enforcing} for indoors. While Monodepth2 proposed both stereo and video-based models, we choose their stereo model because the predicted depth is in \textit{metric} scale, which enables us to study perturbations to alter the scale of the scene without changing its topology.

In \secref{sec:finding_targeted_adversarial_perturbations}, we discuss our method. We show perturbations for altering entire predictions to a target scene in \secref{sec:attack_the_entire_scene} and localized attacks on specific categories and object instances in \secref{sec:attacking_a_specific_portion}. We discuss the transferabilty of such perturbations in \secref{sec:transferability} and their robustness against defenses in Supp. Mat.

\vspace{-1em}
\section{Finding Targeted Adversarial Perturbations}
\label{sec:finding_targeted_adversarial_perturbations}
\vspace{-0.8em}
Given a pretrained depth prediction network, $f_d:\mathbb R^{H \times W \times 3} \rightarrow \mathbb R_{+}^{H \times W}$, $f_d: x \mapsto d(x)$ our goal is to find a small additive perturbation $v(x) \in \mathbb R^{H \times W \times 3}$, as a function of the input image $x$, which can change its prediction to a target depth $f_d(x+v(x))= d^t(x) \neq d(x)$ with some norm constraint $||v(x)||_{\infty} \leq \xi$ and high probability $\mathbb P (f_d(x+v(x)) = d^t(x)) \geq 1-\delta$.

We begin by examining Dense Adversarial Generation (DAG) proposed by \cite{xie2017adversarial} for finding adversarial perturbations for the semantic segmentation task. The perturbations from DAG can be formulated as the sum of a gradient ascent term (that pushes the predictions away from those of the original image) and a gradient descent term (that pulls predictions towards the target predictions). 
In the case of semantic segmentation, this formulation works well because the gradient ascent term suppresses the probability for the original predictions, which naturally increases the probability of the target predictions (zero-sum) driven by the gradient descent term. However, such is not the case for regression tasks, which requires the network to predict a real-valued scalar (as opposed to probability mass) for a targeted scene. Hence, the gradient ascent term maximizes the difference between the original and predicted depth, which results in DAG ``overshooting'' the target depth. 

Instead, we use a simple objective function, similar to \cite{hendrik2017universal}, but we modify it for the regression task by minimizing the normalized difference between predicted and target depth, 
\begin{equation}
    \ell(x, v(x), d^t(x), f_d) := \frac{||f_d(x+v(x)) - d^t(x)||_1}{d^t(x)}.   
\label{eqn:target_loss}
\end{equation}

We minimize this objective with respect to an image $x$ by following an iterative optimization procedure (\algref{alg:pseudo_code}). The $\texttt{CLIP}\big(v_{n}(x), -\xi, \xi \big)$ operation clamps any value of $v(x)$ larger than $\xi$ to $\xi$ and any value smaller than $-\xi$ to $-\xi$. For all the experiments, $\xi \in \{2 \times 10^{-3}, 5 \times 10^{-3}, 1 \times 10^{-2}, 2 \times 10^{-2} \}$.

\begin{algorithm}[t]
\caption{Proposed method to calculate targeted adversarial perturbations for a regression task.}
\begin{algorithmic} 
    \STATE \textbf{Parameters:} Learning rate $\eta$, noise upper norm $\xi$.
    \STATE \textbf{Inputs:} Image $x$, target depth map $d^t(x)$, pretrained depth network $f_d$.
    \STATE \textbf{Outputs:} Perturbation $v_N(x)$.
    \STATE \textbf{Init:} $v_{0}(x) = 0$.
    \FOR{$n = 0 : N-1$}
        \STATE $v_{n}(x) = \texttt{CLIP}\big(v_{n}(x), -\xi, \xi \big)$
        \STATE Calculate $\ell(x, v_{n}(x), d^t(x), f_d)$ as defined in \eqnref{eqn:target_loss}.
        \STATE $v_{n+1}(x) = v_{n}(x) - \eta \nabla \ell(x, v_{n}(x), d^t(x), f_d)$
    \ENDFOR
    \STATE $v_{N}(x) = \texttt{CLIP}\big(v_{N}(x), -\xi, \xi \big)$
\end{algorithmic} 
\label{alg:pseudo_code}
\end{algorithm}

\vspace{-0.7em}
\subsection{Implementation Details}
\label{sec:experiments_details}
\vspace{-0.5em}
We evaluate adversarial targeted attacks on KITTI semantic split \cite{Alhaija2018IJCV}. This is a dataset of $200$ outdoor scenes, captured by car-mounted stereo cameras and a LIDAR sensor, with ground-truth semantic segmentation and instance labels. The semantic and instance labels in this split enables our experiments in \secref{sec:attacking_a_specific_portion} for targeting specific categories or instances in a scene.

The depth models (Monodepth, Monodepth2) are trained on the KITTI dataset \cite{geiger2012we} using Eigen split \cite{eigen2015predicting}. The Eigen split contains 32 out of the total 61 scenes, and is comprised of 23,488 stereo pairs with an average size of $1242 \times 375$. Images are resized to $640 \times 192$ as a preprocessing step and perturbations are computed with 500 steps of SGD. Entire optimization for each frame takes $\approx 12$s (Monodepth2 takes $22$ms for each forward pass and $11\textrm{s} \approx 500 \times 22$ms in total) using a GeForce GTX 1080. Details on hyper-parameters are provided in the Supp. Mat.

For all the experiments, we use absolute relative error (ARE), computed with respect to the target depth $d^t(x)$, as our evaluation metric:
\vspace{-0.5em}
\begin{equation}
    \textrm{ARE} = \frac{||f_d(x+v(x)) - d^t(x)||_1}{d^t(x)}.
\label{eqn:are}
\end{equation}

\vspace{-1.5em}
\subsection{Summary of Findings}
\label{sec:summary}
\vspace{-0.5em}
By studying targeted adversarial perturbations and their effect on monocular depth prediction networks, we found (i) monocular depth prediction networks heavily rely on edges to determine scale of the scene (\secref{sec:scaling_the_scene}); (ii) with large perturbations (but still imperceptible), an adversary can fool the network to predict scenes completely different from the ones observed (\secref{sec:symmetrically_flipping_the_scene}, \ref{sec:preset_scene}); (iii) depth prediction networks exhibit heavy biases present in the dataset (\secref{sec:symmetrically_flipping_the_scene}, \ref{sec:preset_scene}); (iv) regions belonging to different semantic categories demonstrate different levels of robustness against targeted attacks (\secref{sec:category_conditional_scaling}); (v) depth prediction networks leverage global context for local predictions, making them susceptible to non-local attacks (\secref{sec:instance_conditional_removing}, \ref{sec:instance_conditional_removing_with_spatial_constraints}); (vi) targeted perturbations do not transfer between networks, but, when crafted for multiple networks, can fool all of them equally well (\secref{sec:transferability}).

\begin{figure}[]
    \centering
    \subfloat[][]{
        \includegraphics[width=0.24\textwidth]{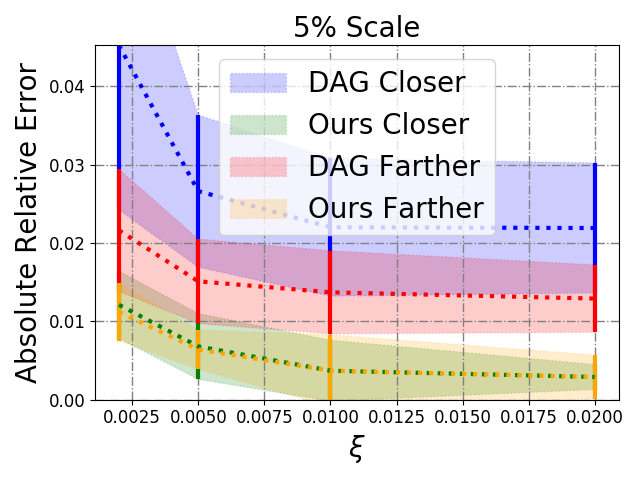}
    }
    \subfloat[][]{
        \includegraphics[width=0.24\textwidth]{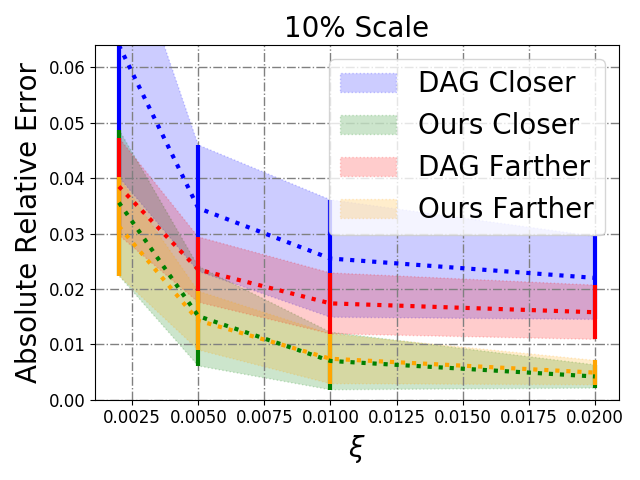}
    }
    \subfloat[][]{
        \includegraphics[width=0.24\textwidth]{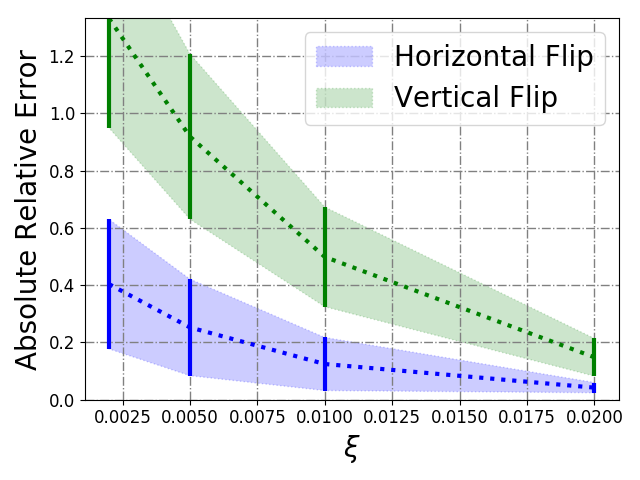}
    }
    \subfloat[][]{
        \includegraphics[width=0.24\textwidth]{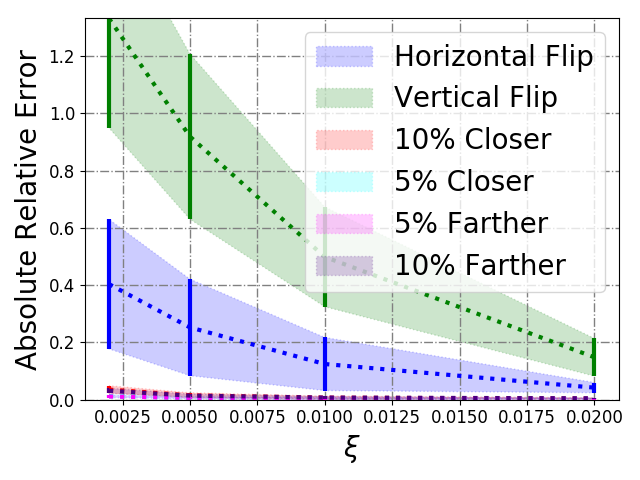}
    }
    \vspace{-0.5em}
    \caption{\textbf{ARE with various upper norm $\xi$ for scaling and flipping Monodepth2 predictions.} (a) and (b) Comparisons between DAG and the proposed method for scaling the scene by $\pm5\%$ and $\pm10\%$. (c) Results for horizontally and vertically flipping the predictions. (d) comparison between scaling and flipping tasks. Vertically flipping proves to be the most challenging.}
    \label{fig:plots_flip_scaleall_dag}
    \vspace{-1.5em}
\end{figure}

\begin{figure}[]
    \centering
    \subfloat[][]{
        \includegraphics[width=0.24\textwidth]{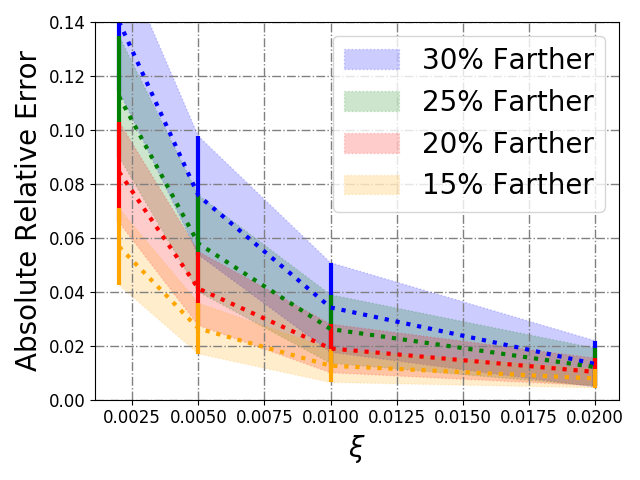}
    }
    \subfloat[][]{
        \includegraphics[width=0.24\textwidth]{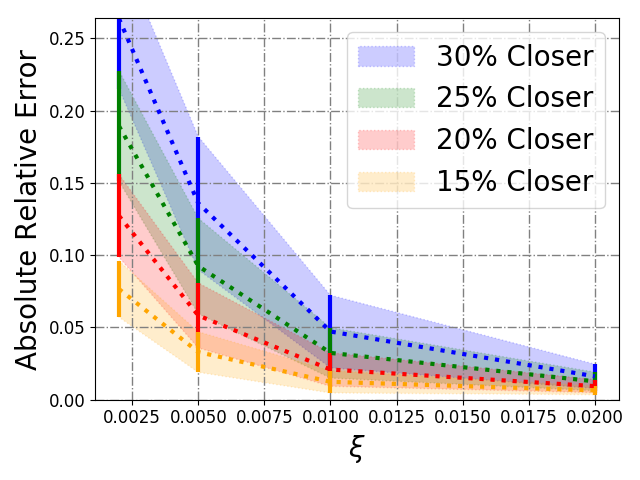}
    }
    \subfloat[][]{
        \includegraphics[width=0.24\textwidth]{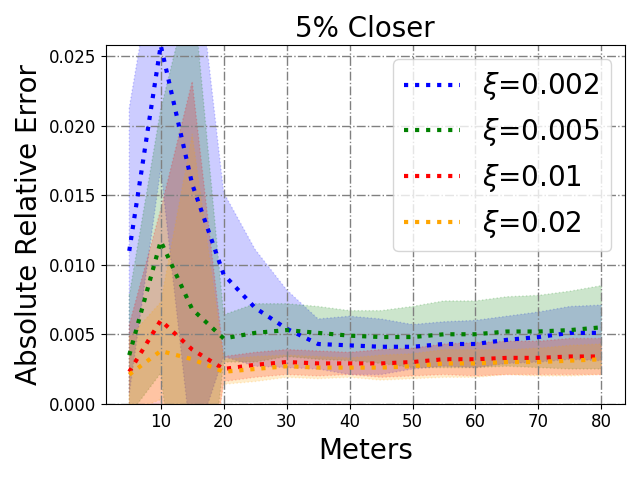}
    }
    \subfloat[][]{
        \includegraphics[width=0.24\textwidth]{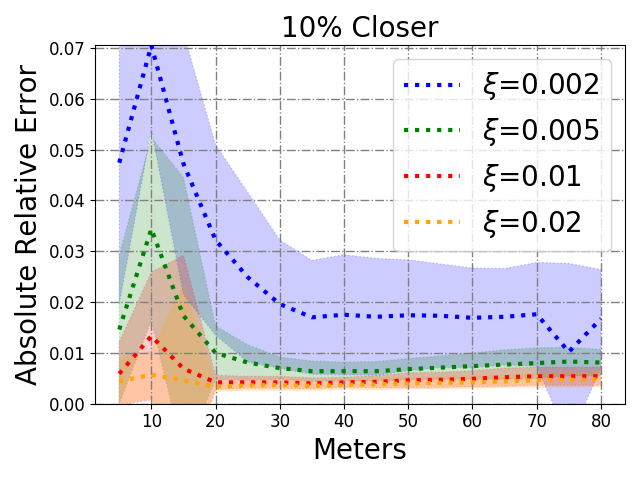}
    }
    \vspace{-0.5em}
    \caption{\textbf{Assessing larger scaling factors and error at various distances.} (a, b) ARE with various upper norm $\xi$ for scaling up to 30\% closer and farther. With $\xi=2 \times 10^{-2}$, an adversary to scale the scene up to 30\% with little error. (c, d) ARE binned every 5m for scaling 5\% and 10\% closer. Most errors are concentrated around 10m. Note: scaling 5\% and 10\% farther also have similar plots.}
    \label{fig:plots_larger_scaling_binned_error}
\end{figure}

\vspace{-1em}
\section{Attacking the Entire Scene}
\label{sec:attack_the_entire_scene}
\vspace{-0.8em}
Given a depth network $f_d$, our goal is to find adversarial perturbations to alter the predictions to a target scene $d^{t}(x)$ for an image $x$. For this, we examine three settings (i) scaling the entire scene by a factor, (ii) symmetrically flipping the scene, and (iii) altering the scene to a preset scene.

\vspace{-0.7em}
\subsection{Scaling the Scene}
\label{sec:scaling_the_scene}
\vspace{-0.5em}
For autonomous navigation, misjudging an obstacle to be farther away than it is could prove disastrous. Hence, to examine the possibility of altering the distances in the predicted scene without changing the scene topology or structure, we study perturbations that will scale the scene (bringing the scene closer to or farther away from the camera) by a factor of $1 + \alpha$. The target scene is defined as:
\begin{equation}
    d^t(x) = \texttt{scale}(f_d(x), \alpha) = (1 + \alpha) f_d(x)
\end{equation}
for $\alpha \in \{-0.10, -0.05, +0.05, +0.10\}$ or $-10\%$, $-5\%$ (closer), $+5\%$, $+10\%$ (farther), respectively. Column two of \figref{fig:teaser} shows the scene scaled 10\% closer to the camera by applying visually imperceptible perturbations with $\xi = 2 \times 10^{-2}$. On average, scaling the scene by $-10\%$, $-5\%$, $+5\%$, $+10\%$ with $\xi = 2 \times 10^{-2}$ require an $\|v(x)\|_1$ of 0.0160, 0.0124, 0.0126, and 0.0161, respectively. We note that scaling the scene by $\pm 5\%$ requires less perturbations than $\pm 10\%$ and the magnitude required for both directions is approximately symmetric. Also, perturbations are typically located along the object boundaries with concentrations on the road. For a side by side visualization of comparisons between different scaling factors, please see the Supp. Mat. 

In \figref{fig:plots_flip_scaleall_dag}-(a, b), we compare our approach with DAG (\secref{sec:finding_targeted_adversarial_perturbations}), re-purposed for depth prediction task. While both are bounded by the same upper norm, DAG consistently produces results with higher error and generally with a higher standard deviation. As seen in \figref{fig:plots_flip_scaleall_dag}-(a, b), even with $\xi = 2 \times 10^{-3}$, we are able to find perturbations that can scale the scene to be $\approx 1\%$ from $\pm 5\%$ and $\approx 3\%$ from $\pm 10\%$. With $\xi = 2 \times 10^{-2}$, we are able to fully reach all four targets with less than $\approx 0.5\%$ error.

We also consider larger scaling factors (up to 30\% closer and farther) in \figref{fig:plots_larger_scaling_binned_error}-(a, b). While an adversary using the smallest upper norm, $\xi = 2 \times 10^{-3}$, is unable to alter the scene to the desired scale, one using the largest, $\xi = 2 \times 10^{-2}$, can still achieve target with little error. For a more detailed discussion on larger scaling factors and limitations, please refer to the Supp. Mat. To examine where the errors generally occur, we binned ARE every 5 meters (m) in \figref{fig:plots_larger_scaling_binned_error}-(c,d) and found that the error is generally concentrated around 10 meters.

\begin{figure}[t]
\centering
    \subfloat[][]{
        \includegraphics[width=0.46\textwidth]{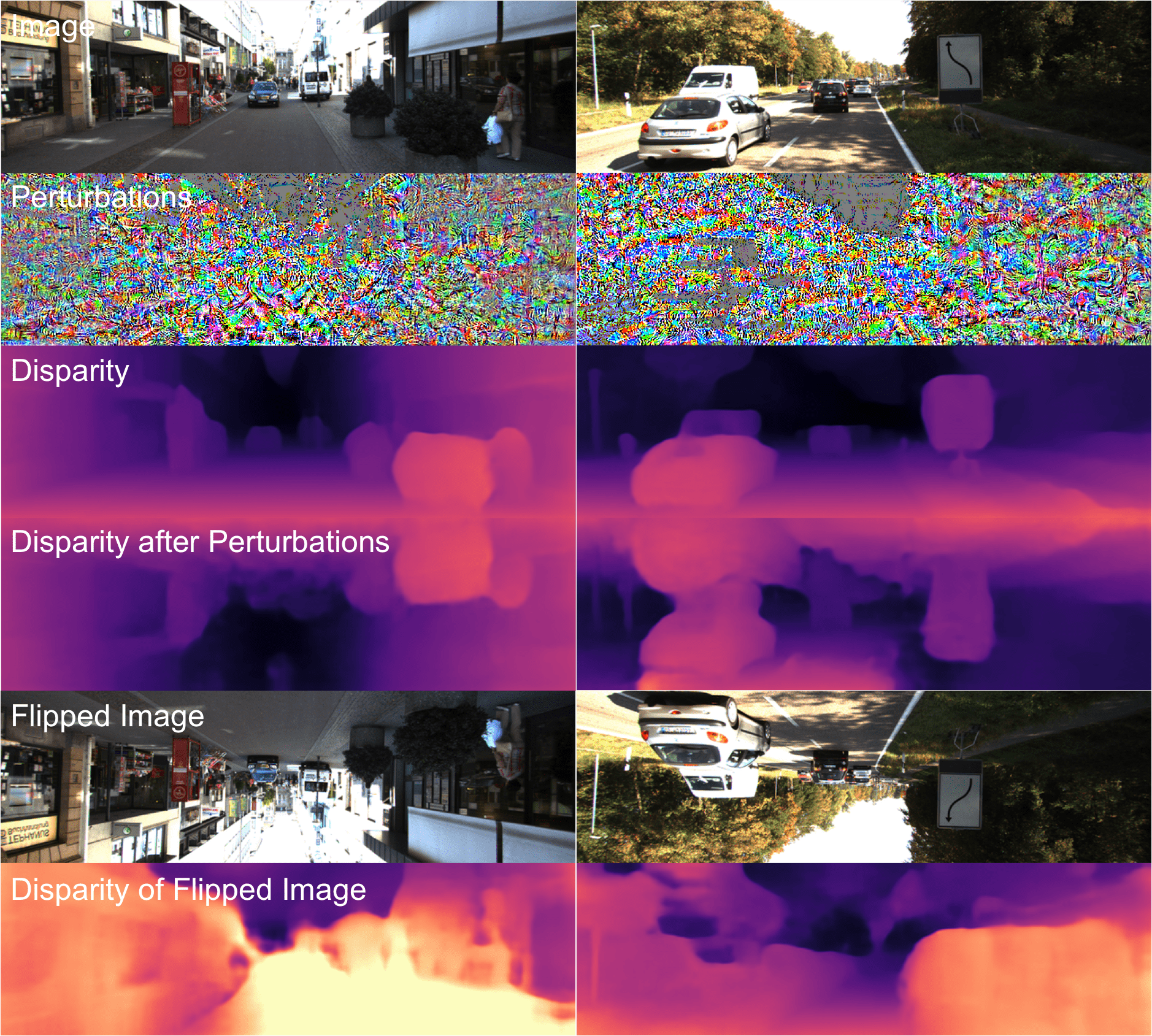}
    }
    \subfloat[][]{
        \includegraphics[width=0.46\textwidth]{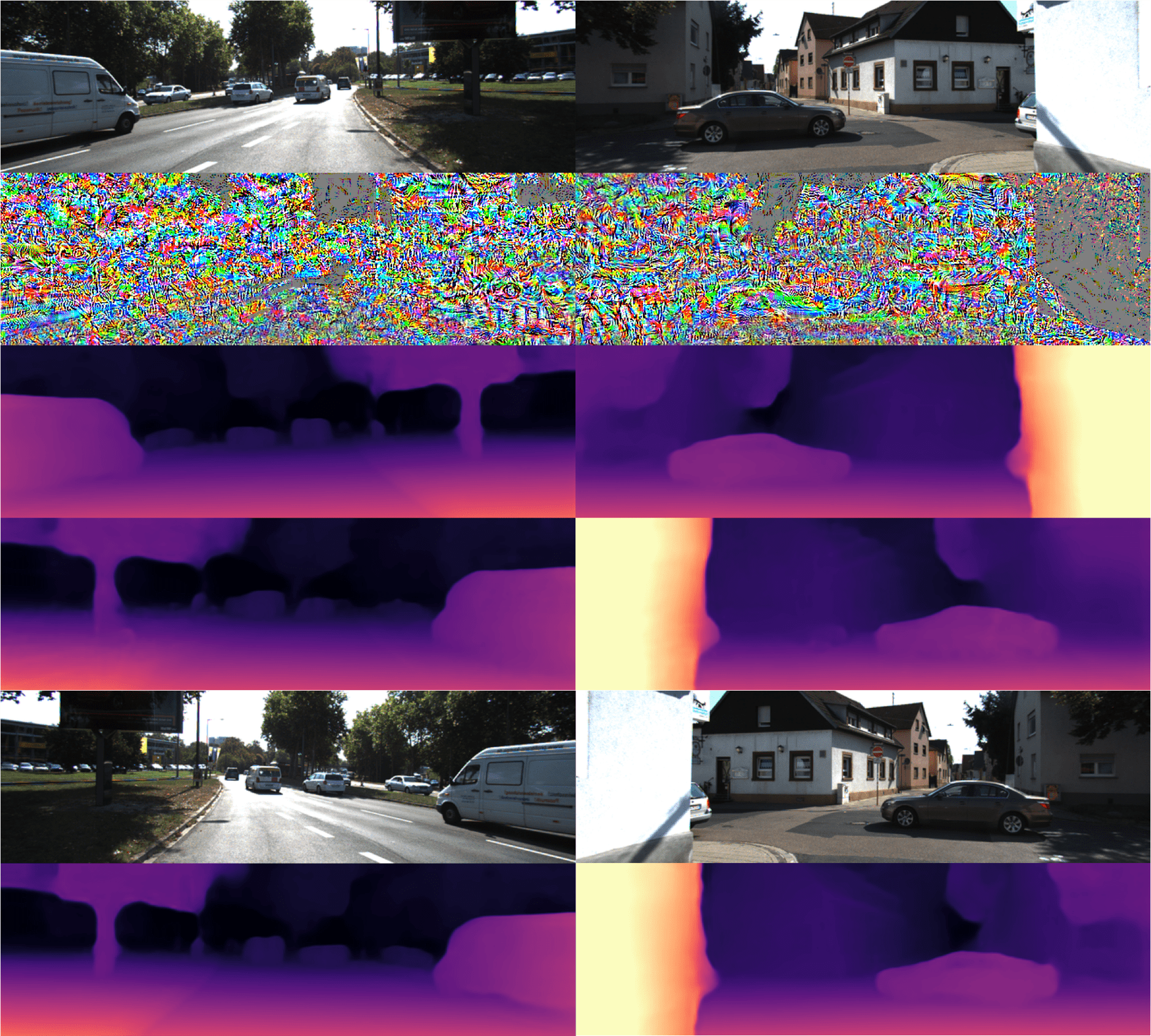}
    }
    \vspace{-0.6em}
    \caption{(a) \textbf{Examples of success (left) and failure (right) cases for vertical flip}. For the failure case, the car and road still remain on the bottom of the predictions. This is likely because the network is biased to predict closer structures on the bottom half of the image and farther ones on the top half (last two rows). (b) \textbf{Examples of horizontal flip.} Here, we observe the noise required to create and remove surfaces. Surprisingly, removing the white wall (right) requires very little perturbations.}
\label{fig:horizontal_vertical_flip}
\vspace{-1.8em}
\end{figure}

\begin{figure}[]
    \centering
    \subfloat[][]{
        \includegraphics[width=0.24\textwidth]{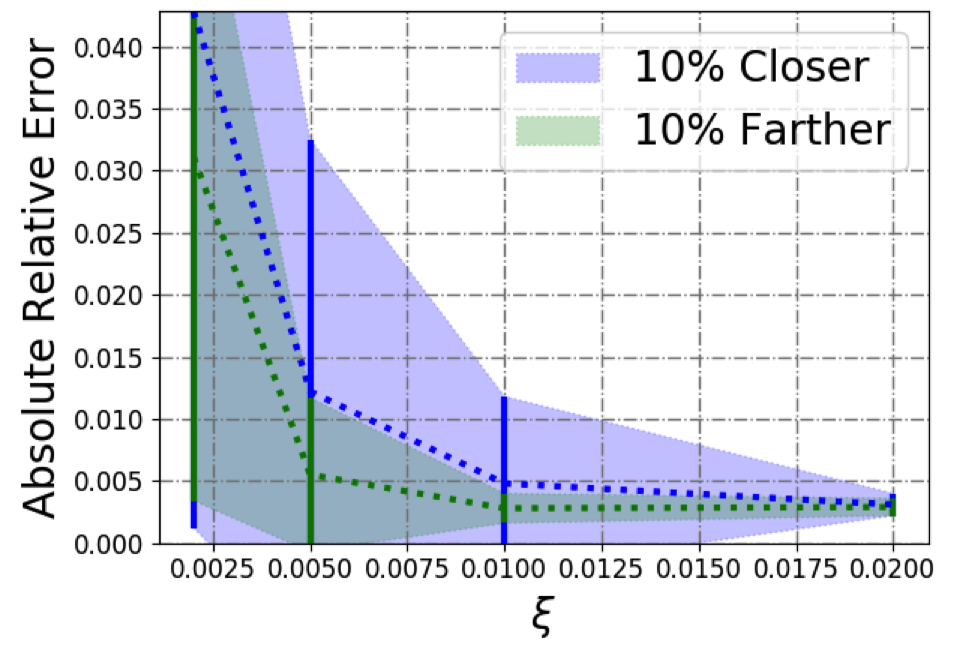}
    }
    \subfloat[][]{
        \includegraphics[width=0.24\textwidth]{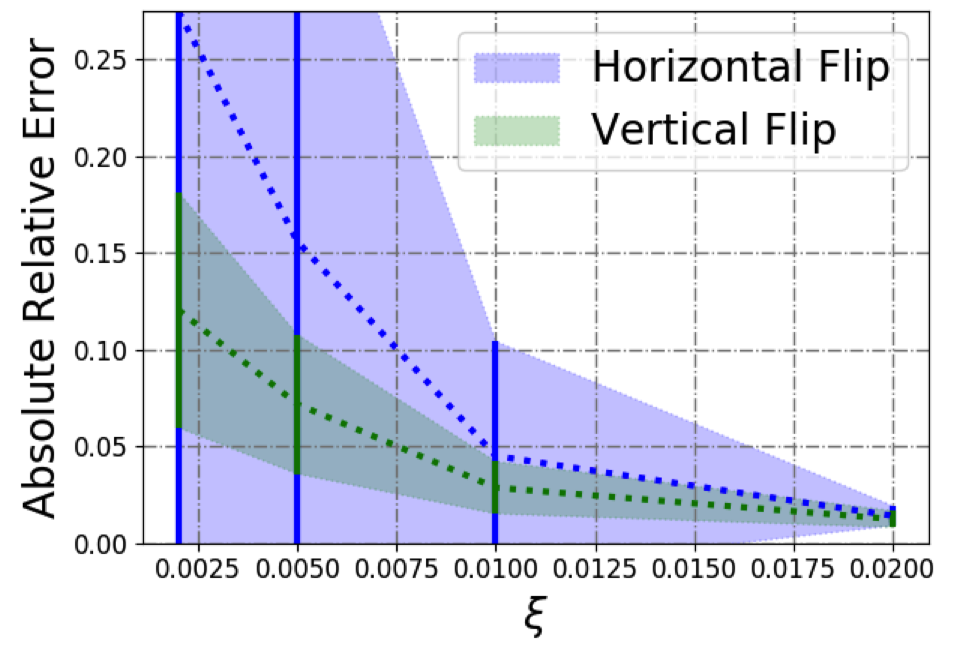}
    }
    \subfloat[][]{
        \includegraphics[width=0.24\textwidth]{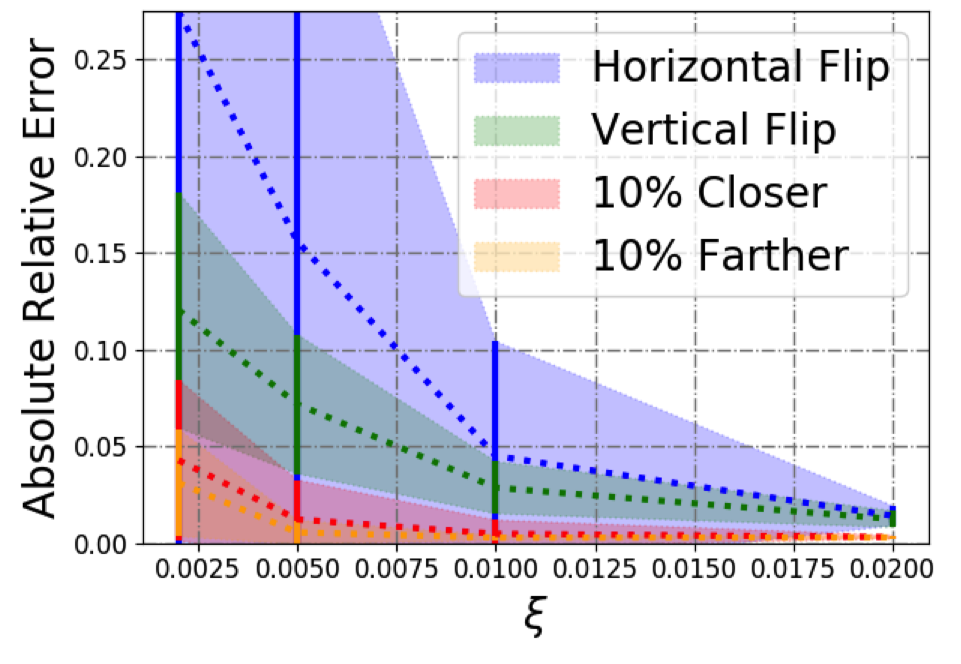}
    }
    \vspace{-0.5em}
    \caption{\textbf{ARE with various upper norm $\xi$ for scaling and flipping VNL predictions on NYUv2}. (a) Results for horizontally and vertically flipping the predictions, (b) Results for scaling the scene by $\pm10\%$, and (c) comparison between scaling and flipping tasks. Unlike Monodepth2 trained on KITTI, VNL trained on NYUv2 is less biased by the orientation of the scene.}
    \label{fig:plots_scale_flip_all_indoors}
    \vspace{-0.2em}
\end{figure}

\vspace{-0.8em}
\subsection{Symmetrically Flipping the Scene}
\vspace{-0.5em}
\label{sec:symmetrically_flipping_the_scene}
We now examine the problem setting where the target scene still retains the same structures given by the image, however, they are mirrored across the y-axis (horizontal flip) or x-axis (vertical flip). 

For the \textit{horizontal flip} scenario, we denote the target depth as
$d^t(x) = \texttt{fliph}(f_d(x))$ 
where the $\texttt{fliph}$ operator horizontally flips the predicted depth map $f_d(x)$ across the y-axis.

\figref{fig:horizontal_vertical_flip}-(b) shows that the perturbations can fool the network into predicting horizontal flipped scenes. For scenes with different structures on either side, $v(x)$ fools the network into \textit{creating and removing} surfaces. Note that the amount of noise required to horizontally flip the scene is much more than that to scale the scene (i.e. for $\xi = 2 \times 10^{-2}$, $\|v(x)\|_1 = 0.0161$ for scaling $+10\%$, and  $\|v(x)\|_1 = 0.0326$ for horizontal flip), which illustrates the difficulty in altering the scene structures. Interestingly, the amount of noise required to remove the white wall (\figref{fig:horizontal_vertical_flip}-(b)) is significantly less than the rest.

For the \textit{vertical flip} scenario, we denote the target depth as 
$d^t(x) = \texttt{flipv}(f_d(x))$
where $\texttt{flipv}$ operator vertically flips the predicted depth map $f_d(x)$ across the x-axis. 

As seen in \figref{fig:horizontal_vertical_flip}-(a), perturbations cannot fully flip the predictions vertically. Even on successful attempts (left), there are still artifacts in the output. For failure cases (right), portions of the cars still remain on the bottom half of the predictions. This experiment reveals the potential $\textit{biases}$ learned by the network. To verify this, we feed vertically flipped images to the network. As seen in the last two rows of \figref{fig:horizontal_vertical_flip}-(a), the network still assigns closer depth values to the bottom half of the image (now sky) and farther depth values to the top half (now road and cars). Unlike outdoor datasets, indoor ones do not exhibit this orientation bias. \figref{fig:plots_scale_flip_all_indoors} shows that it is in fact easier to fool VNL \cite{yin2019enforcing}, trained on NYUv2 \cite{silberman2012indoor}, into vertically flipping the scene than it is to achieve a horizontal flip.

In \figref{fig:plots_flip_scaleall_dag}-(d) and \figref{fig:plots_scale_flip_all_indoors}-(c), we plot the ARE achieved by the proposed method for different methods and target depth maps: horizontal flip, vertical flip and different scales. Both flipping tasks are much harder than the scaling tasks. Especially for outdoor settings, fooling the network to produce vertical flipped predictions is the most challenging task as the error is $\approx 15\%$, even with $\xi = 2 \times 10^{-2}$.

\begin{figure}[t]
    \centering
    \includegraphics[width=0.92\textwidth]{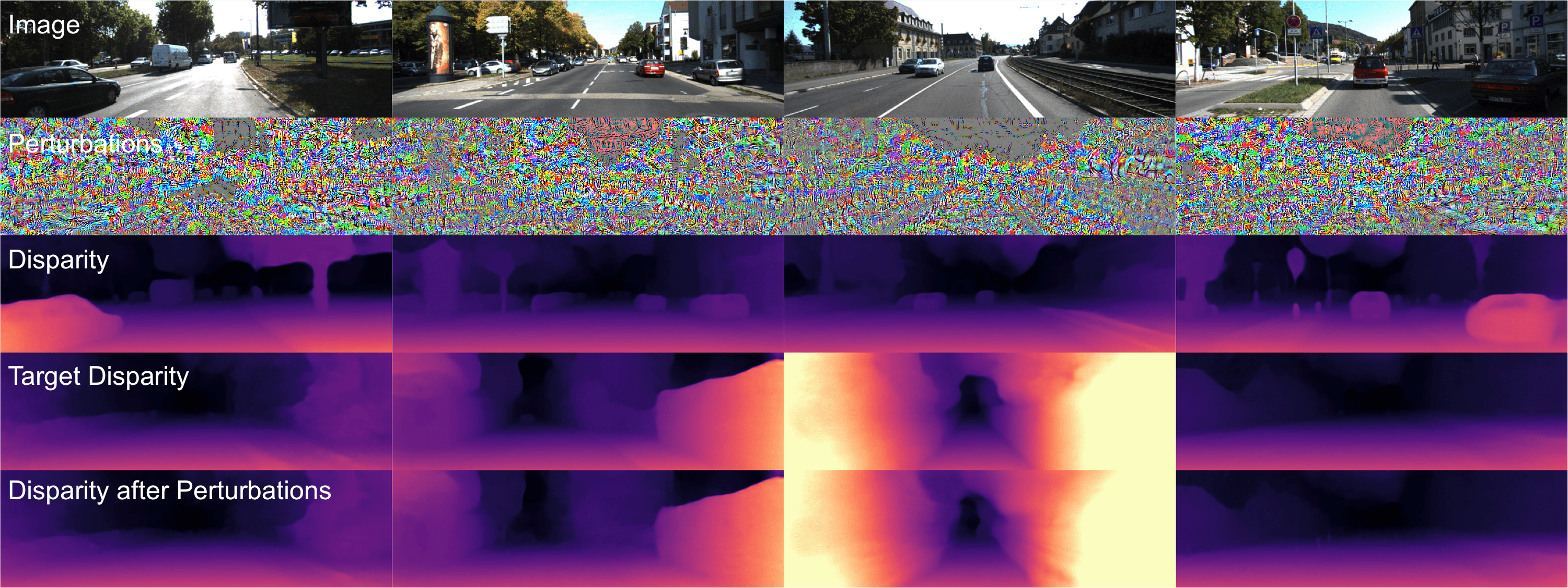}
    \vspace{-0.3em}
    \caption{\textbf{Altering the predicted scene to a preset scene.} Adversarial perturbations can remove a car and replace a road sign with trees (leftmost), add walls (column two), and vegetation (column three) to open streets, and transform an urban environment with vehicles to an open road (rightmost).}
    \vspace{-0.2em}
    \label{fig:preset_scene}
\end{figure}

\vspace{-0.8em}
\subsection{Altering Predictions to Fit a Preset Scene}
\label{sec:preset_scene}
\vspace{-0.5em}
We now examine perturbations for altering the predicted scene $d(x_1) = f_d(x_1)$ to an entirely different pre-selected one $d^t(x_1) = f_d(x_2)$ obtained from images sampled from the same training distribution $x_1,x_2 \sim \mathbb P(x)$: $d(x_1) = f_d(x_1) \neq d^t(x_1) = f_d(x_2)$.

\figref{fig:preset_scene} shows that cars can be removed and road signs can be replaced with trees (leftmost), walls (column two) and vegetation (column three) can be added to the scene, and an urban street with vehicles can be transformed to an open road (rightmost). While perturbations are visually imperceptible, we note that $\|v(x)\|_1 = 0.0348$ (similar to that required for flipping). Even with this amount of noise, an adversary can fool the network to return scenes that are different from the one observed.

Additionally, this experiment also confirms the biases learned by network discussed in \secref{sec:symmetrically_flipping_the_scene}. While perturbations can alter the scene to a preset one with \textit{structures not present} in the image, we have difficulties finding perturbations that can vertically flip the predicted scene.

\vspace{-1.2em}
\section{Localized Attacks on the Scene}
\vspace{-0.8em}
\label{sec:attacking_a_specific_portion}
Given semantic and instance segmentation \cite{Alhaija2018IJCV}, we now examine adversarial perturbations to target localized regions in the scene. Our goal is to fool the network into (i) predicting depths that are closer or farther by a factor of $1 + \alpha$ for all objects belonging to a semantic category, (ii) removing specific instances from the scene, and (iii) moving specific instances to different regions of the scene, all the while keeping the rest of the scene unchanged. 

\vspace{-0.8em}
\subsection{Category Conditioned Scaling}
\vspace{-0.5em}
\label{sec:category_conditional_scaling}
Unlike \secref{sec:scaling_the_scene}, we want to alter a subset of the scene, partitioned by semantic segmentation, such that predictions belonging to an object category (e.g. vehicle, nature, human) are brought closer to or farther from the camera by a factor of $1 + \alpha$ for $\alpha \in \{-0.10, -0.05, +0.05, +0.10\}$. 

\begin{figure}[t]
    \centering
    \includegraphics[width=0.24\textwidth]{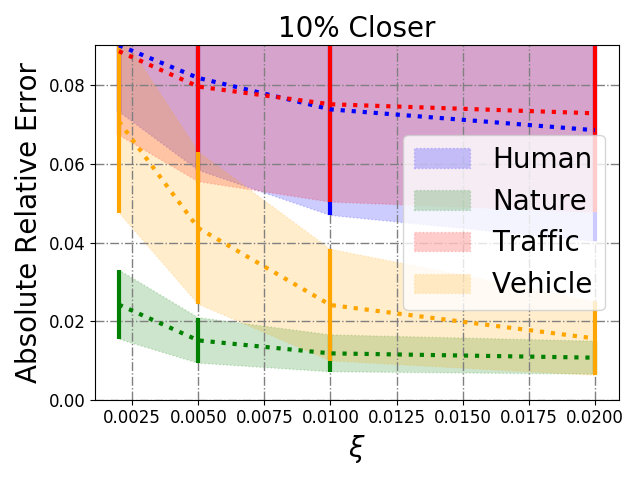}
    \includegraphics[width=0.24\textwidth]{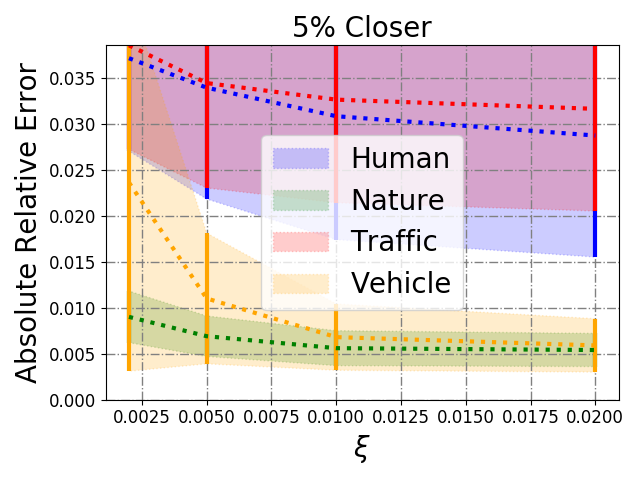}
    \includegraphics[width=0.24\textwidth]{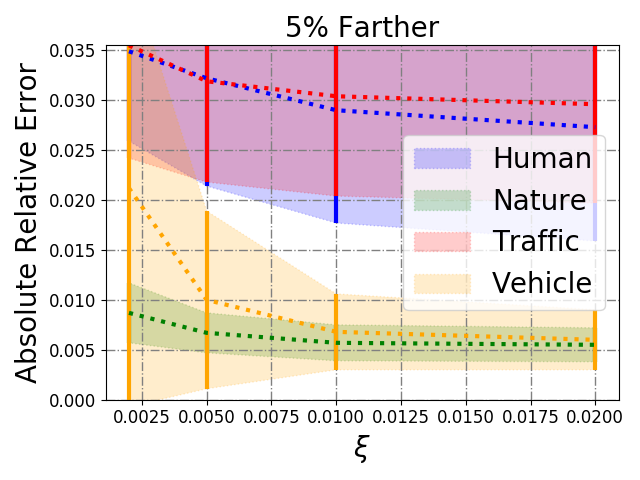}
    \includegraphics[width=0.24\textwidth]{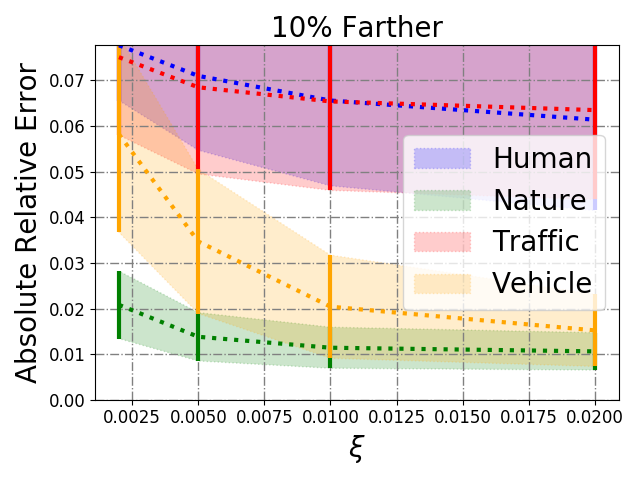}
    \vspace{-0.6em}
    \caption{\textbf{ARE for scaling different categories closer and farther.} It is easier to fool the network to predict vehicle and nature categories closer and farther than is to fool human and traffic categories.}
    \vspace{-0.2em}
    \label{fig:plots_scale_category_human_vehicle_nature}
\end{figure}

We assume a binary category mask $M \in \{0, 1\}^{H \times W}$ derived from a semantic segmentation where all pixels belonging to a category are marked with $1$ and $0$ otherwise. We denote the target depth as
\begin{equation}
    d^{t}(x) = (\mathbf{1}-M) \circ f_d(x) + (1 + \alpha) M \circ f_d(x)  
\end{equation} 
where $\mathbf{1}$ is an $H \times W$ matrix of $1$s. Column three of \figref{fig:teaser} illustrates this problem setting where the perturbations fool Monodepth2 into predicting all vehicles to be 10\% closer to the camera. \figref{fig:plots_scale_category_human_vehicle_nature} shows a comparative study between different categories. Unlike \secref{sec:scaling_the_scene}, it is more difficult to alter a specific portion of the scene without affecting the rest. We surmise that this is due to the network's dependency on global context and hence altering one portion also affects other regions.

Moreover, each category exhibits a different level of robustness to adversarial noise. \textit{Some categories are harder to attack than others}, e.g. traffic signs and human categories ($\approx 3\%$ error for $\alpha = \pm5\%$ and $\approx 6\%$ for $\alpha = \pm10\%$) are harder to alter than vehicle and nature ($\approx 0.5\%$ error for $\alpha = \pm5\%$ and $\approx 1\%$ for $\alpha = \pm10\%$). For interested readers, please see Supp. Mat. for visualizations, additional experiments, and performance comparisons amongst all categories.

\begin{figure}[t]
    \centering
    \includegraphics[width=0.92\textwidth]{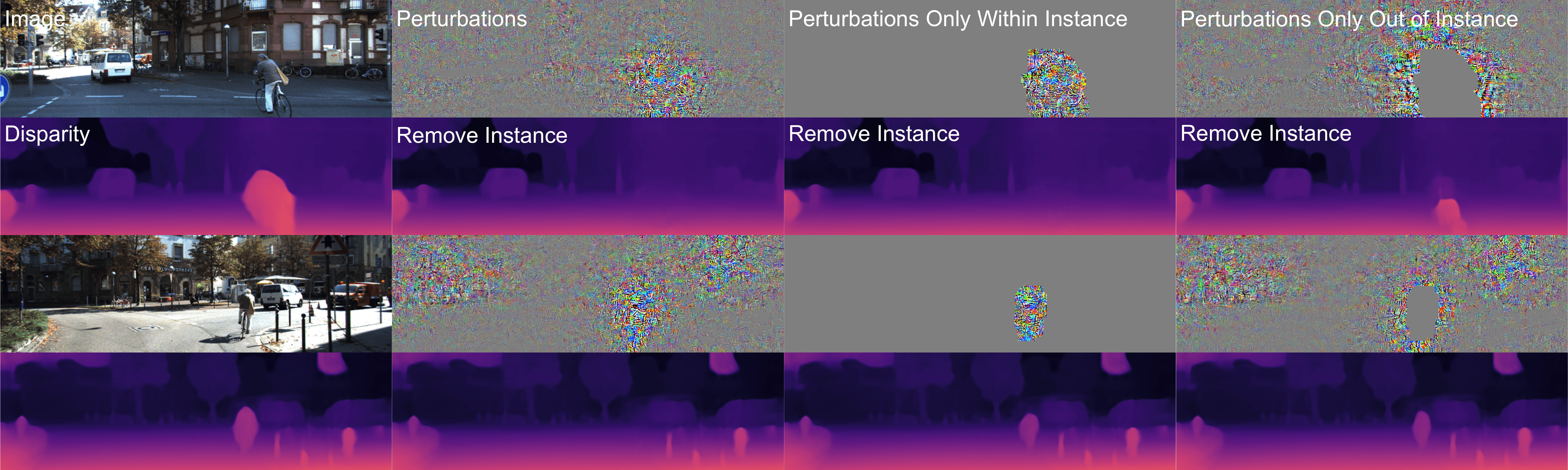}
    \vspace{-0.3em}
    \caption{\textbf{Selectively removing instances of human (bikers and pedestrians).} Removing a localized target requires attacking non-local contextual information. Moreover, one can attack an instance without perturbing it at all. We demonstrate this by constraining the perturbation to be either completely within the instance mask or completely out of the mask.}
    \vspace{-0.2em}
    \label{fig:remove_human}
\end{figure}

\vspace{-0.8em}
\subsection{Instance Conditioned Removing}
\vspace{-0.5em}
\label{sec:instance_conditional_removing}
We now consider the case where instance labels are available for removing a specific instance (e.g. car, pedestrian) from the scene. By examining this scenario, we hope to shed light on the possibility that a depth prediction network can ``miss'' a human or car, which may cause incorrect rendering in augmented reality or an accident in the autonomous navigation scenario. 

Similar to \secref{sec:category_conditional_scaling}, we assume a binary mask $M$, but in this case, of specific instance(s) in the scene, e.g. all pixels belonging to a specific pedestrian are marked with 1 and 0 otherwise. To obtain $d^{t}(x)$, we first remove the depth values in $f_d(x)$ belonging to $M$ by multiplying $f_d(x)$ by $\mathbf{1}-M$. Then, we use the depth values $f_d(x)$ on the contour of $M$ to linearly interpolate the depth in the missing region:
\begin{equation}
\label{eqn:remove_instance}
    d^t(x) = (\mathbf{1}-M) \circ f_d(x) + M \circ d_{M}^{t}(x)
\end{equation}
where $d_{M}^{t}(x) := \texttt{interp}(\texttt{contour}(f_d(x), M))$. Indeed, this scenario is possible. \figref{fig:remove_human} shows examples of pedestrian and biker removal in the driving scenario where perturbations completely remove the targeted instance. With this attack, the road ahead becomes clear, which makes the agent susceptible to causing an accident. 

Even though perturbations are concentrated on the targeted instance region, non-zero perturbations can be observed in the surrounding regions. While the target effect is localized (e.g., make a pedestrian disappear), the perturbation is non-local, implying that the network exploits non-local context, which presents a vulnerability to attacks against a target instance by perturbing other parts of the image. To validate this claim, in the next section, we study perturbations with spatial constraints (either within or outside of the mask) to restrict the information that the adversary can attack.

\vspace{-0.9em}
\subsection{Instance Conditioned Removing with Spatial Constraints on the Perturbations}
\vspace{-0.6em}
\label{sec:instance_conditional_removing_with_spatial_constraints}
Motivated by our results in \secref{sec:instance_conditional_removing}, we extend the instance conditioned removal task to more constrained scenarios where the perturbations have to either exist (i) completely within the target instance mask $M$ or (ii) completely outside of it. 

For perturbations within the targeted instance mask $M$, we constrain $v(x)$ to satisfy $||M \circ v(x)||_{\infty} \leq \xi$ and $||(\mathbf{1}-M) \circ v(x)||_{\infty} = 0$. When constrained within $M$, perturbations can only remove \textit{some instances} successfully (e.g. biker is completely removed in row two, column three of \figref{fig:remove_human}). In other cases, the perturbations can only remove the outer part of the instance, leaving parts of the instance in the scene (row four). This shows that depth prediction networks leverage global context; without attacking the contextual information located outside of $M$ (e.g. without perturbing the entire image as in column two of \figref{fig:remove_human}), it is not always possible to completely remove the target instance. 

Second, we want to answer the question posed in \secref{sec:introduction}. Can perturbations remove an object by attacking anywhere (e.g. a billboard), \textit{but the object} (e.g. a car)? In this more challenging case, the perturbations are constrained to be outside of the instance mask: $||(\mathbf{1}-M) \circ v(x)||_{\infty} \leq \xi$ and $||M \circ v(x)||_{\infty} = 0$. Column four of \figref{fig:remove_human} shows that even though there are no ``direct attacks on the object'' (perturbations in the masked region), the perturbations can \textit{still remove parts of the target instance}. While some of the target instance still remains, our experiment demonstrates that depth prediction networks are indeed susceptible to \textit{attacks against a target instance that does not require perturbing the instance at all}.

\begin{figure}[t]
    \centering
    \subfloat[][]{
        \includegraphics[width=0.46\textwidth]{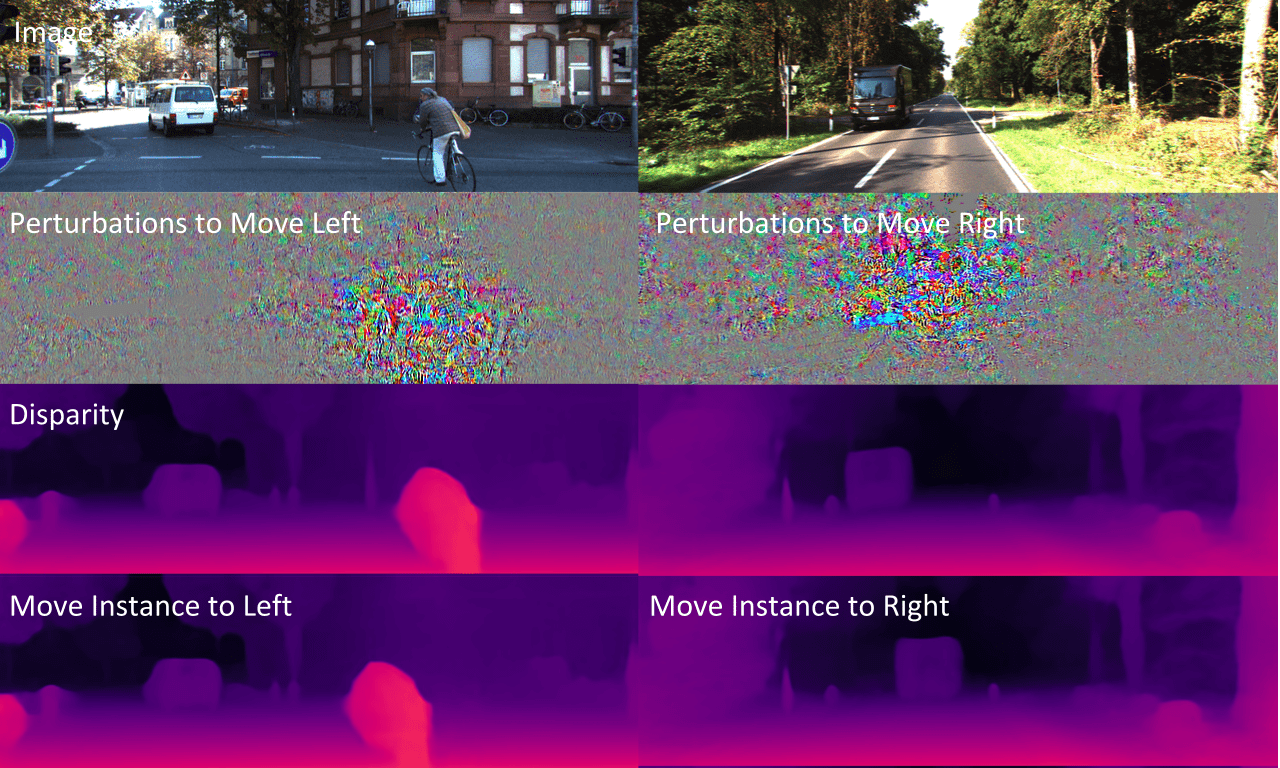}
    }
    \subfloat[][]{
        \includegraphics[width=0.46\textwidth]{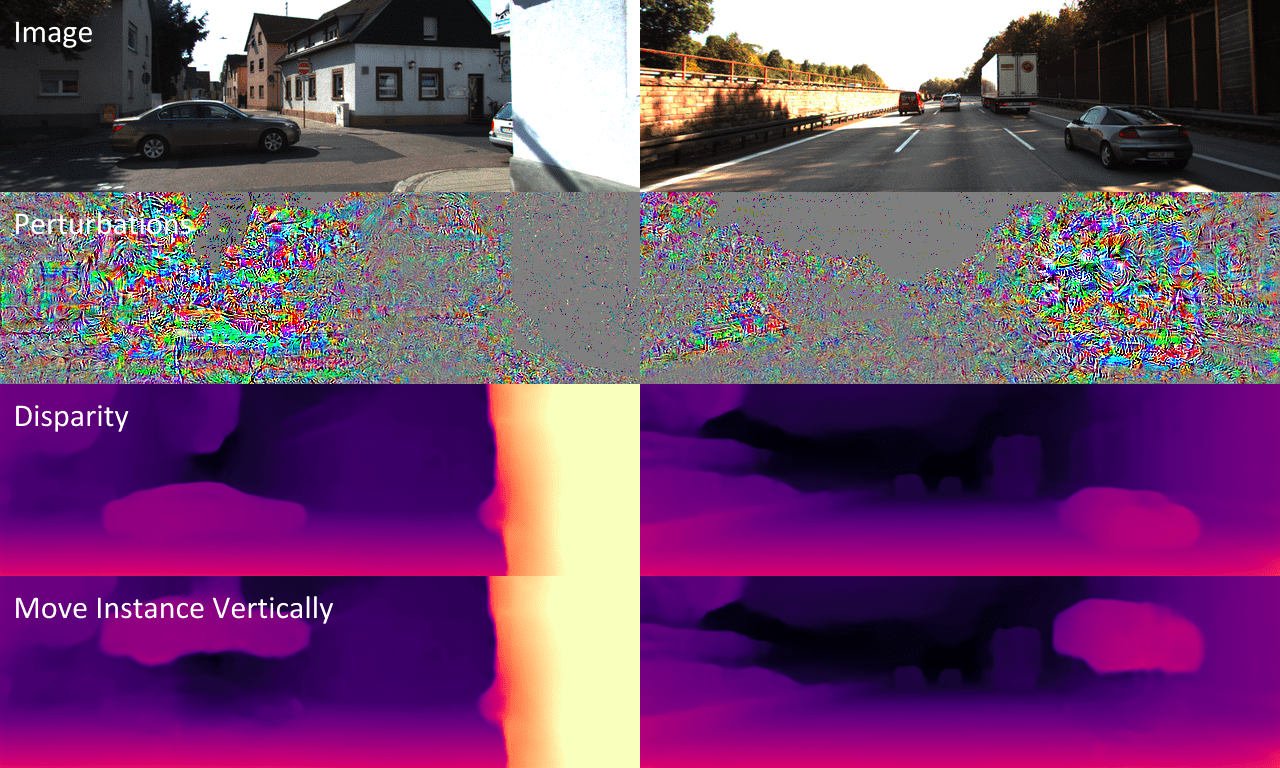}
    }
    \vspace{-0.6em}
    \caption{(a) \textbf{Moving horizontally.} Selected instances is moved by $\approx 8\%$ in the left and right directions while rest of the scene is preserved. (b) \textbf{Flying cars.} A vehicle instance is moved $\approx 42\%$ upward while rest of the scene is preserved. Noise is concentrated around the targeted instance.}
    \label{fig:horizontal_vertical_move}
\end{figure}

\vspace{-0.9em}
\subsection{Instance Conditioned Translation}
\label{sec:instance_conditioned_translation}
\vspace{-0.6em}
In this case study, we examine perturbations for moving an instance (e.g. vehicle, pedestrian) horizontally or vertically in the image space. As \secref{sec:instance_conditional_removing} and \ref{sec:instance_conditional_removing_with_spatial_constraints} have demonstrated the ability to remove localized objects from the scene, we now show that it is possible for perturbations to move such objects to different locations in the scene (removing the instance and creating it elsewhere) while keeping the rest of the scene unchanged.

\figref{fig:horizontal_vertical_move}-(a) shows that perturbations can fool a network to move the target instance by $\approx 8\%$ across the image in the left and right directions. When moved left, the biker (left column) is now in front our vehicle. When moved right, the truck (right column) is in the wrong lane and looks to be on-coming traffic. Moreover, \figref{fig:horizontal_vertical_move}-(b) shows that perturbations can move select instances by $\approx 42\%$ in the upward direction, creating the illusion that there are ``flying cars'' in the scene.

\begin{figure}[t]
    \centering
    \includegraphics[width=0.24\textwidth]{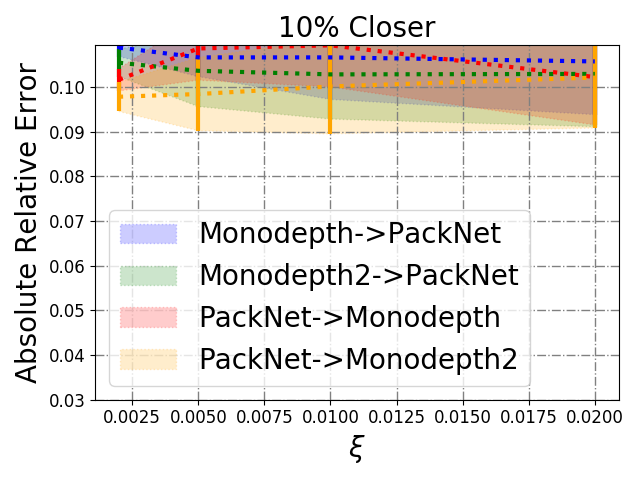}
    \includegraphics[width=0.24\textwidth]{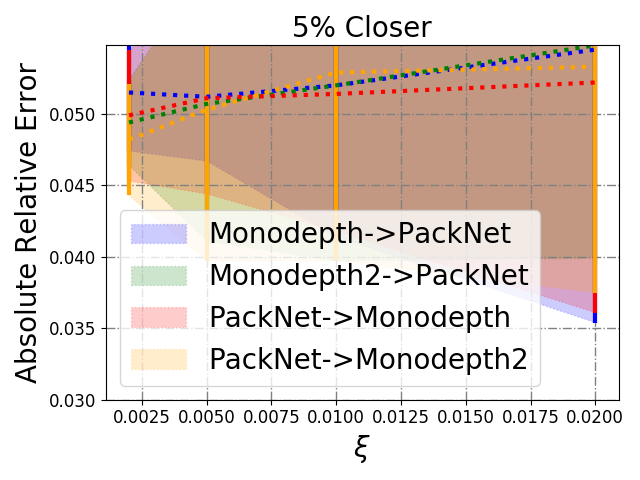}
    \includegraphics[width=0.24\textwidth]{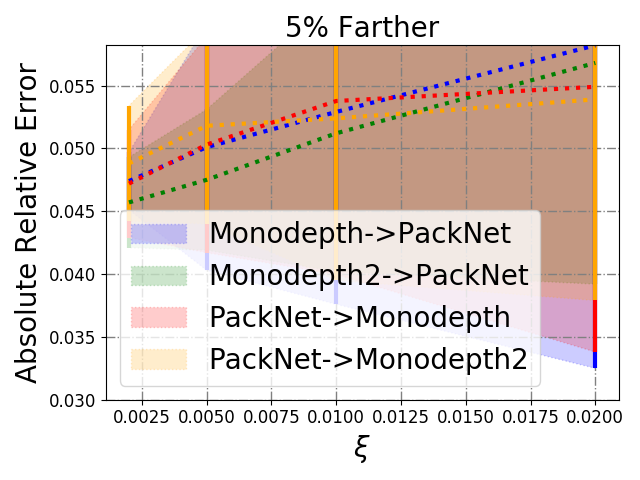}
    \includegraphics[width=0.24\textwidth]{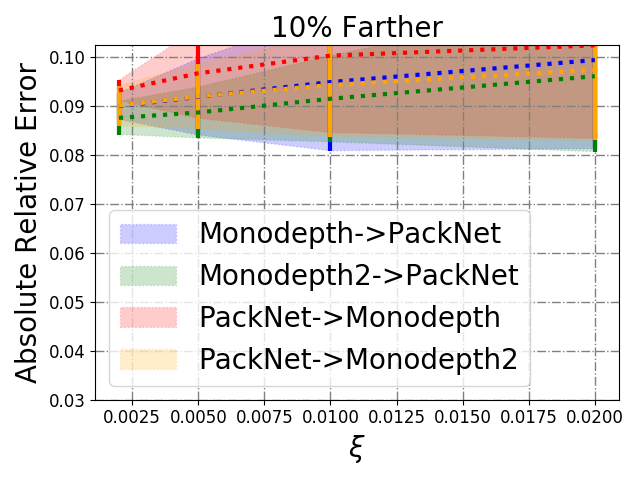}
    \vspace{-0.6em}
    \caption{\textbf{Transferability between networks with different architectures and loss functions.} Perturbations are optimized for Monodepth, Monodepth2 (both with 2D convolutions) and PackNet (3D convolutions) separately and are transferred from Monodepth models to PackNet and vice versa.}
    \vspace{-1em}
    \label{fig:plots_transferable_mono_packnet}
\end{figure}

\begin{figure}[t]
    \centering
    \includegraphics[width=0.24\textwidth]{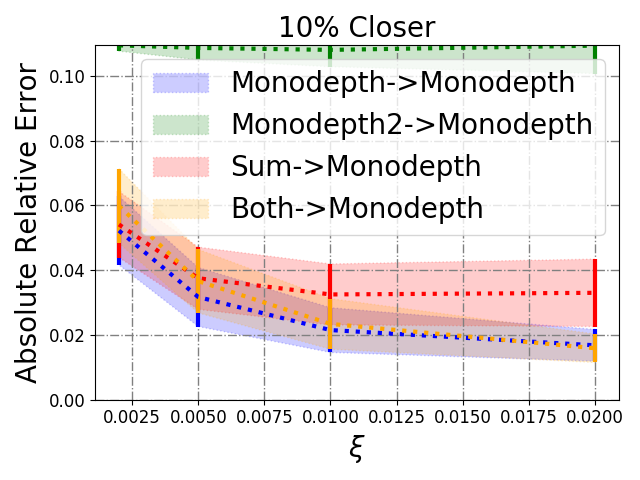}
    \includegraphics[width=0.24\textwidth]{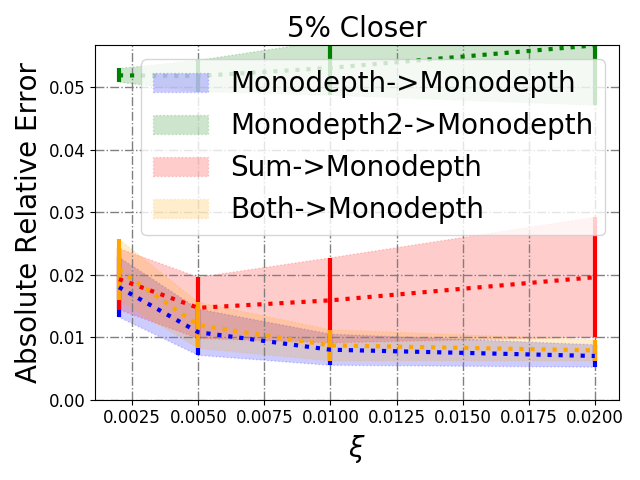}
    \includegraphics[width=0.24\textwidth]{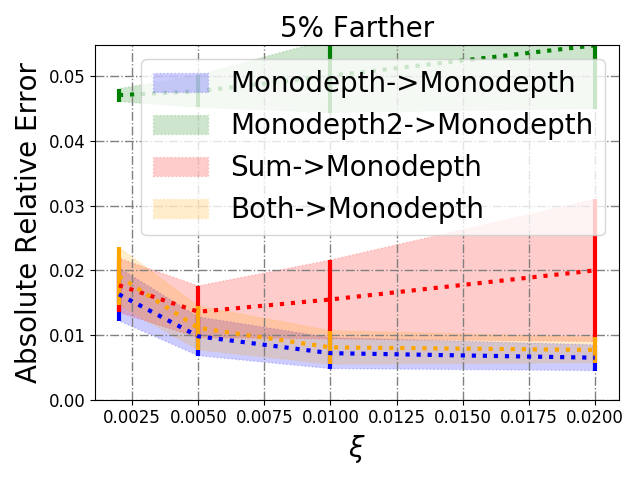}
    \includegraphics[width=0.24\textwidth]{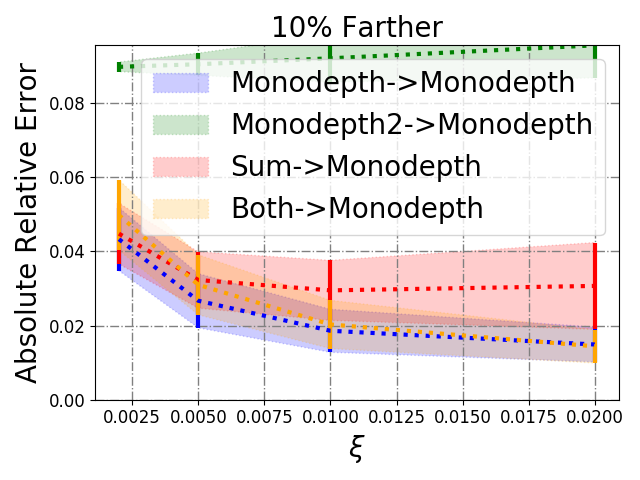}
    \includegraphics[width=0.24\textwidth]{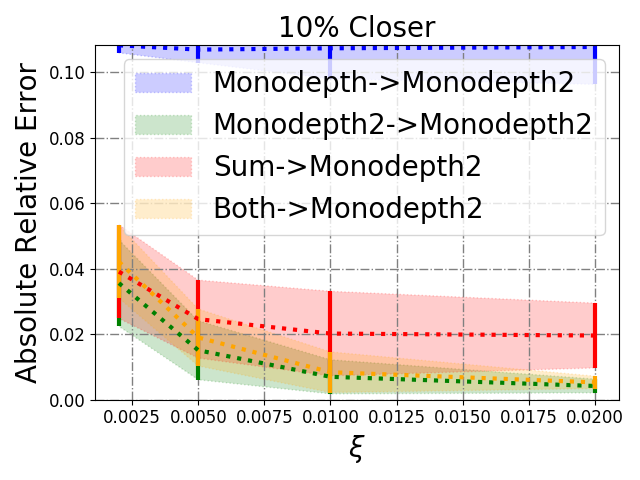}
    \includegraphics[width=0.24\textwidth]{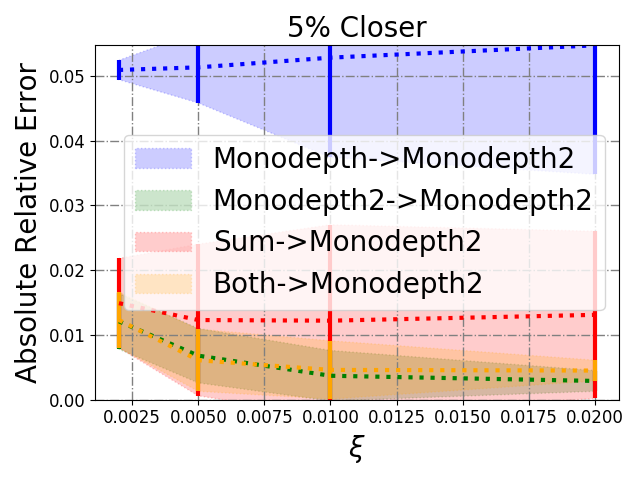}
    \includegraphics[width=0.24\textwidth]{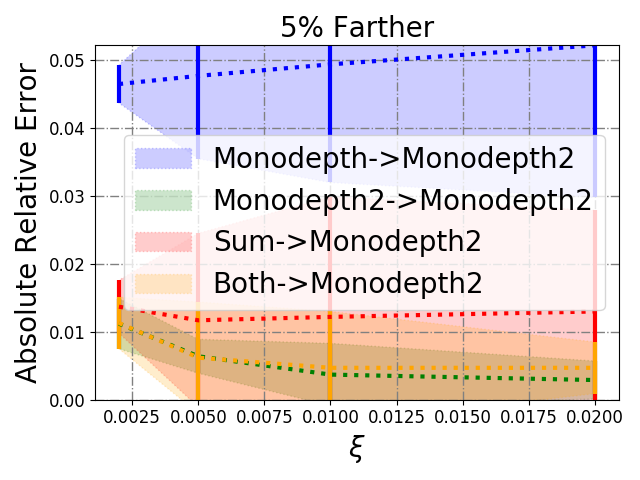}
    \includegraphics[width=0.24\textwidth]{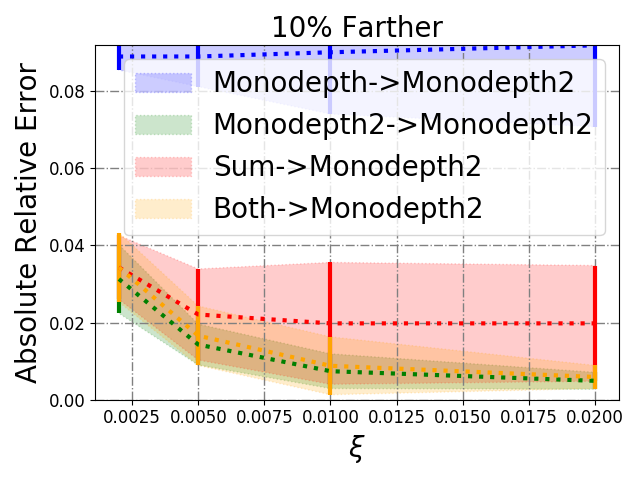}
    \vspace{-0.6em}
    \caption{\textbf{Transferability between networks with similar architectures and loss functions.} Perturbations are (i) optimized for Monodepth and Monodepth2 separately, (ii) optimized for both together and (iii) summed over perturbations calculated for Monodepth and Monodepth2 separately.}
    \vspace{-0.2em}
    \label{fig:plots_transferable_mono}
\end{figure}

\vspace{-1.2em}
\section{Transferability Across Different Models}
\label{sec:transferability}
\vspace{-0.9em}
Transferability is important for black-box scenarios, a practical setting where the attacker does not have access to the target model or its training data. To examine transferability, we test our perturbations crafted for Monodepth2 \cite{godard2019digging} to fool its predecessor Monodepth \cite{godard2017unsupervised} (both use 2D convolutions and similar loss functions) in Monodepth2$\rightarrow$Monodepth, and vice versa in Monodepth$\rightarrow$Monodepth2, for the scene scaling task. (\secref{sec:scaling_the_scene}). To consider networks with different architectures and loss functions, we also test perturbations crafted for Monodepth and Monodepth2 on PackNet \cite{guizilini20203d} (3D convolutions) in \figref{fig:plots_transferable_mono_packnet}. 

To this end, we also optimized perturbations for Monodepth and PackNet to scale the entire scene. Overall, the perturbations optimized for one model does not transfer to another. \figref{fig:plots_transferable_mono_packnet} shows that perturbations crafted for networks with different architectures do not transfer. When comparing models with similar architectures and loss, interestingly, transferability decays with increasing norm (\figref{fig:plots_transferable_mono}), which may be due to perturbations overfitting to the model. We summed the perturbations for Monodepth and Monodepth2 (``Sum'' in \figref{fig:plots_transferable_mono}) and found that their summation can \textit{affect both models} with reduced effects as the upper norm increases. For $\xi = 2 \times 10^{-3}$, the potency is nearly unaffected, meaning, for small norms, their summation can attack both models equally well. Lastly, by optimizing for both models (``Both'' in \figref{fig:plots_transferable_mono}), the \textit{same perturbation} can fool both as if it was optimized for the models individually, with performance indistinguishable from Monodepth$\rightarrow$Monodepth and Monodepth2$\rightarrow$Monodepth2 \textit{across all norms}. This shows that both models share a space that is vulnerable to adversarial attacks. Hence, crafting perturbations for an array of potential models may be an avenue towards achieving absolute transferability across models.

\vspace{-1.2em}
\section{Conclusion}
\vspace{-0.9em}
Depth prediction networks are indeed vulnerable to adversarial perturbations. Not only can such perturbations alter the perception of the scene, but can also affect specific instances, making the network behave unpredictably, which can be catastrophic in applications that involve interaction with physical space. These perturbations also shed light on the network's dependency on non-local context for local predictions, making non-local targeted attacks possible. Despite these vulnerabilities, targeted attacks do not transfer and cannot be crafted in real time, so we are safe for now. Yet, the fact that they exist suggests that there is room to improve the representation learned. We hope that our findings on the network's biases, the effect of context, and robustness of the perturbations can help design more secure and interpretable models that are not susceptible to such attacks.

\section*{Broader Impact}
\vspace{-0.8em}
Adversarial perturbations highlight limitations and failure modes of deep networks. They have captured the collective imagination by conjuring scenarios where AI goes awry at the tune of imperceptible changes. Some popular media and press has gone insofar as suggesting them as proof that AI cannot be trusted.

While monocular depth prediction networks are indeed vulnerable to these attacks, we want to assure the reader that these perturbations cannot cause harm outside of the academic setting. As mentioned in \secref{sec:experiments_details}, optimizing for these perturbations is computationally expensive and hence it is infeasible to craft these perturbations in real time. Additionally, they also do not transfer; so, we see little negative implications for real-world applications. However, the fact that they exist implies that there is room for improvement in the way that we learn representations for depth prediction.

Hence, we see the existence of adversaries as an opportunity. Studying their effects on deep networks is also the first step to render a system robust to such a vulnerability. The broader impact of our work is to understand the corner cases and failure modes in order to develop more robust representations. This, in turn, will improve interpretability (or, rather, reduce nonsensical behavior). In the fullness of time, we expect this research to pay a small contribution to benefit transportation safety. 

\section*{Acknowledgements}
\vspace{-0.8em}
This work was supported by ONR N00014-19-1-2229 and ARO W911NF-17-1-0304.

\small
{
\footnotesize
\bibliographystyle{apalike}
\bibliography{ref}
}

\newpage

\begin{appendices}

\hrule
\hrule height 4pt
\vskip 0.29in
\begin{center}
    {\LARGE{\textbf{Supplementary Materials}}}
\end{center}
\vskip 0.25in
\hrule height 1pt
\vskip 0.29in

\section{Summary of Contents}

In \secref{sec:robustness_against_defense_mechanisms}, the robustness of perturbations against defenses is discussed. Additional implementation details that we could not fit into main text due to space constraints are given in \secref{sec:additional_implementation_details_outdoor_scenario}. We verify our claim that targeted adversarial perturbations are visually imperceptible in \secref{sec:impercetibility}. More experimental results on changing the scale of the scene are provided in \secref{sec:scaling_with_larger_factors}. In \secref{sec:adversarial_attacks_for_indoor_scenes}, existence of the successful adversarial attacks for indoor scenes (NYU-V2) is shown for state-of-the-art indoor monocular depth prediction model. In \secref{sec:linear_operations}, we examine how predictions behave when linear operations are applied to perturbations (sum of two perturbations and linear scaling of a perturbation). Failure cases for the perturbations are analyzed in \secref{sec:failure_cases}.
Finally, in \secref{sec:additional_results_on_outdoors}, more qualitative and quantitative results are provided for the experiments whose compressed versions are presented in the main text.

\section{Robustness of the Targeted Attacks Against Defense Mechanisms}
\label{sec:robustness_against_defense_mechanisms}
In the main text, we have shown that depth prediction networks are prone to adversarial attacks. In this section, we will examine the robustness of the perturbations against common defense mechanisms: (i) Gaussian blurring and (ii) adversarial training.

\subsection{Defense through Gaussian Blurring} 
\label{sec:defense_gaussian_blur}
\begin{figure}[h]
    \centering
    \includegraphics[width=0.24\textwidth]{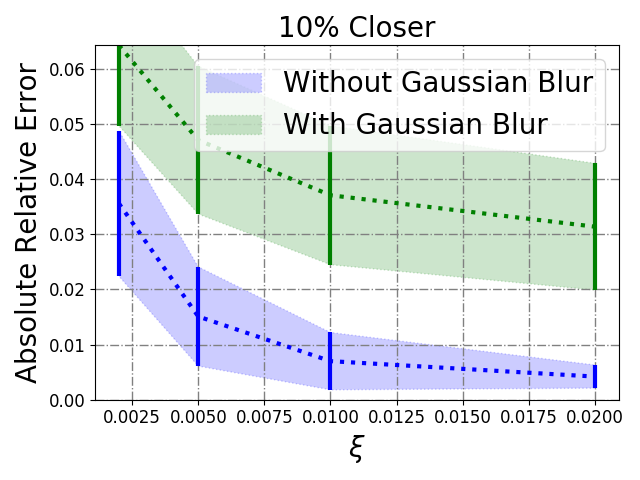}
    \includegraphics[width=0.24\textwidth]{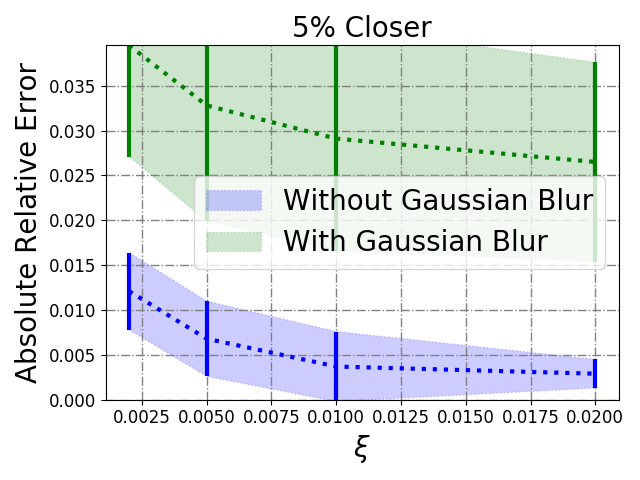}
    \includegraphics[width=0.24\textwidth]{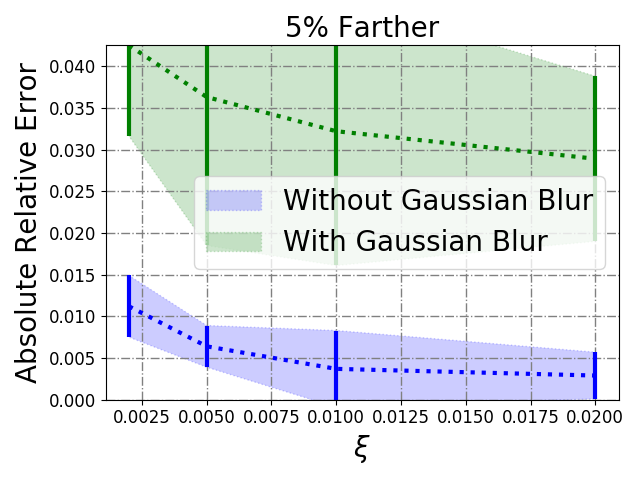}
    \includegraphics[width=0.24\textwidth]{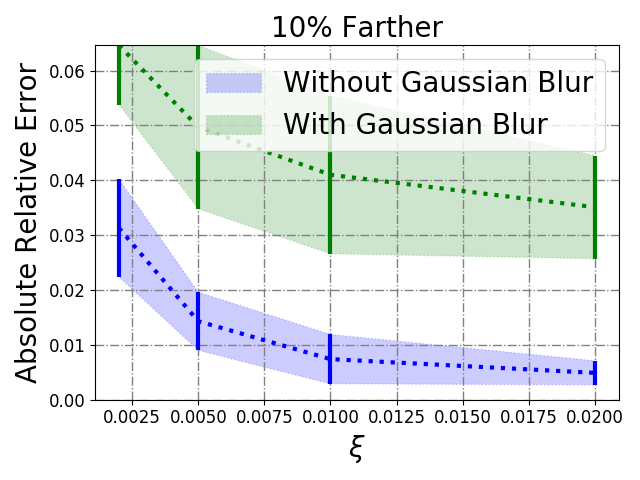}
    \caption{\textbf{Gaussian blur.} Absolute relative error (ARE) achieved by adversarial perturbations of different norms ($\xi$) for different scales. For each scale, we plot the ARE with and without Gaussian blur. Even though absolute relative error increases with the Gaussian blur, the proposed method can still find a small norm noise to alter the scene.}
    \label{fig:plots_gaussian_blur}
\end{figure}

In \figref{fig:plots_gaussian_blur}, we show the effect of Gaussian blurring as a simple defense mechanism on our targeted attacks by blurring the image with additive perturbations. Even though Gaussian blur does reduce the effectiveness of the perturbations (increased ARE over all scales), the resulting scene is still only $\approx 3\%$ away from a target depth that is 10\% closer or farther than the original predictions for $\xi = 2 \times 10^{-2}$. This is the performance that the method achieves for the case of $\xi = 2 \times 10^{-3}$ without blurring. In other words, the effect of the blurring can be suppressed simply by increasing the upper norm of the noise by $10\times$ for scaling the scene by $\pm 10\%$.

\subsection{Defense through Adversarial Training} 
\begin{figure}[t]
    \centering
    \includegraphics[width=0.24\textwidth]{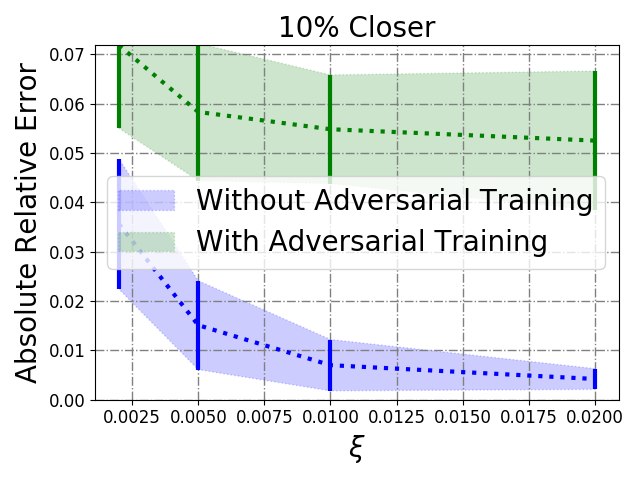}
    \includegraphics[width=0.24\textwidth]{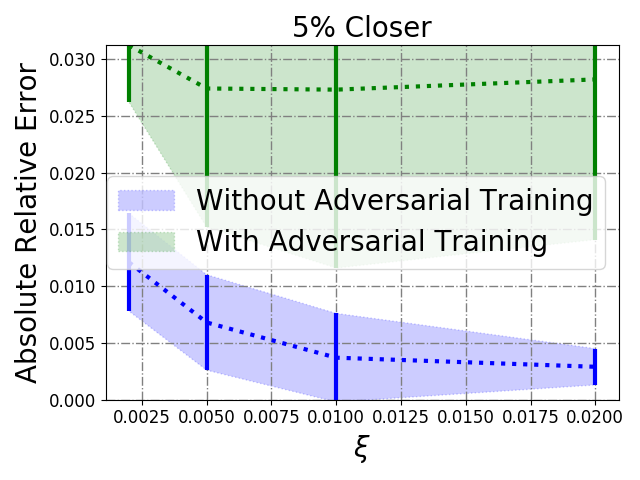}
    \includegraphics[width=0.24\textwidth]{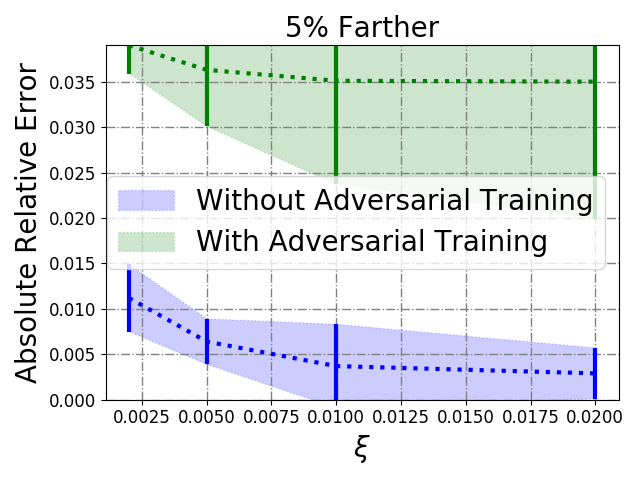}
    \includegraphics[width=0.24\textwidth]{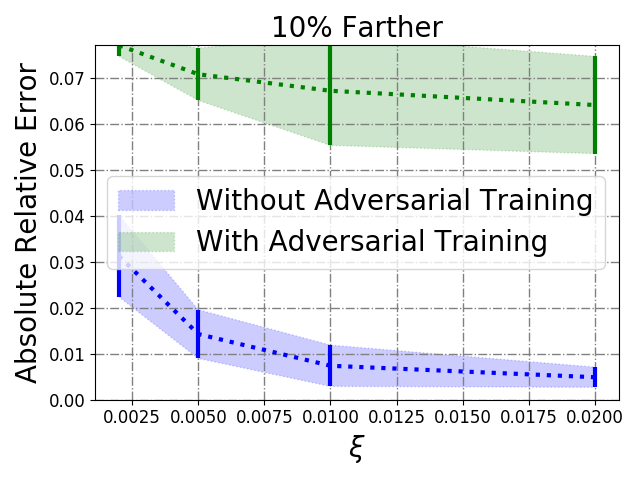}
    \caption{\textbf{Adversarial training.} Absolute relative error (ARE) achieved by adversarial perturbations of different norms ($\xi$) for different scales. For each scale, we plot the ARE with and without adversarial training. Even though absolute relative error increases with the adversarial training, the perturbations can still affect the predicted scene with small norm noise.}
    \label{fig:plots_adversarial_training}
\end{figure}

To examine the robustness of adversarial perturbations to adversarial training, we crafted adversarial perturbations for scaling the scene by a factor of $1 + \alpha$ where $\alpha \in \{ -0.10, -0.05, +0.05, +0.10 \}$ for the KITTI Eigen split \cite{eigen2015predicting} (consisting of 22600 stereo pairs). We trained Monodepth2 \cite{godard2019digging} by minimizing the normalized discrepancy between the predicted depth of a perturbed image ($f_d(x + v(x))$) and its prediction for the original image ($f_d(x)$).
\begin{equation}
    \ell(x, v(x), f_d) = \frac{\| f_d(x) - f_d(x + v(x)) \|_1}{f_d(x)}
\label{eqn:adversarial_loss}
\end{equation}
\figref{fig:plots_adversarial_training} shows the effect of the perturbations on Monodepth2 after adversarial training. While training does reduce the influence of adversarial perturbations on the scene scaling task, it does not make the network invariant to the adversarial perturbations. With perturbations of $\xi = 2 \times 10^{-3}$, the predicted scene is $\approx 7\%$ from the target scene that is $10\%$ closer than or farther from the original and $\approx 3\%$ from the target scene that is $5\%$ closer or farther. For $\xi = 2 \times 10^{-2}$, perturbations can still fool the network to be predict a target scene scaled by $\pm 10\%$ with  $\approx 5\%$ absolute relative error and $\approx 2\%$ error for fooling the network to predict a target scene that is scaled by $\pm 10\%$. 

\begin{figure}[t]
    \centering
    \includegraphics[width=0.24\textwidth]{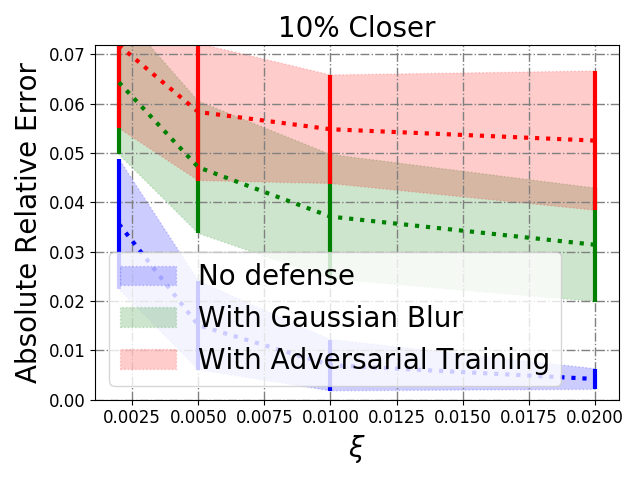}
    \includegraphics[width=0.24\textwidth]{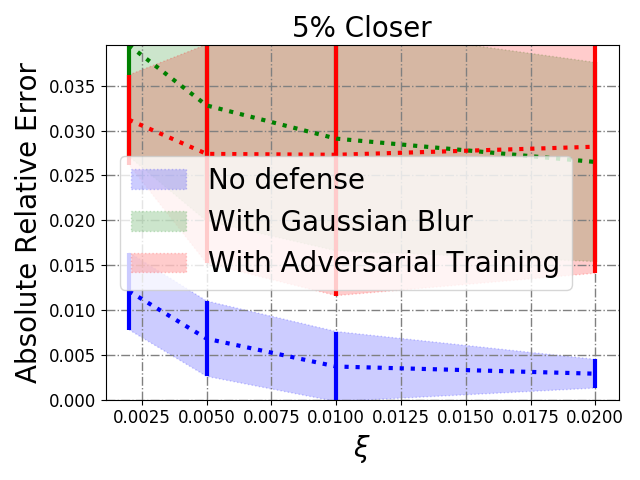}
    \includegraphics[width=0.24\textwidth]{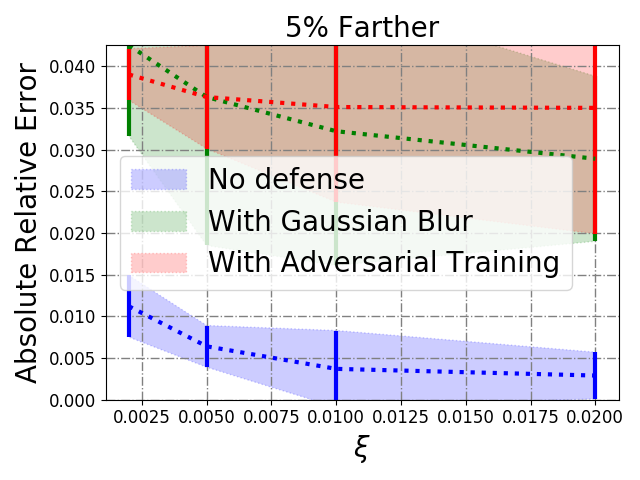}
    \includegraphics[width=0.24\textwidth]{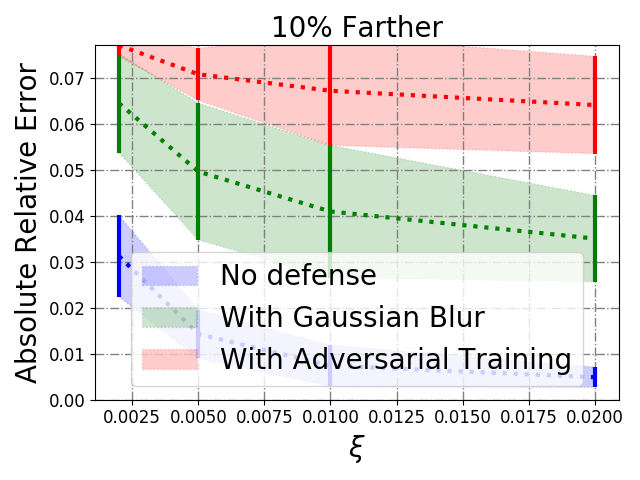}
    \caption{\textbf{Adversarial training vs Gaussian blur.} Absolute relative error (ARE) achieved by adversarial perturbations of different norms ($\xi$) for different scales. For each scale, we plot the ARE without any defense, with Gaussian blur and with adversarial training. Both Gaussian blur and adversarial training makes the depth prediction network more robust to perturbations. Performances of the two defense mechanisms are comparable for small norms ($\xi$), but adversarial training is more effective on larger norms.}
    \label{fig:plots_blur_adversarial_training}
\end{figure}

To compare the two defense mechanisms, we refer to \figref{fig:plots_blur_adversarial_training}. For smaller norms, e.g. $\xi = 2 \times 10^{-3}$, we observe a similar performance in using Gaussian blur (\secref{sec:defense_gaussian_blur}) and adversarial training as defenses against adversarial perturbations; whereas, adversarial training is clearly better for larger $\xi$. This may be due to Gaussian blur's ability to destroy the perturbation for small norms and, hence, able to mitigate the effect of the perturbations. However, for larger norms, the blurring does not corrupt the perturbations enough and therefore does not reduce the effect of perturbations by as much.

\section{Additional Implementation Details for Outdoor Scenario}
\label{sec:additional_implementation_details_outdoor_scenario}
In this section, we provide the additional implementation details for crafting adversarial perturbations for Monodepth \cite{godard2017unsupervised} and Monodepth2 \cite{godard2019digging} on the KITTI dataset (outdoor driving scenario) as discussed in the main text.
\subsection{Hyper-parameters}
\begin{table}[h]
    \centering
    \begin{tabular}{l c c c c}
        \midrule 
       Upper Norm & $\xi = 2 \times10^{-3}$ & $\xi = 5 \times 10^{-3}$ & $\xi = 1 \times 10^{-2}$ & $\xi = 2 \times 10^{-2}$ \\
        \midrule 
        Monodepth & $\eta=1.0$ & $\eta=2.0$ & $\eta=3.0$ & $\eta=4.0$ \\
        \midrule 
        Monodepth2 & $\eta=0.1$ & $\eta=1.0$ & $\eta=3.0$ & $\eta=5.0$ \\
        \midrule 
        \end{tabular}
    \caption{\textbf{Learning rates.} We achieve the best performances with the given learning rates.}
    \label{table:learning_rates}
\end{table}

Regarding hyper-parameters for crafting \textit{adversarial perturbations}: We search the learning rate for each noise norm from the set $\{0.1, 1.0, 2.0, 3.0, 4.0, 5.0, 10.0\}$. We report the best performing ones in \tabref{table:learning_rates}. Regarding our choice for the number of steps to run, we experimented with 200, 400, 500, 800, and 1000 steps and found little difference in performance measured by ARE between 500 steps, and 800 and 1000 steps. While an increased number of steps will obtain slight performance improvements, conscious of the time complexity, we chose 500 for our experiments. 

Regarding hyper-parameters for \textit{adversarial training}: As a defense against adversarial perturbations, we optimized  \eqnref{eqn:adversarial_loss} for Monodepth2 using Adam \cite{kingma2014adam} with $\beta_1 = 0.9$ and $\beta_2 = 0.999$. We used a batch size of 4 and starting learning rate of $1 \times 10^{-5}$. We decreased the learning rate to $5 \times 10^{-6}$ after 10 epochs and to $2.5 \times 10^{-6}$ after 20 epochs and $1 \times 10^{-6}$ after 30 epochs for a total of 40 epochs. Training takes approximately 4 hours using an Nvidia GeForce GTX 1080 GPU.

\subsection{Monodepth, Monodepth2, and PackNet}
We study the effects of adversarial perturbations on the state-of-the-art monocular depth prediction method, PackNet \cite{guizilini20203d}, Monodepth2 \cite{godard2019digging} and its predecessor Monodepth \cite{godard2017unsupervised}. Monodepth2 and Monodepth models utilize similar 2D convolutional network architectures and are also trained with similar loss functions. PackNet is built on 3D convolutions and uses a different loss function than Monodepth models. In this section, we provide details on the three methods.

Regarding \textit{Monodepth}: Monodepth uses a ResNet50 encoder architecture as its backbone and a standard decoder with skip connections. Monodepth predicts both left and right disparities from a single image (assuming it is the left image of a stereo-pair) and uses image reconstruction as supervision. Additionally, it is trained with a standard local smoothness term weighted by image gradients and a left-right disparity consistency term as its regularizers.

Regarding \textit{Monodepth2}: Monodepth2, unlike Monodepth, uses ResNet18 encoder (pretrained on ImageNet) as its backbone network architecture. Rather than simply minimizing an image reconstruction loss, Monodepth2 leverages a heuristic to discount occluded pixels and also uses a criterion to discount static frames. Similar to Monodepth, Monodepth2 also minimizes a local smoothness regularizer weighted by image gradients.

Regarding \textit{PackNet}: PackNet uses 3D convolutions. While the general architecture is still of the encoder-decoder form, they use ``packing'' and ``unpacking'' to convert 2D features into 3D features. PackNet and the Monodepth methods all leverage structure from motion, but PackNet also uses velocity at every time step as a prior to obtain metric depth.

\section{Imperceptibility}
\label{sec:impercetibility}
While we have provided the $L_\infty$ norm (from $2 \times 10^{-3}$ to $2 \times 10^{-2}$) and $L1$ norm (ranging from $\approx$0.0124 to $\approx$0.0348, depending on the target scene) of the perturbations in the main text, it is difficult to quantify how ``visually imperceptible'' these perturbations really are. The real test is whether a human can spot the difference between the original and perturbed image. Hence, we conducted a study with 28 people where we ask if two images (the original image paired with itself or its associated perturbed image with the highest norm, $\xi = 0.02$) are the same. The perturbation types were randomly chosen from scaling $\pm10\%$, horizontal, or vertical flipping. Of the 40 random pairs (20 perturbed), users identified only 1 perturbed image on average, which verifies our claim on the imperceptibility of the perturbations. 

\section{Scaling with Larger Factors}
\label{sec:scaling_with_larger_factors}

\begin{figure}[t]
    \centering
    \includegraphics[width=0.33\textwidth]{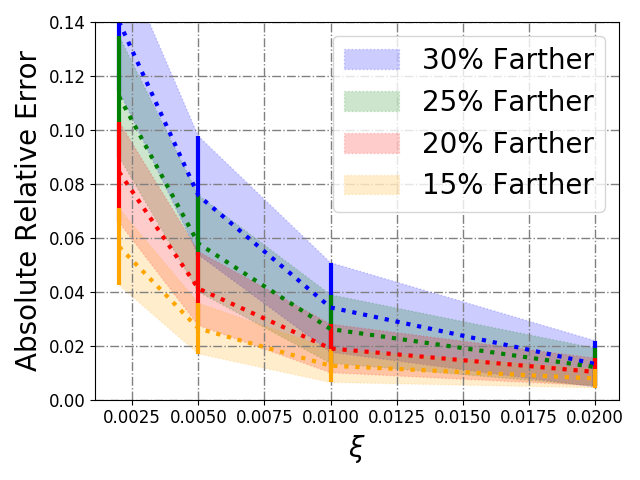}
    \includegraphics[width=0.33\textwidth]{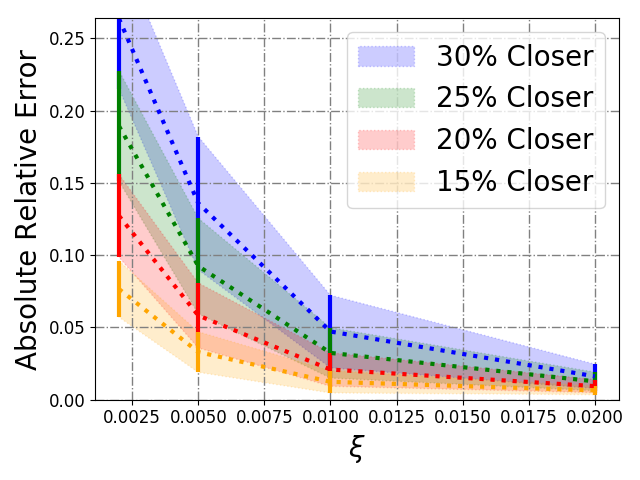}\caption{\textbf{ARE with various upper norm $\xi$ for scaling Monodepth2 predictions by larger factors.} We increased the scaling to $15\%$, $20\%$, $25\%$ and $30\%$ closer and farther. We can see that for $\xi = 2 \times 10^{-2}$, perturbations are still able to scale the entire scene by $\approx 1\%$ error.}
    \label{fig:plots_large_scales}
\end{figure}

In Sec. 4.1 and Fig. 2-(a, b) of the main text, we showed that it is possible, even for small norms such as $\xi = 2 \times 10^{-3}$ to scale the scene to be $5\%$ or $10\%$ closer or farther with small error. In this section, we demonstrate that it is possible to scale the scene by larger amounts (up to $30\%$ closer or farther). \figref{fig:plots_large_scales} shows that with $\xi = 2 \times 10^{-3}$, it is only able to scale the up to $15\%$ with reasonable error; whereas perturbations with $\xi = 5 \times 10^{-3}$ can achieve this up to $20\%$ closer or farther. However, using larger norms ($1 \times 10^{-2}$ and $\xi = 2 \times 10^{-2}$), one can scale the scene up to $30\%$ with small errors (less than $5\%$ ARE for $\xi = 1 \times 10^{-2}$ and $\approx 1\%$ ARE for $\xi = 2 \times 10^{-2}$). 

To see how far we can push for each upper norm, \figref{fig:plots_scale_limitation} shows various scales that each upper norm is capable of achieving. We note that $\xi = 2 \times 10^{-2}$, is still able to obtain less than $2\%$ ARE when scaling the scene by $45\%$; however, standard deviation starts to grow larger as the scaling increases. 

\begin{figure}[t]
    \centering
    \subfloat[][]{
        \includegraphics[width=0.33\textwidth]{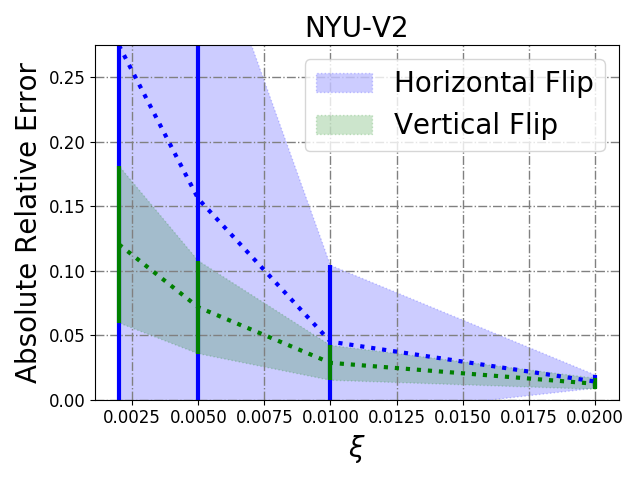}
    }
    \subfloat[][]{
        \includegraphics[width=0.33\textwidth]{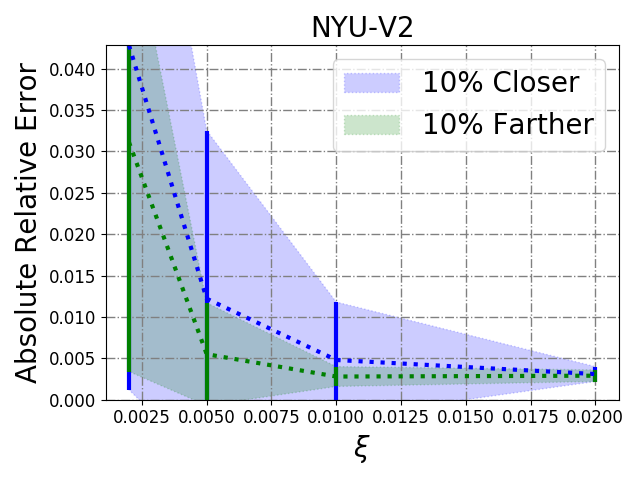}
    }
    \subfloat[][]{
        \includegraphics[width=0.33\textwidth]{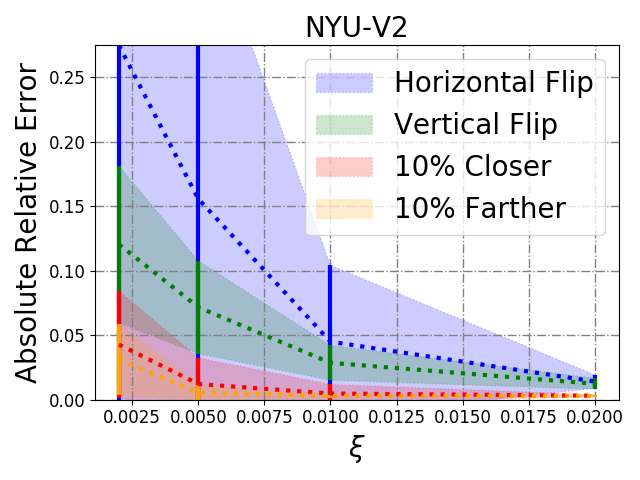}
    }
    \caption{\textbf{Indoor quantitative results.} ARE with various upper norm $\xi$ for scaling and flipping VNL predictions. (a) Results for horizontally and vertically flipping the predictions. (b) Results for scaling the scene by $\pm10\%$. (c) comparison between scaling and flipping tasks. }
    \label{fig:indoor_plots}
\end{figure}

\section{Adversarial Attacks for Indoor Scenes}
\label{sec:adversarial_attacks_for_indoor_scenes}

\begin{figure}[h]
    \centering
    \subfloat[][]{
        \includegraphics[width=0.24\textwidth]{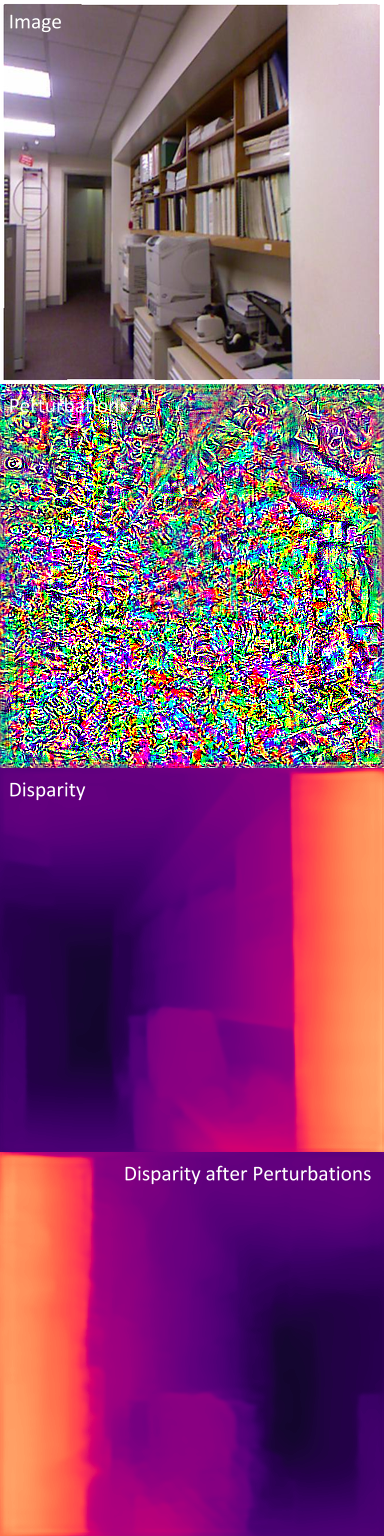}
    }
    \subfloat[][]{
        \includegraphics[width=0.24\textwidth]{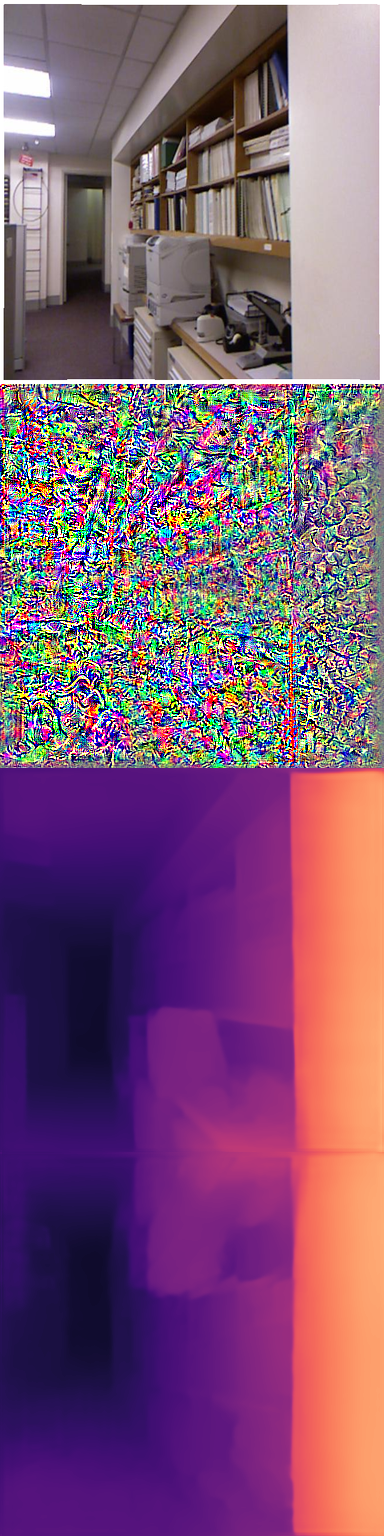}
    }
    \subfloat[][]{
        \includegraphics[width=0.24\textwidth]{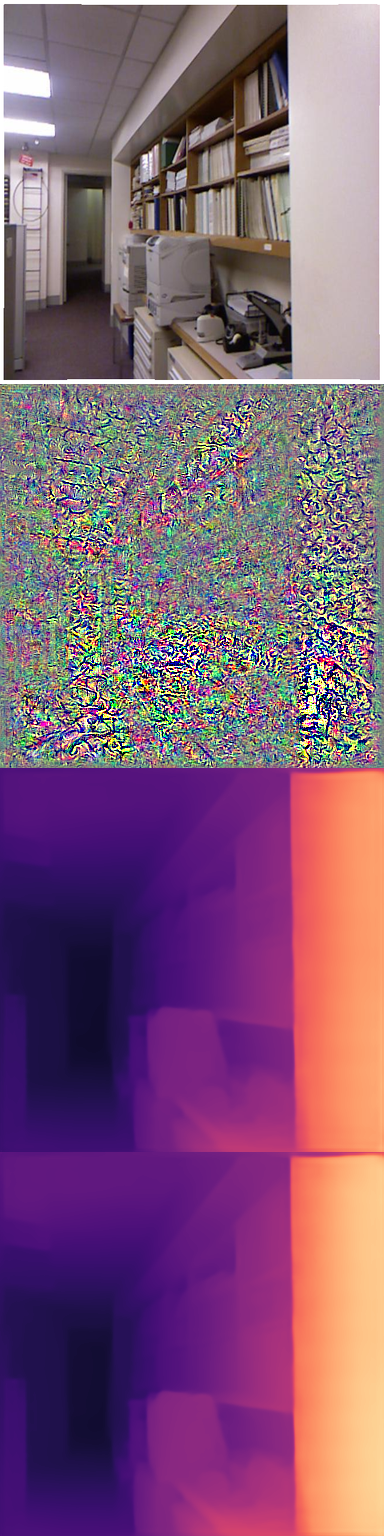}
    }
    \subfloat[][]{
        \includegraphics[width=0.24\textwidth]{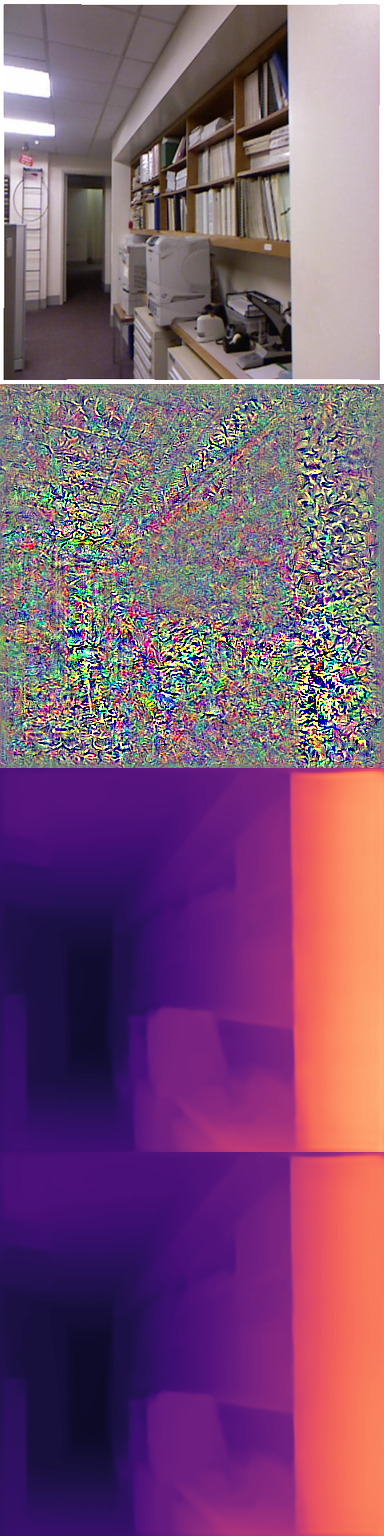}
    }
    \caption{\textbf{Indoor qualitative results.} From left to right: (a) horizontal flip, (b) vertical flip, (c) scale 10\% closer and (d) scale 10\% farther. From top to bottom: RGB, noise, original disparity, disparity prediction for the perturbed image.}
    \label{fig:indoor_figures}
\end{figure}

To show the applicability of the adversarial method on indoor scenes, we examine the adversarial perturbations for Kinect Dataset NYU Depth V2 (NYU-V2) \cite{silberman2012indoor}. We tested the effectiveness of adversarial perturbations on Virtual Normal Loss (VNL) \cite{yin2019enforcing} which is the state-of-the-art monocular depth prediction method for NYU-V2, trained in the supervised setting. 

\subsection{Implementation Details}
NYU-V2 consists of 1449 RGBD images gathered from a wide range of buildings, comprising 464 different indoor scenes across 26 scene classes. The images were hand-selected from 435,103 video frames, to ensure diversity. 1449 labeled samples are split into 795 training and 654 test images. 

In the method proposed by \cite{yin2019enforcing}, a 3D point cloud is reconstructed from the estimated depth map. Then, three non-colinear points are randomly sampled with large distances to form a virtual plane. The deviation between ground truth and prediction for the direction of the normal vector corresponding to the plane is penalized. The pre-trained ResNeXt-101 \cite{xie2017aggregated} model on ImageNet \cite{deng2009imagenet} is used as the backbone architecture. During training, images are cropped to the size 384$\times$384 for NYU-V2. We use the same image resolution for our experiments. The training set is randomly sampled from 29K images of the raw unlabeled training set. 

The time it takes to forward an image with this method is $\approx 0.15$ seconds ($\approx 7$ times more than Monodepth2). Due to computational limitations, we choose the first $20$ images out of $654$ images of the test split for our experiments. We run SGD for $1000$ steps. The learning rate is kept at $10.0$ for the entire optimization.

\subsection{Scaling and Symmetrically Flipping the Scene}
In \figref{fig:indoor_plots}, we compare the performance for different target depth maps: scaling the scene by $\pm10\%$, horizontal and vertical flipping. Unlike outdoor case (KITTI), in the indoor (NYU-V2) horizontal flipping is a harder task than vertical flipping. Achieving a vertically flipped scene was expected to be simpler as layouts of indoor scenes are more diverse. Hence, the depth network does not overfit to a particular layout type e.g. the one in which there are large depth values only at the top of the image. Horizontal flipping being relatively harder for indoor scenes can be explained by the large divergence between the depth distributions of the original predictions and the target depths. The reason is that for the indoor scenes, most scenes are not symmetric in the horizontal direction, unlike the outdoor driving scenario where the left and right parts of the scene from the ego-view are usually symmetric.

Since \cite{yin2019enforcing} normalizes images with the deviation of the dataset which approximately scales the image by $5$, it also effectively scales the noise with the same deviation. But, since the relative norm is still the same, we use the same norm values $\xi \in \{2 \times10^{-3}, 5 \times 10^{-3}, 1 \times 10^{-2}, 2 \times 10^{-2}\}$ when plotting the ARE in \figref{fig:indoor_plots}. 

In \figref{fig:indoor_figures}, we present qualitative results for NYU-V2 for $\xi = 2 \times 10^{-2}$. Small, white borders around RGB images exist in the raw dataset. For all the tasks, including vertical flip, adversarial perturbations manage to fool the model to predict the target depth with small errors. For horizontally flipped target depth (a), predictions have more artifacts than vertical flipped depth (b) and scaled depths (c,d). 

\section{Linear Operations} 
\label{sec:linear_operations}

\begin{figure}[h]
    \centering
    \includegraphics[width=0.49\textwidth]{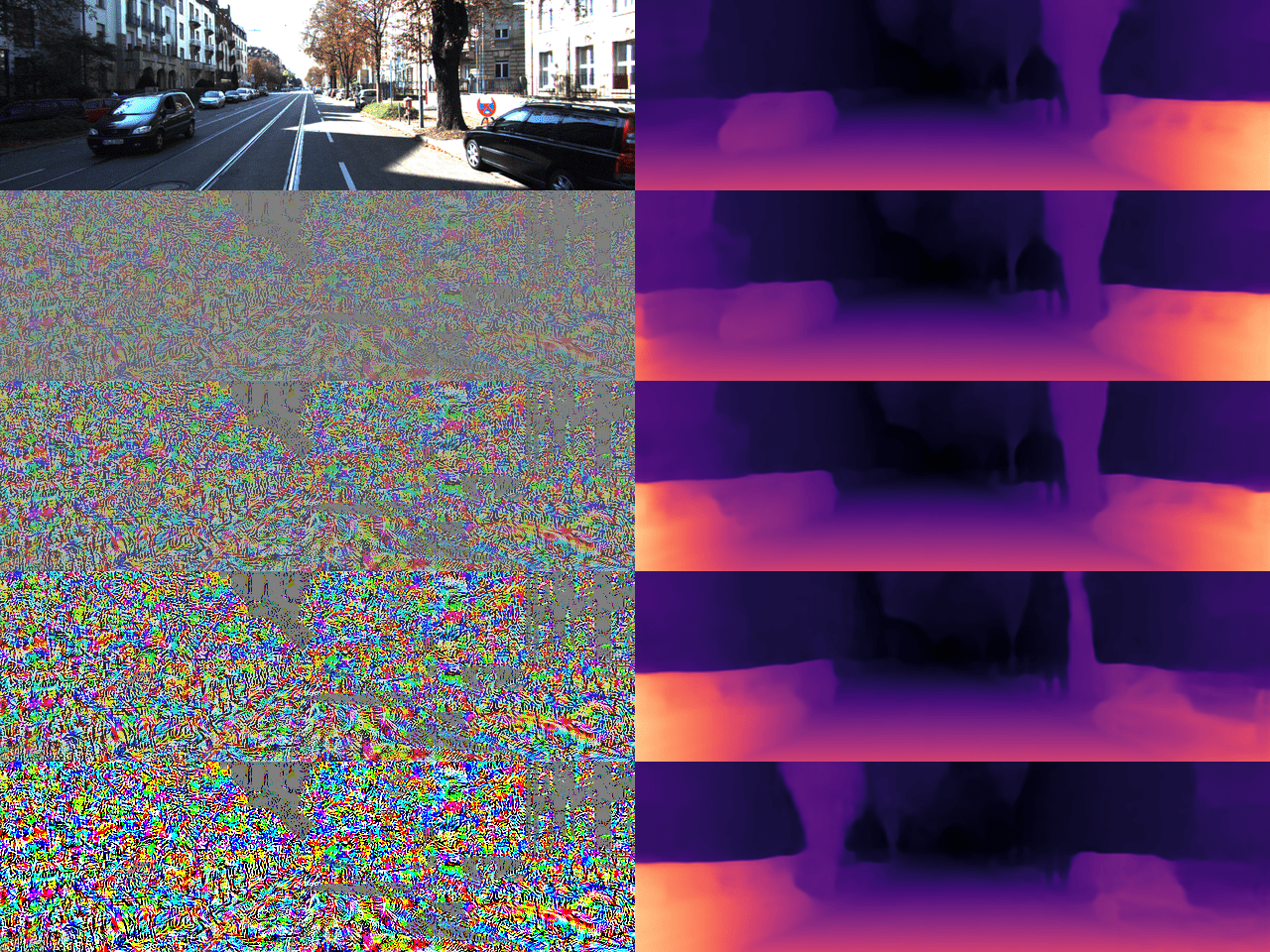}
    \includegraphics[width=0.49\textwidth]{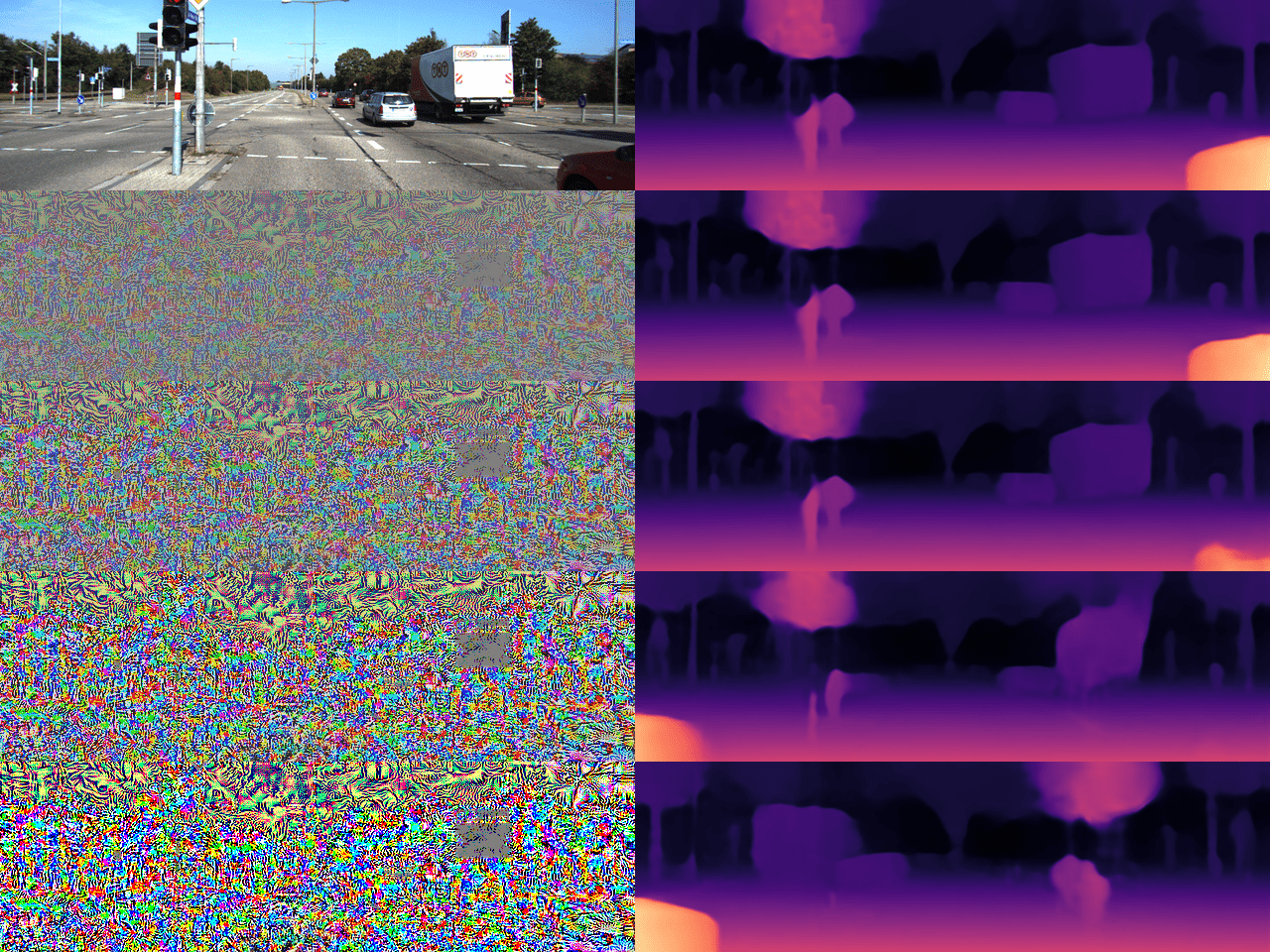}
    \caption{\textbf{} Disparity for $x + \gamma v(x)$ where $\gamma$ is $0.0$, $0.25$, $0.5$, $0.75$, $1.0$ from top to bottom. $v(x)$ is calculated for $d^t = \texttt{fliph}(f_d(x))$. So, the top is the original disparity map while bottom most is the flipped one. In between, portions of the scene are flipped smoothly.}
    \label{fig:scaled_noise_horizontal_flip}
\end{figure}

\begin{figure}[h]
    \centering
    \includegraphics[width=0.49\textwidth]{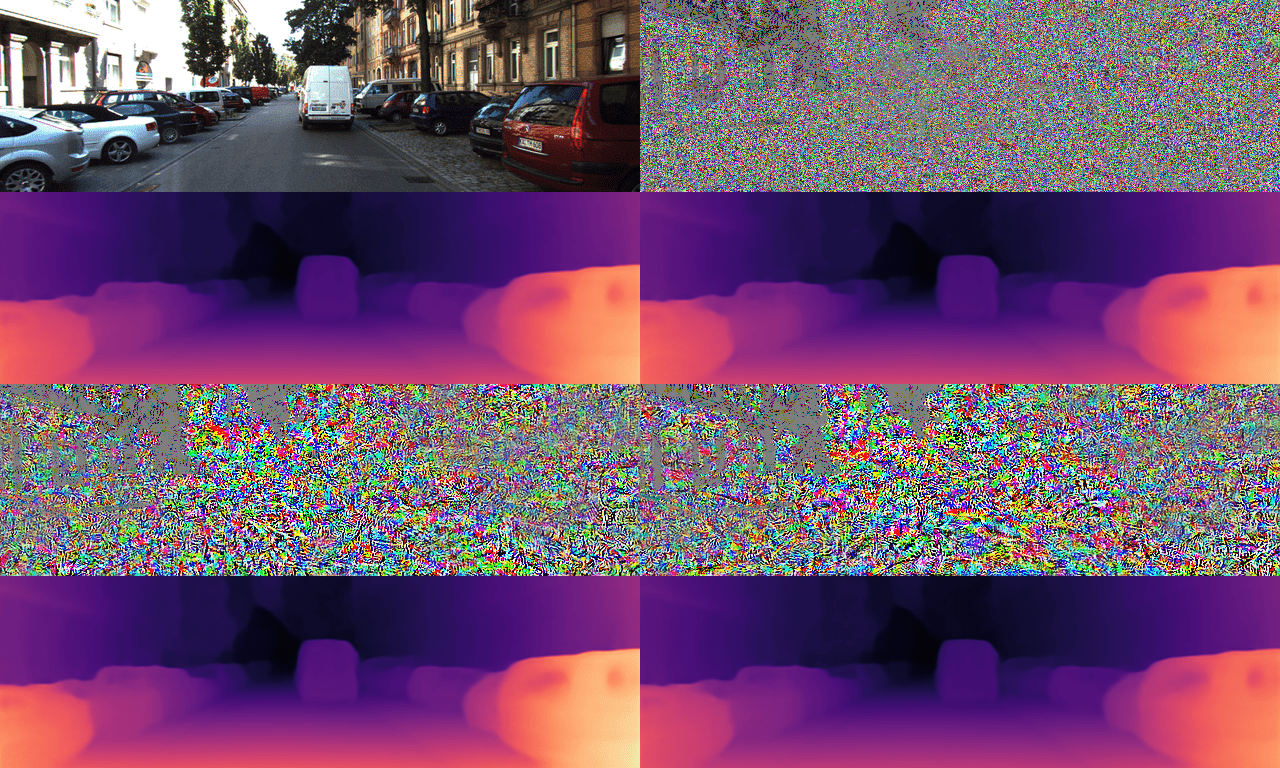}
    \includegraphics[width=0.49\textwidth]{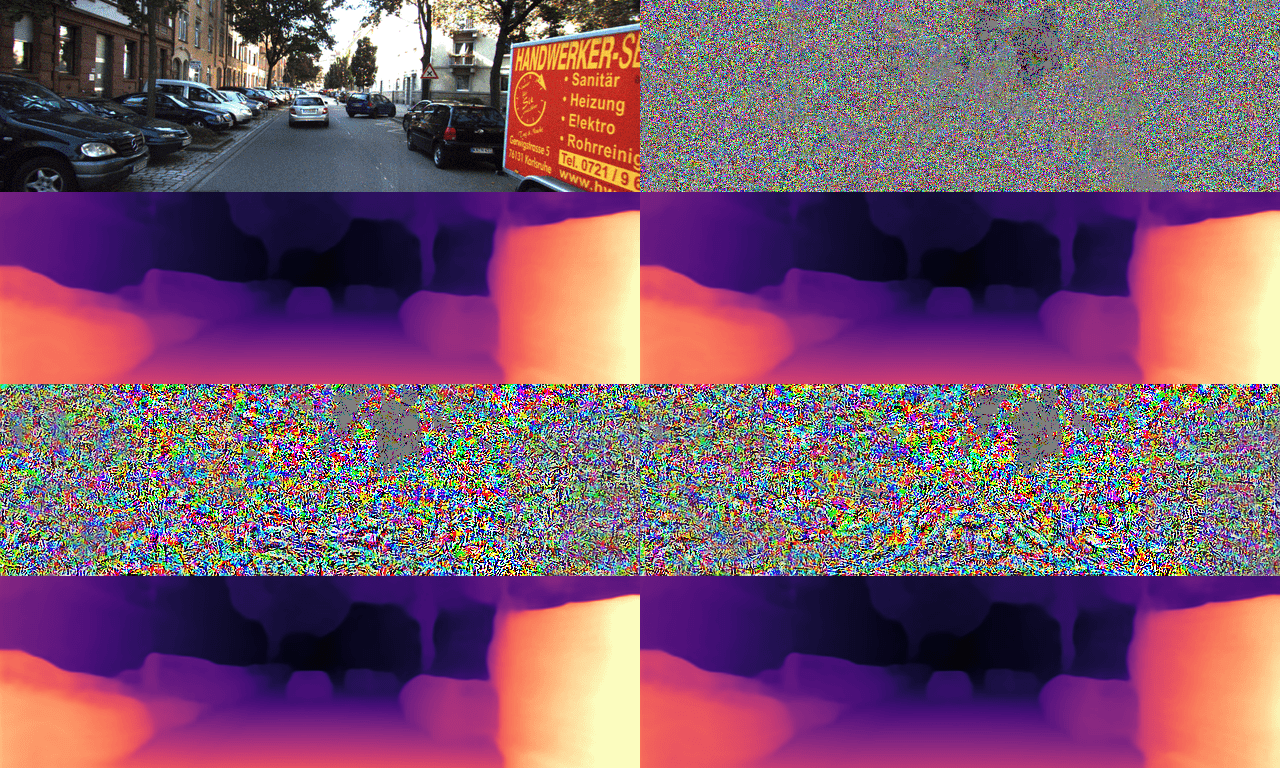}
    \caption{\textbf{} (1st row) left to right: RGB, sum of noises. (2nd row) left to right: original disparity, disparity when two noises $v_1(x)+v_2(x)$ are added to the image. (3rd row) left to right: noise for $10 \%$ closer, noise for $10 \%$ farther. (4th row): Disparity predictions for the images perturbed with the noises in the 3rd row. When added, two noises cancel each other's effect: the scale for $f_d(x+v_1(x)+v_2(x))$ is close to the original one $f_d(x)$.}
    \label{fig:combine_10_-10_scale}
\end{figure}

To better understand how predictions of the depth network changes within a ball of small radius, we examine the effect of linear operations on perturbations. Specifically, we visualize the predictions for the scaled perturbations and for the perturbations which we get after summing two perturbations calculated for two different target depth maps. 

In \figref{fig:scaled_noise_horizontal_flip}, we take the perturbation $v(x)$, which we calculated to horizontally flip the prediction for the given image, and we visualize the prediction of the network for $x + \gamma v(x)$ where $\gamma \in \{0.0, 0.25, 0.5, 0.75, 1.0\}$. As can be seen, between $\gamma=0$ and $\gamma=1$, the scene is smoothly flipped. This implies that the adversarial perturbations can be used to control the depth prediction in a disentangled way. In other words, one causal factor (e.g. horizontal orientation) of the prediction can be independently controlled by tweaking $\gamma$ only, keeping everything else the same. 

As observed before, noise is small for the white regions. See the third column, where there is a gray rectangle in the noise corresponding to the white region of the truck. We speculate the reason behind this phenomenon as the white color being on the border of the support of RGB images. But, the noise is still large for black regions which are at the other extreme of the support (see perturbations corresponding to black vehicles). So, we left further understanding of this phenomenon as future work.

In \figref{fig:combine_10_-10_scale}, we take $v_1(x)$ and $v_2(x)$ which are optimized to scale the scene to 10\% closer and 10\% farther. Then, visualize the summed perturbation, $v(x) = v_1(x) + v_2(x)$ and the prediction for $x + v(x)$. As can be seen, two noises cancel each other: $||v_1(x)|| \approx ||v_2(x)|| \gg ||v_1(x) + v_2(x)||$. Furthermore, the prediction for the image perturbed with the summed noise is close to the original prediction: $f_d(x) \approx f_d(x + v_1(x) + v_2(x))$. This shows that two perturbations with inverse functionalities can neutralize their effects when applied together.

\section{Failure Cases} 
\label{sec:failure_cases}

\begin{figure}[]
    \centering
    \includegraphics[width=0.24\textwidth]{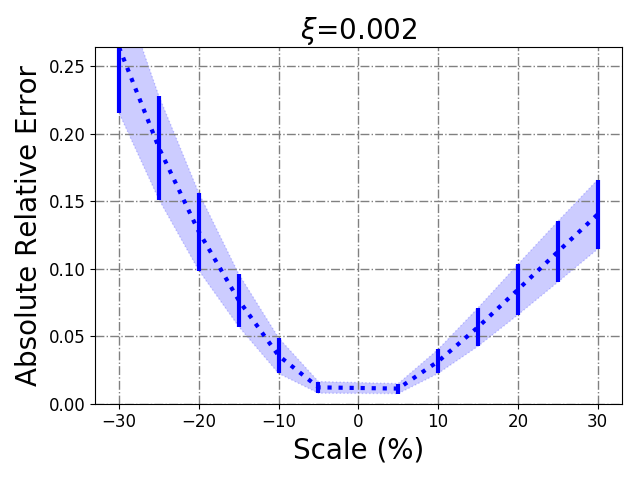}
    \includegraphics[width=0.24\textwidth]{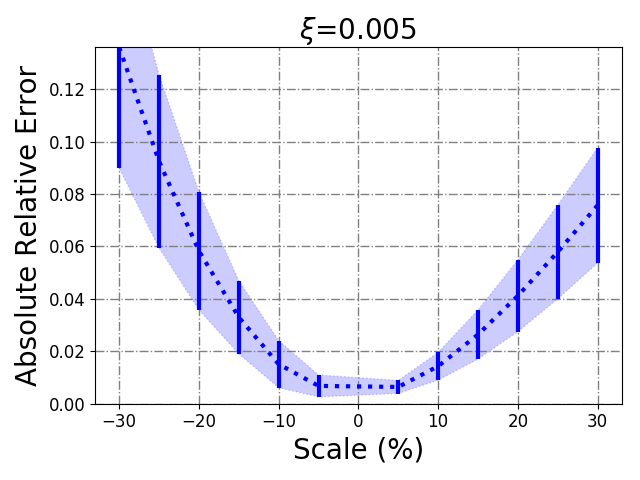}
    \includegraphics[width=0.24\textwidth]{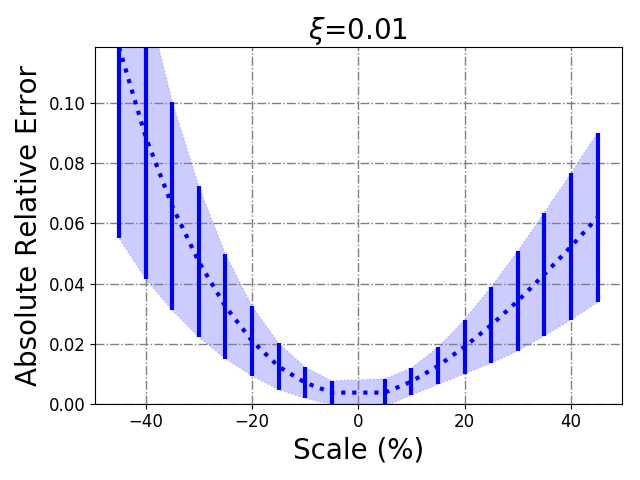}
    \includegraphics[width=0.24\textwidth]{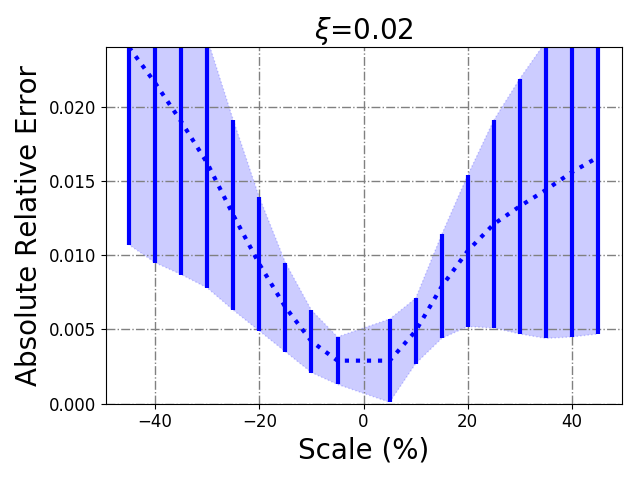}
    \caption{\textbf{ARE with various upper norm $\xi$ for scaling Monodepth2 predictions.} This time error is plotted for large scale ratios $15\%$, $20\%$, $25\%$ and $30\%$ (up to $45\%$ for larger $\xi$), for scaling both closer and farther, showing the limitations of each norm for the scaling task. While ARE is still relatively small for larger norms, standard deviation grows larger -- meaning the perturbations can no longer scale the scene consistency with low error.}
    \label{fig:plots_scale_limitation}
\end{figure}
\begin{figure}[]
    \centering
    \includegraphics[width=1.00\textwidth]{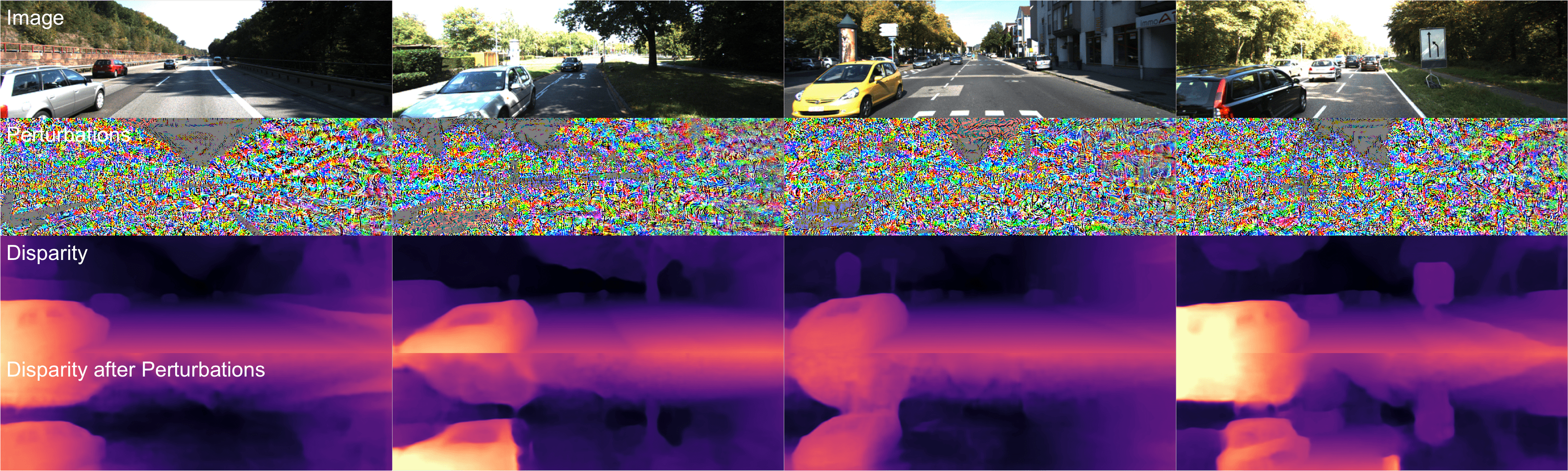}
    \caption{\textit{Additional examples of failure cases for vertical flip}. While adversarial perturbations with $\xi = 2 \times 10^{-2}$, can fool Monodepth2 to predicted scenes that are scaled by large amounts, and horizontally flipped. They cannot cause the network to vertically flip the scene, leaving behind cars and roads as artifacts.}
    \label{fig:vertical_flip_fail}
\end{figure}

\figref{fig:plots_scale_limitation} shows the absolute relative error (ARE) with respect to the target scaling factor for each upper norm. As we can see, for smaller norms of $2 \times 10^{-3}$ and $5 \times 10^{-3}$, the perturbations are limited to scaling the scene by $\approx 15\%$ and $\approx 20\%$, respectively. Scaling factors higher than such increases the ARE by $\approx 4\%$ for every 5\% increase in scaling factor, signaling the limit for these norms. For larger norms of $1 \times 10^{-2}$ and $2 \times 10^{-2}$, the perturbations can afford to scale the scene by a much larger factor. For $\xi = 1 \times 10^{-2}$, perturbations can scale the scene by as much as $30\%$ closer and farther less than $5\%$ error. Whereas, for $\xi = 1 \times 10^{-2}$, perturbations can scale the scene up to $\pm 45\%$ with less than $2\%$ ARE. However, while large scaling still as a low ARE, the standard deviation for larger norms increases drastically showing that it can no longer consistently scale the scene.

While for smaller scales (e.g. $\pm 5$, $\pm 10$) the ARE and amount of noise required is approximately the same (see Sec. 4.1, main text), suggesting similar difficulty levels. As we plot the errors for larger scales in \figref{fig:plots_scale_limitation}, scaling the scene farther generally yields lower error than scaling the scene closer. 

\figref{fig:vertical_flip_fail} shows additional examples of failure cases for vertical flip. While we have shown in the main paper as well as \secref{sec:scaling_with_larger_factors} and \ref{sec:additional_results_on_outdoors} that it is possible to manipulate the scene with small norm perturbations, we show here that perturbations cannot fool a network into vertically flipping the scene.


\newpage

\section{Additional Results on Outdoor Scenarios} 
\label{sec:additional_results_on_outdoors}
In this section, we show (i) side by side visualization of the perturbations required to scale the scene, (ii) additional visualizations of perturbations to horizontally and vertically flip the scene, (iii) quantitative results on targeted attacks to semantic categories and (iv) qualitative results on targeted attacks to instances.

\subsection{Scaling the Scene}
\begin{figure}[h]
    \centering
    \includegraphics[width=1.00\textwidth]{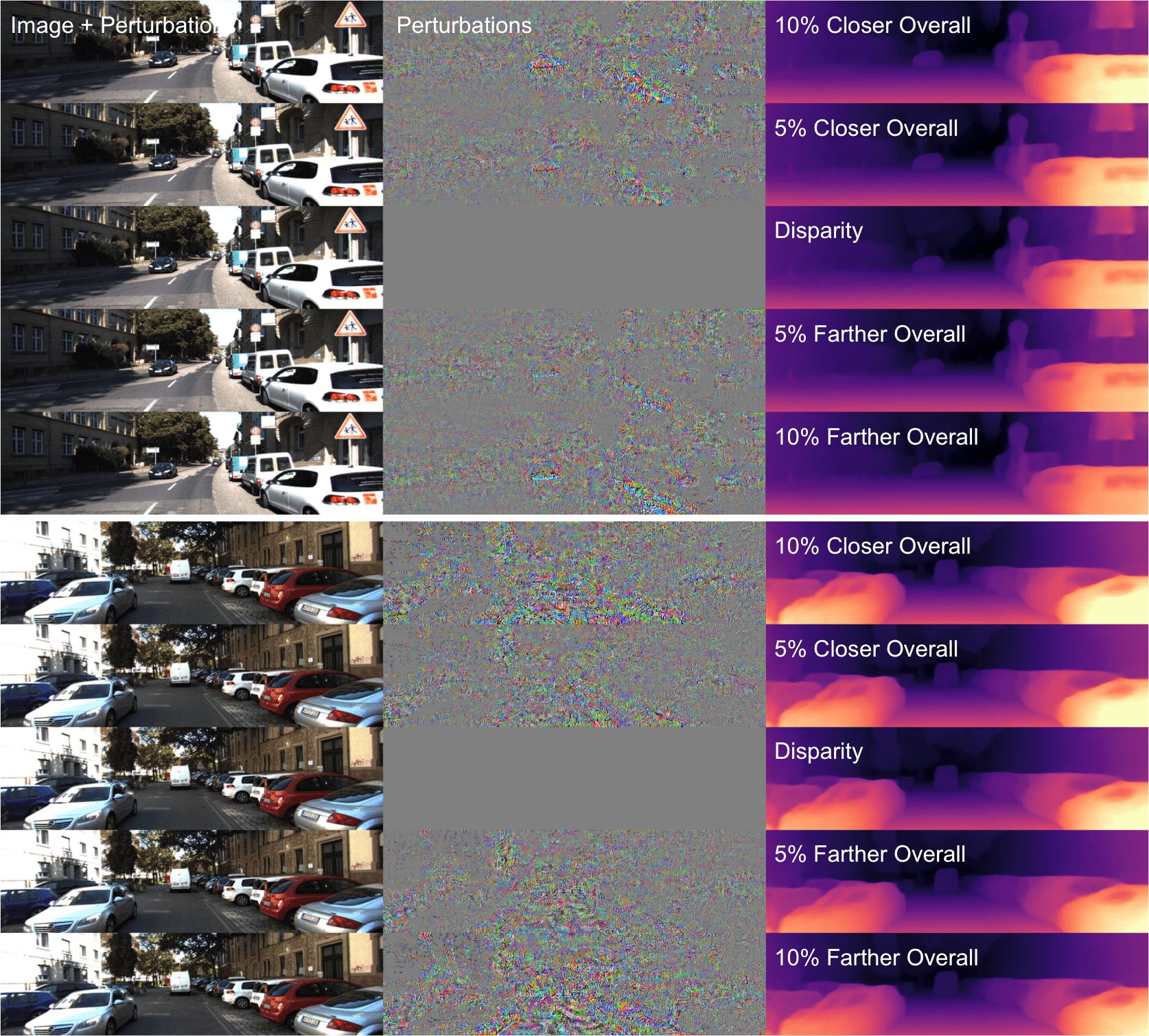}
    \caption{Visually imperceptible perturbations $v(x)$, with $\xi = 2 \times 10^{-2}$, can fool Monodepth2 to predicted scenes that are 5\% or 10\% closer and also 5\% or 10\% farther.}
    \label{fig:scale_overall}
\end{figure}

Here, we show qualitative results for the task of scaling the scene (Sec. 4.1, main text) by a factor of $1 + \alpha$ where $\alpha \in \{ -0.10, -0.05, +0.05, +0.10\}$. As seen in \figref{fig:scale_overall}, the perturbations are successful in fooling state-of-the-art monocular depth prediction method, Monodepth2 \cite{godard2019digging}, into predicting the scene 5\% or 10\% closer and also 5\% or 10\% farther. Additionally, the perturbations are concentrated in similar regions for scaling the scene 5\% or 10\% closer and for 5\% or 10\% farther as well. As noted in the main text, the amount of noise required for scaling the scene by $\pm 5\%$ are approximately the same, as is the amount for scaling the scene by $\pm 10\%$. This is visible in \figref{fig:scale_overall}.

\subsection{Symmetrically Flipping the Scene}
\begin{figure}[t]
    \centering
    \includegraphics[width=1.00\textwidth]{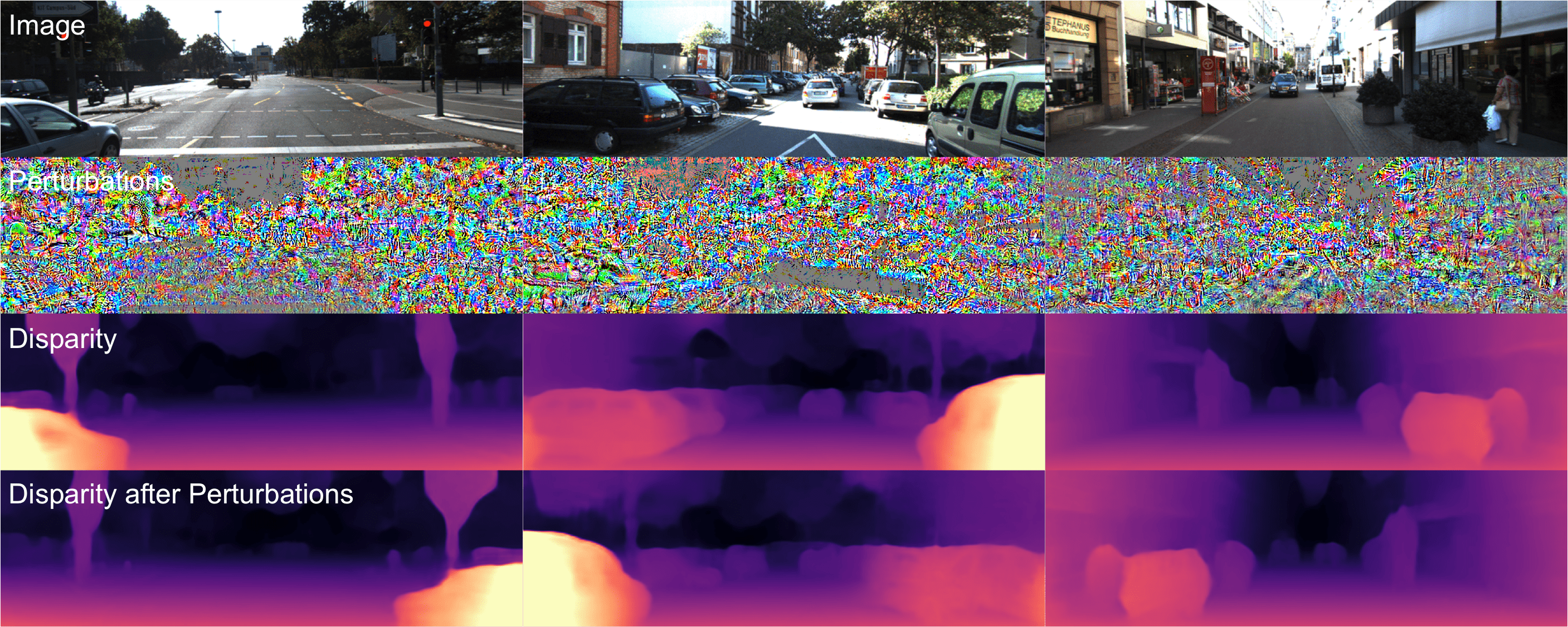}
    \caption{\textit{Additional examples of horizontal flip}. Adversarial perturbations with $\xi = 2 \times 10^{-2}$, can fool Monodepth2 to predicted scenes that horizontally flipped. }
    \label{fig:horizontal_flip}
\end{figure}

\begin{figure}[t]
    \centering
    \includegraphics[width=1.00\textwidth]{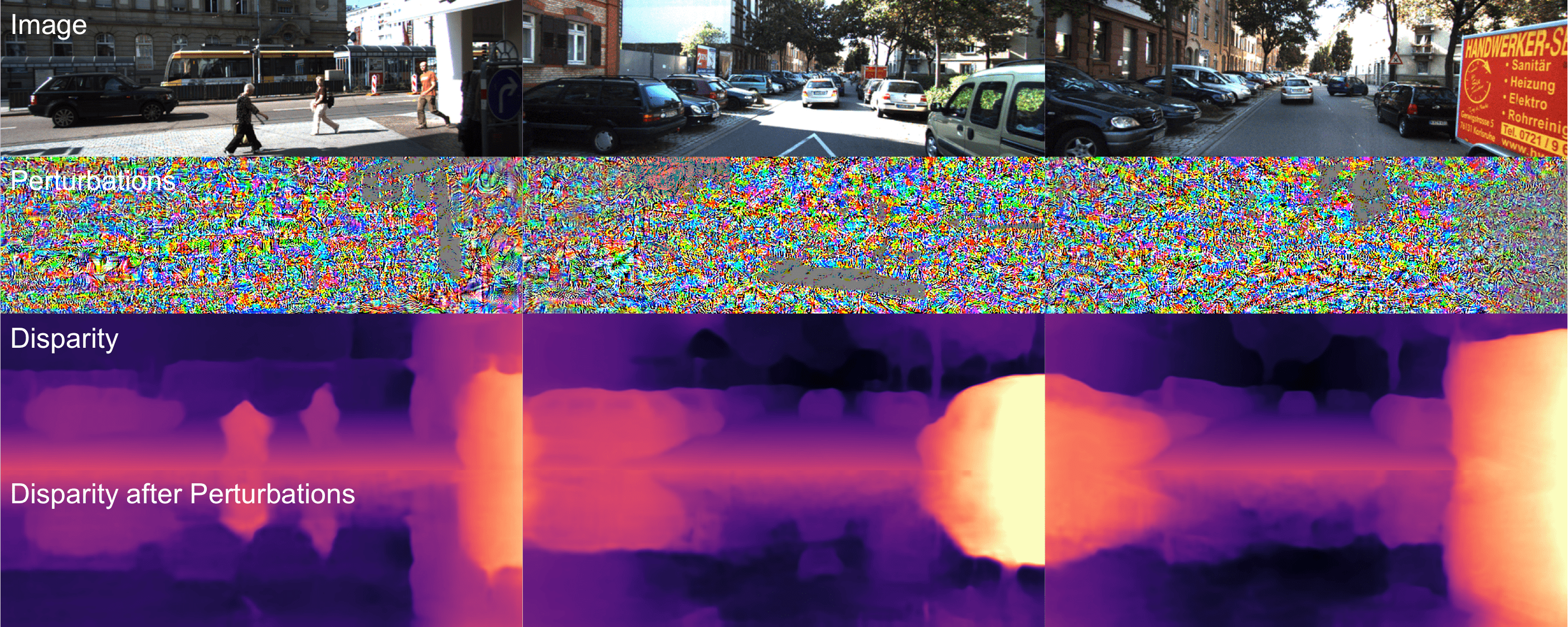}
    \caption{\textit{Additional examples of vertical flip}. Adversarial perturbations with $\xi = 2 \times 10^{-2}$, can fool Monodepth2 to predicted scenes that vertically flipped. Even in these successful examples, there are still artifacts (ripples, wavy-ness) in the output.}
    \label{fig:vertical_flip}
\end{figure}

In Sec. 4.2 and Fig. 3 in the main text, we demonstrated that adversarial perturbation can cause a monocular depth prediction network to predict a horizontally or vertically flipped scene. Here, we show additional qualitative results on the horizontal and vertical flipping tasks in \figref{fig:horizontal_flip} and \figref{fig:vertical_flip}. We note that perturbations can cause the network to predict a horizontally flip scene, they have trouble fooling the network to predict a vertically flipped scene. This is unlike our findings in the indoor scenario (\secref{sec:adversarial_attacks_for_indoor_scenes}) as seen in \figref{fig:indoor_figures} and \ref{fig:indoor_plots}. Fooling the network to vertically flip the scene is in fact \textit{easier} than fooling it to horizontally flip the scene. This confirms the biases (roads on bottom, sky on top) that the network learned from the outdoor dataset that are not present in the indoor dataset. 


\subsection{Category Conditioned Scaling}
In Sec. 5.1 and Fig. 5 of the main text, we showed category specific attacks to scale all objects belonging a given category to be a factor of $1 + \alpha$ closer or farther where $\alpha \in \{-0.10, -0.05, +0.05, +0.10\}$. Here, we provide performance, measured in ARE, of adversarial perturbations crafted for each category. We use the same convention for grouping different classes into categories as the Cityscapes dataset \cite{cordts2016cityscapes}, with the exception of the ``Human'' category, which includes the bicycles that the bikers are riding. 

\begin{figure}[h]
    \centering
    \includegraphics[width=0.24\textwidth]{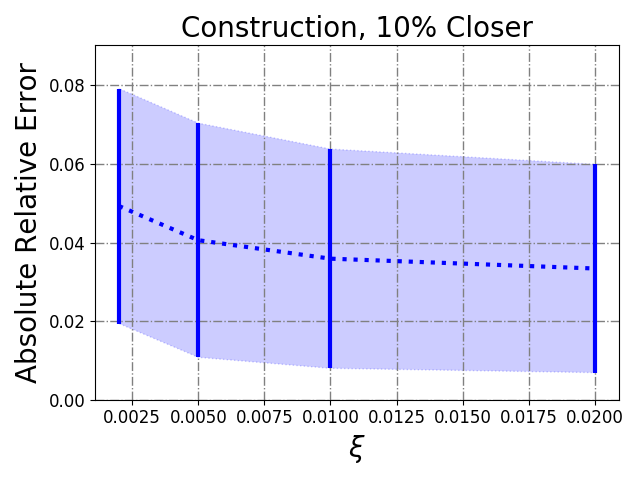}
    \includegraphics[width=0.24\textwidth]{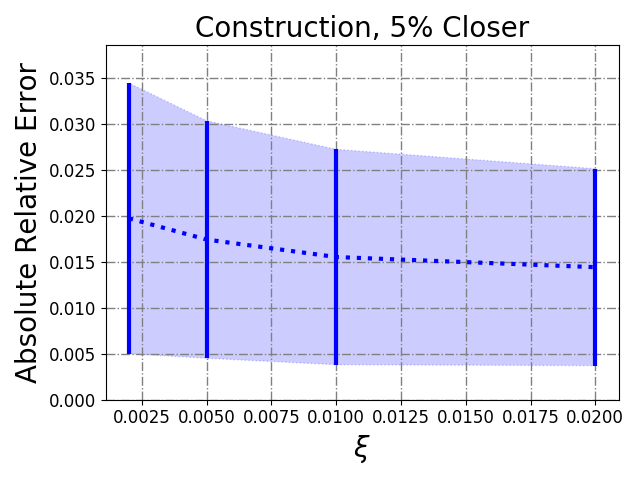}
    \includegraphics[width=0.24\textwidth]{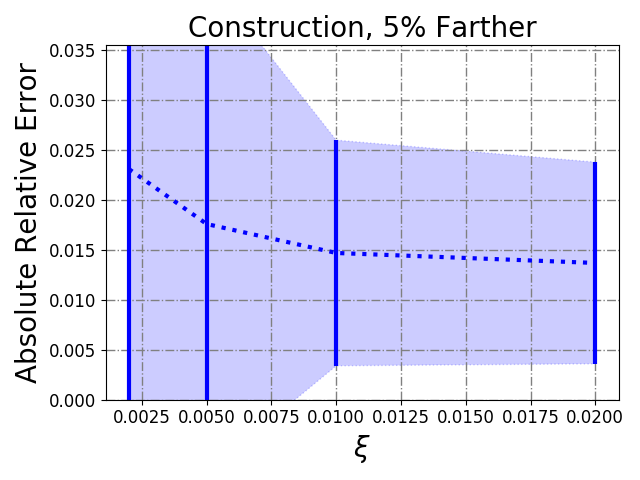}
    \includegraphics[width=0.24\textwidth]{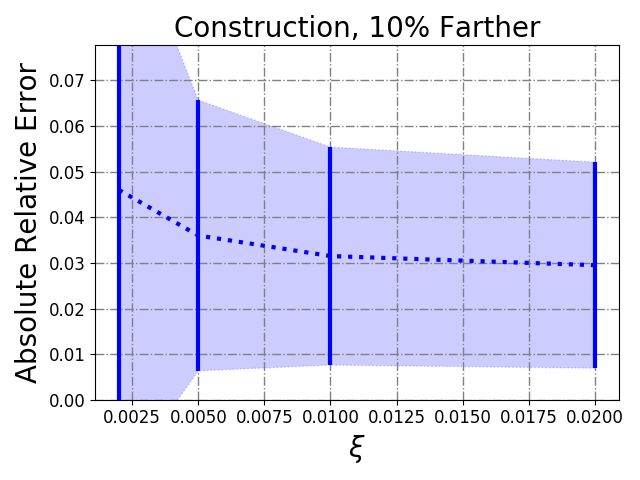}
    
    \includegraphics[width=0.24\textwidth]{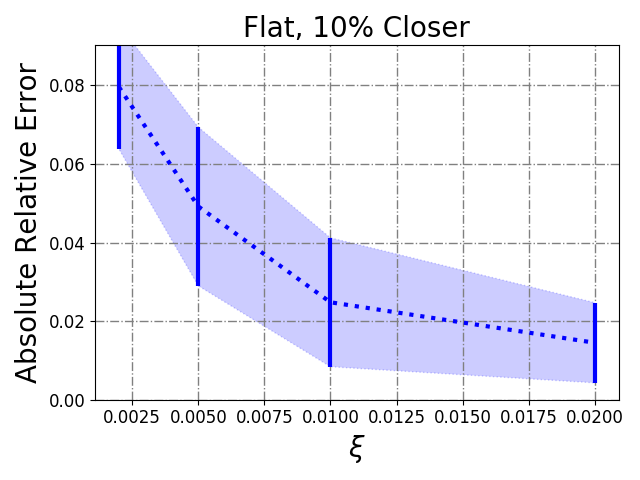}
    \includegraphics[width=0.24\textwidth]{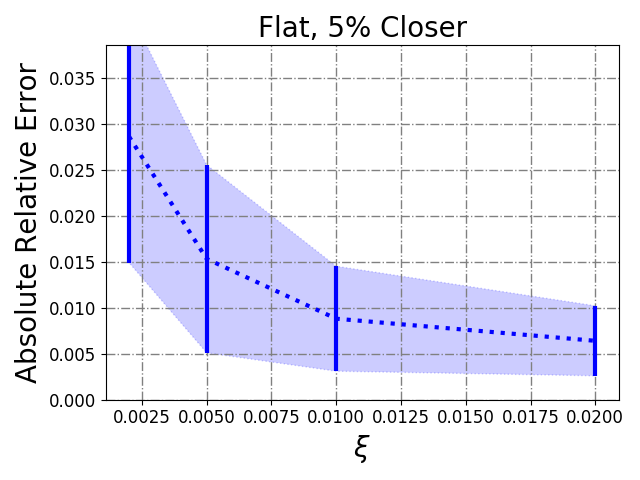}
    \includegraphics[width=0.24\textwidth]{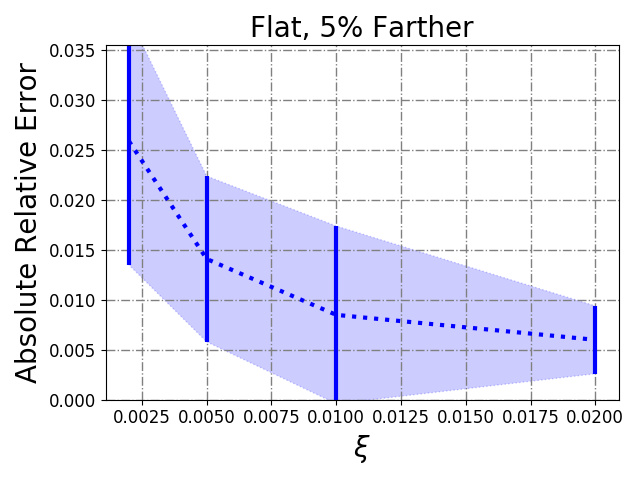}
    \includegraphics[width=0.24\textwidth]{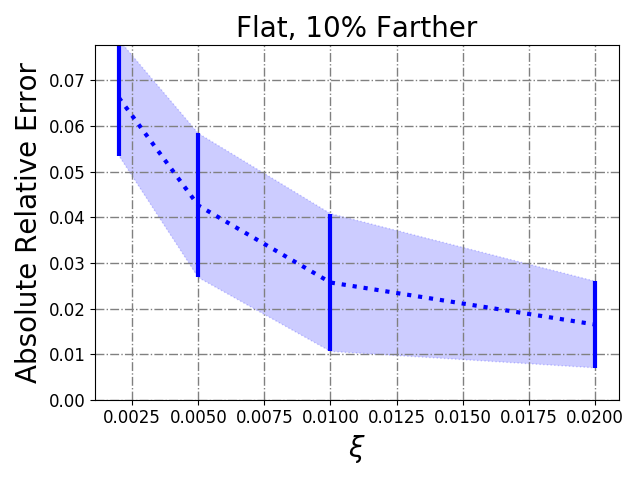}
    
    \includegraphics[width=0.24\textwidth]{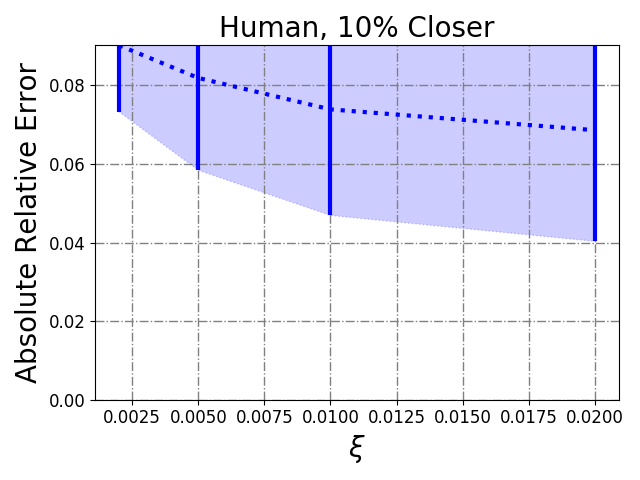}
    \includegraphics[width=0.24\textwidth]{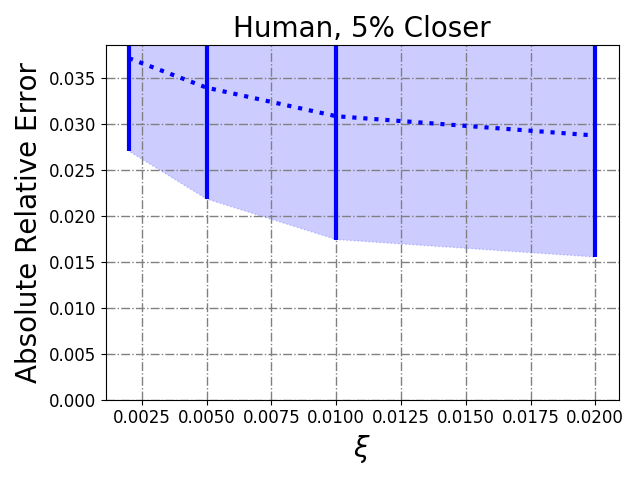}
    \includegraphics[width=0.24\textwidth]{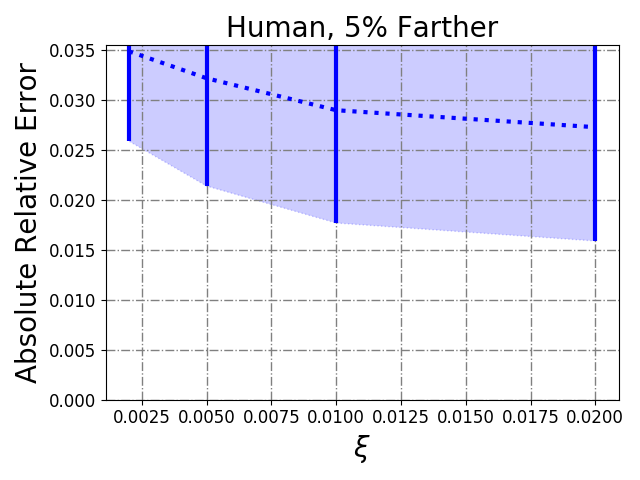}
    \includegraphics[width=0.24\textwidth]{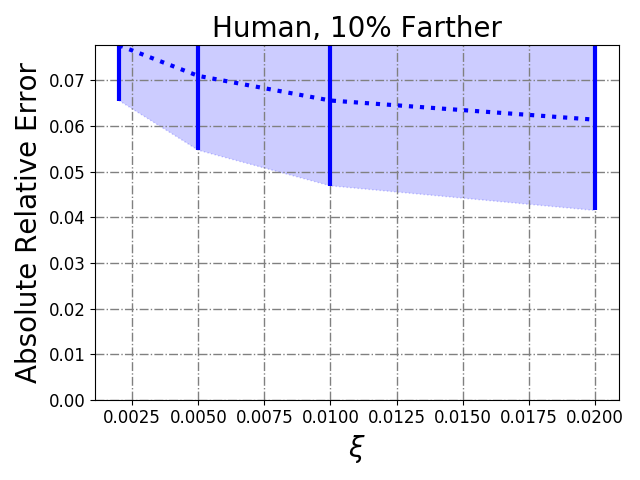}
    
    \includegraphics[width=0.24\textwidth]{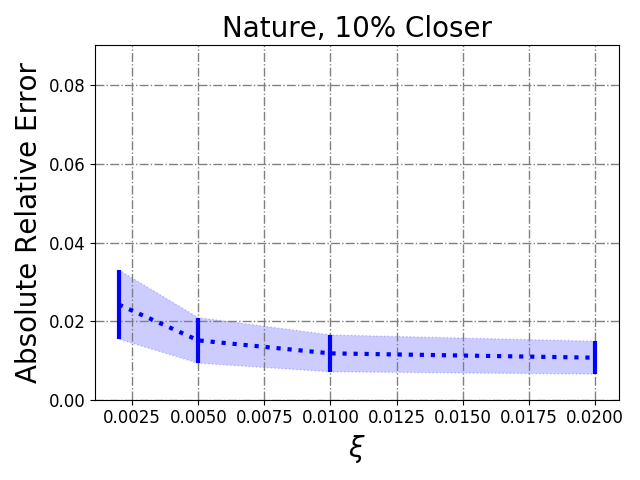}
    \includegraphics[width=0.24\textwidth]{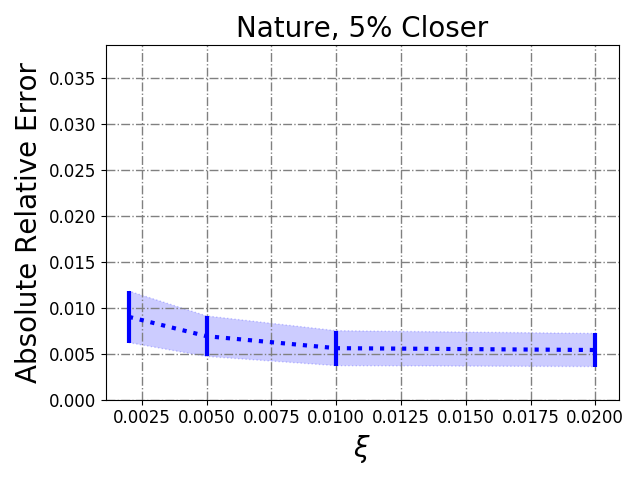}
    \includegraphics[width=0.24\textwidth]{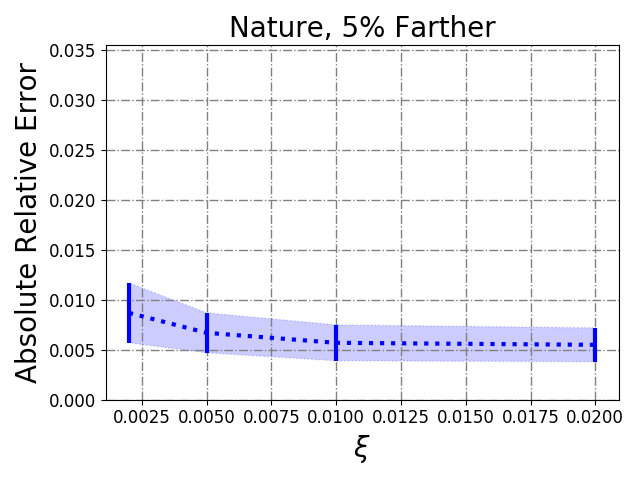}
    \includegraphics[width=0.24\textwidth]{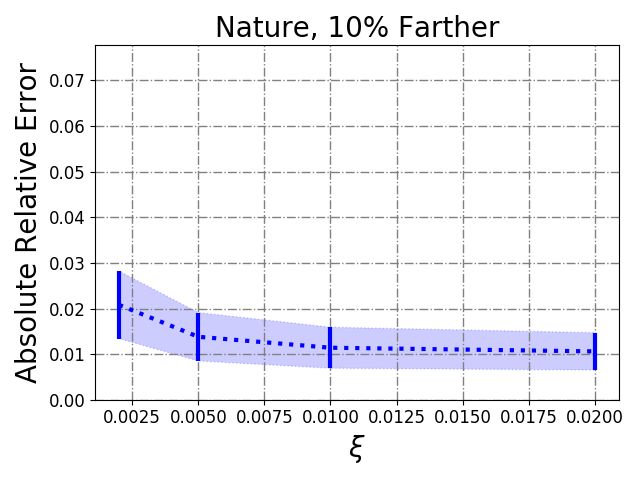}

    \includegraphics[width=0.24\textwidth]{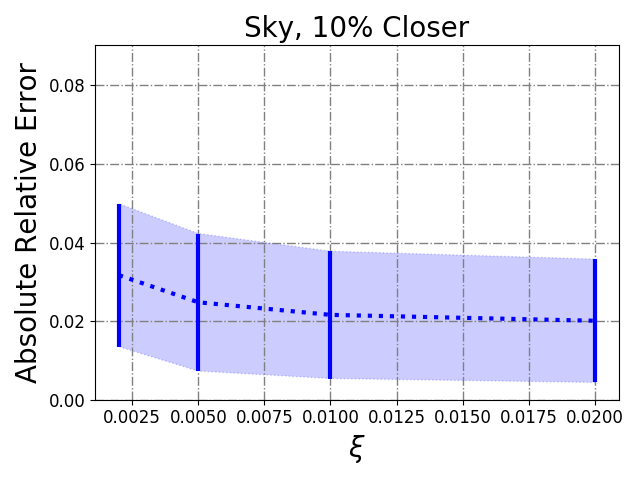}
    \includegraphics[width=0.24\textwidth]{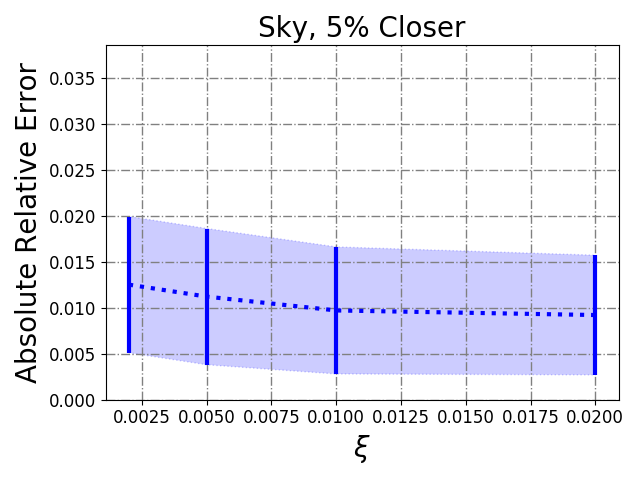}
    \includegraphics[width=0.24\textwidth]{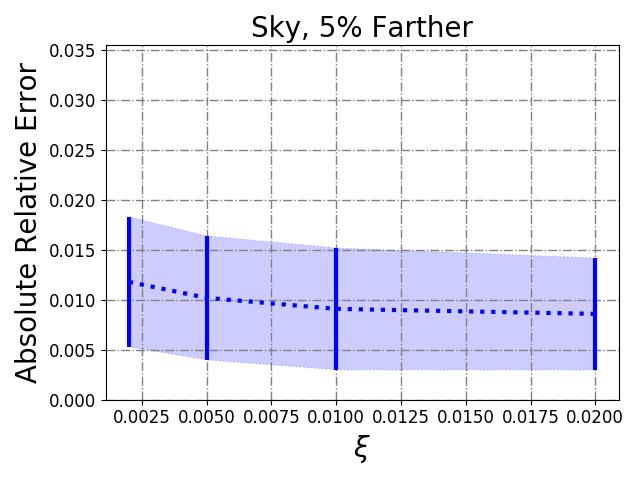}
    \includegraphics[width=0.24\textwidth]{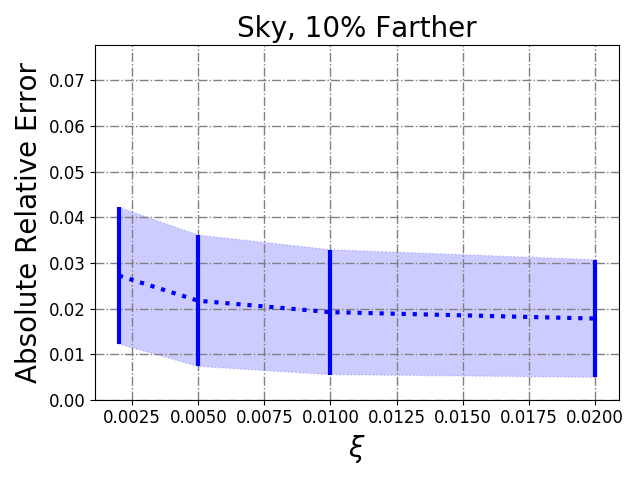}
    
    \includegraphics[width=0.24\textwidth]{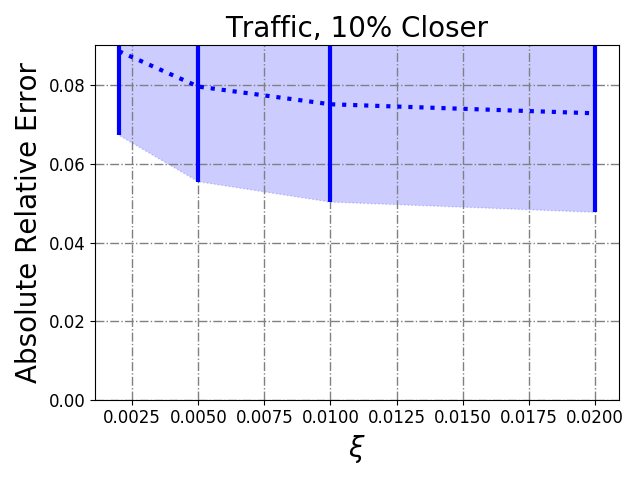}
    \includegraphics[width=0.24\textwidth]{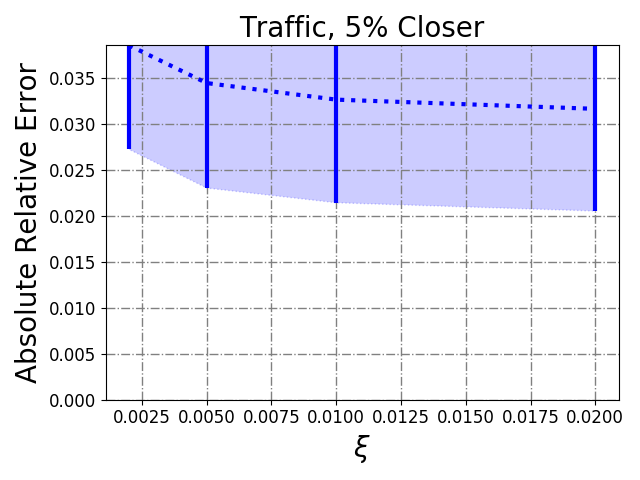}
    \includegraphics[width=0.24\textwidth]{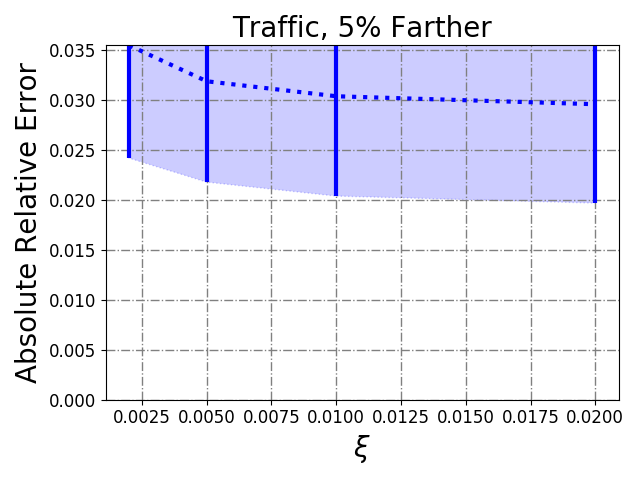}
    \includegraphics[width=0.24\textwidth]{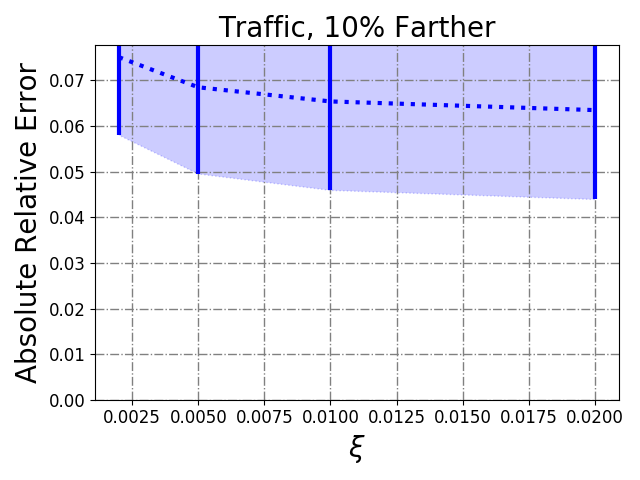}
    
    \includegraphics[width=0.24\textwidth]{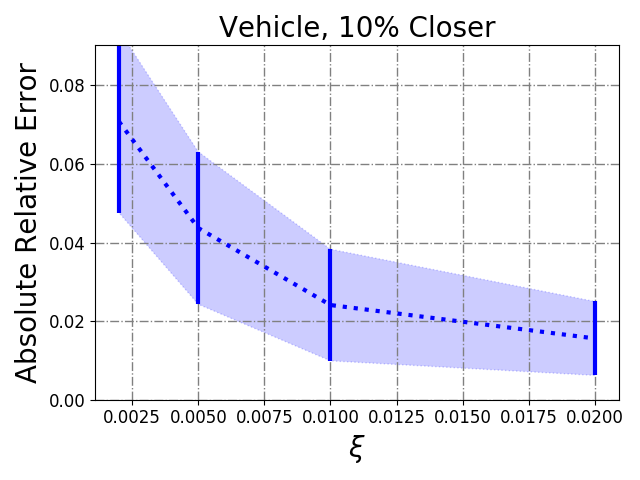}
    \includegraphics[width=0.24\textwidth]{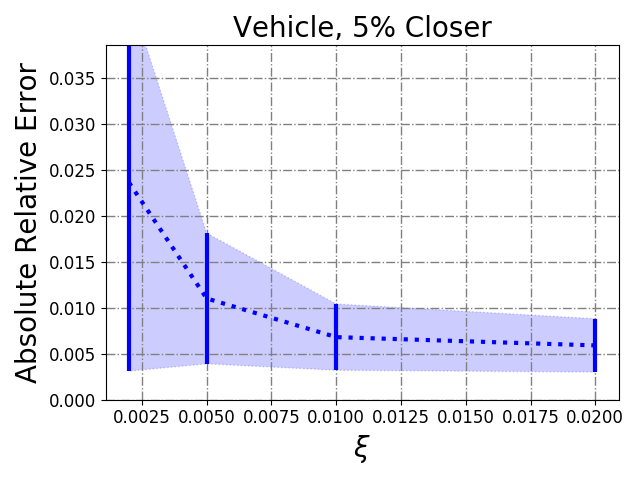}
    \includegraphics[width=0.24\textwidth]{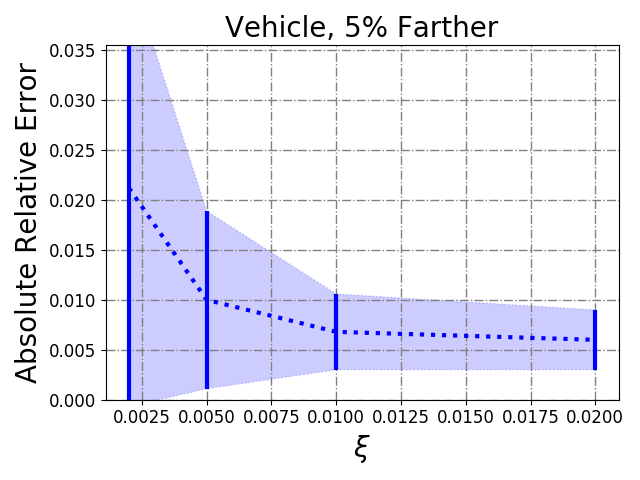}
    \includegraphics[width=0.24\textwidth]{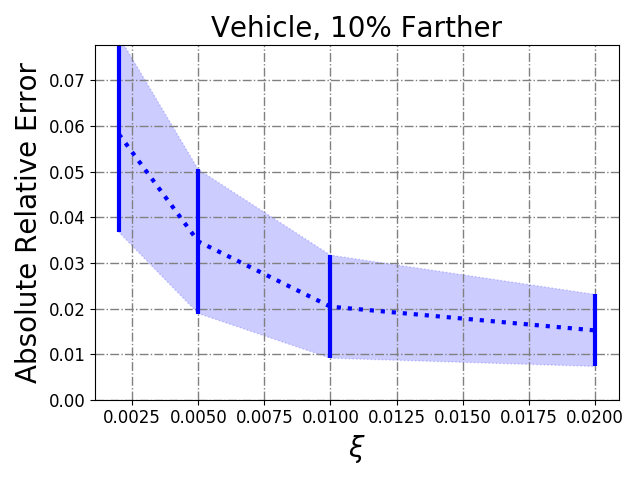}
    \caption{\textbf{ARE for scaling different categories closer and farther.} From top to bottom: ``Construction'', ``Flat'', ``Human'', ``Nature'', ``Sky'', ``Traffic'', ``Vehicle''. From left to right: 10\% closer, 5\% closer, 5\% farther, 10\% farther. Y-axis is kept the same for the same scale, for making the comparison across categories possible. It is easier to fool the network to predict vehicle and nature categories closer and farther than is to fool human and traffic categories. }
    \label{fig:plots_scale_category_all}
\end{figure}

\begin{figure}[t]
    \centering
    \includegraphics[width=0.95\textwidth]{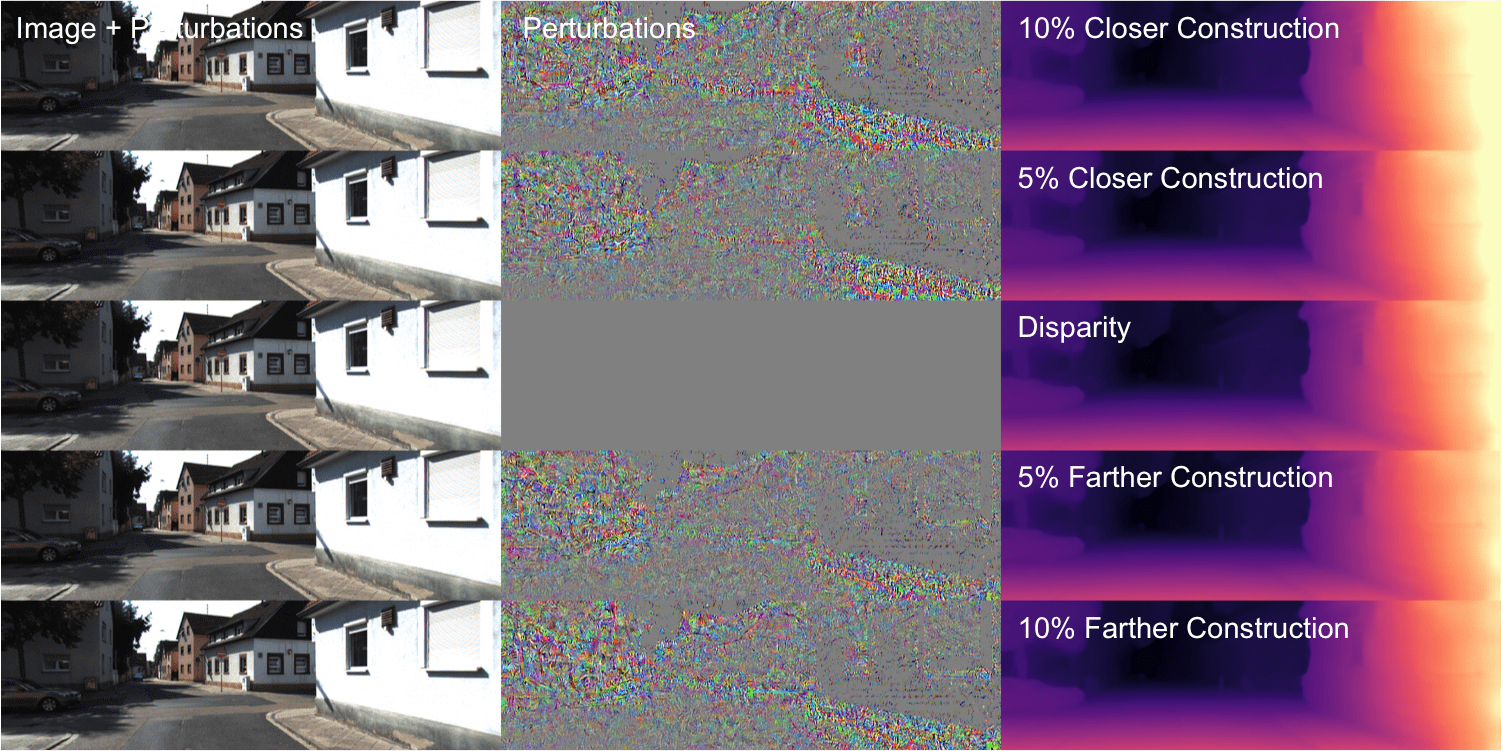}
    \caption{\textbf{Examples of targeted attacks on regions belonging to ``Construction'' category.}}
    \label{fig:scale_construction}
\end{figure}

\begin{figure}[t]
    \centering
    \includegraphics[width=0.95\textwidth]{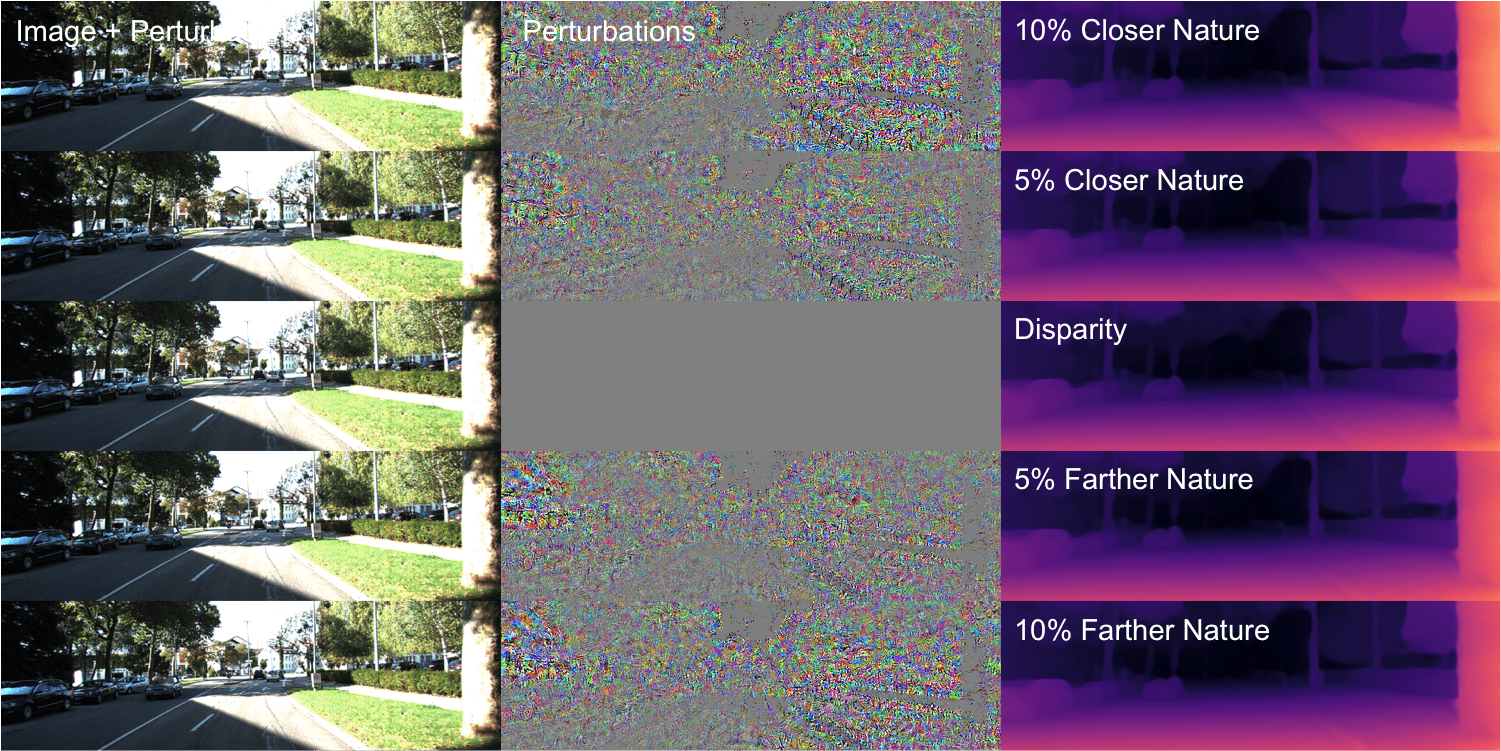}
    \caption{\textbf{Examples of targeted attacks on regions belonging to ``Nature'' category.}}
    \label{fig:scale_nature}
\end{figure}

\begin{figure}[t]
    \centering
    \includegraphics[width=0.95\textwidth]{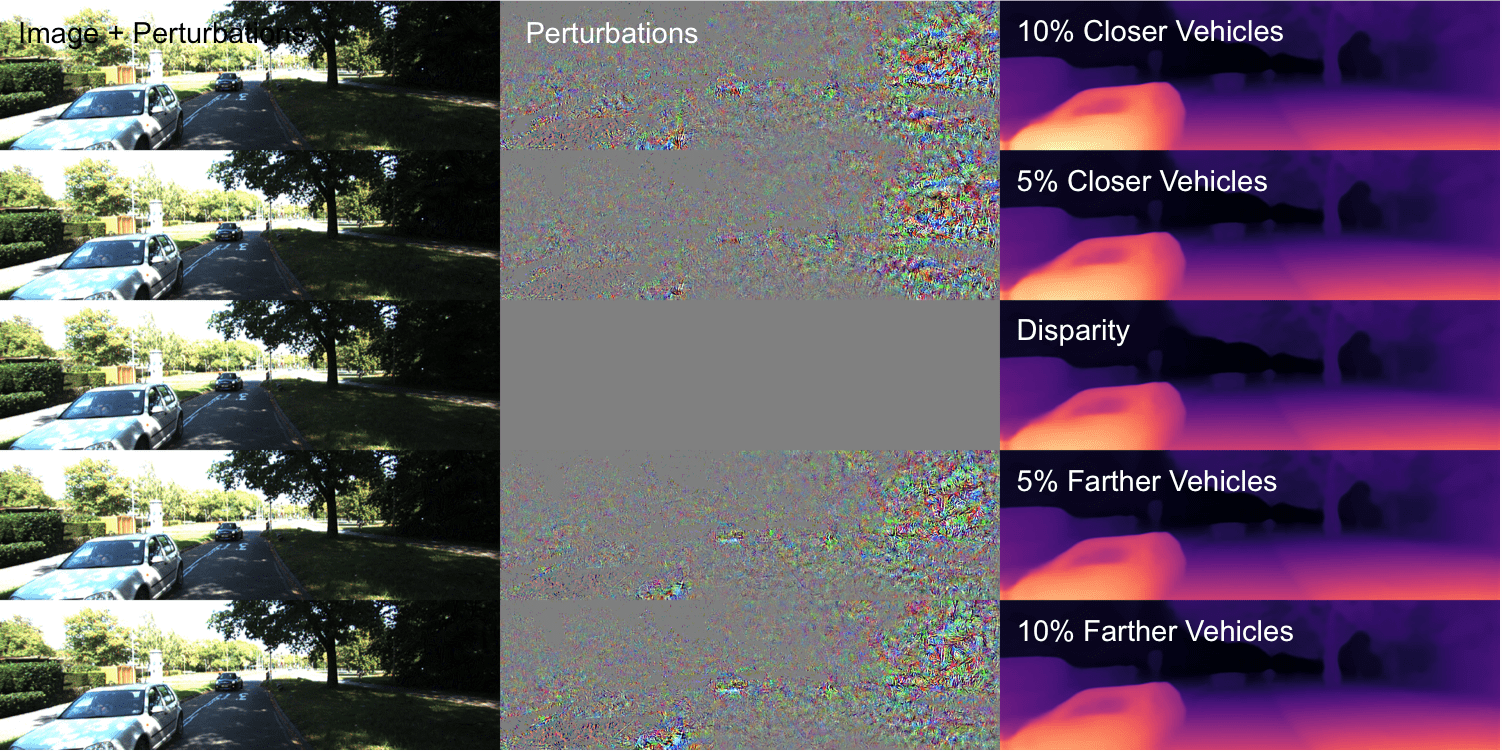}
    \caption{\textbf{Examples of targeted attacks on regions belonging to ``Vehicle'' category.}}
    \label{fig:scale_vehicle}
\end{figure}

\figref{fig:plots_scale_category_all} shows a comparative study between different categories. Not all categories are equally easy to be fooled by the perturbations, some are more robust to adversarial attacks than others. As seen in \figref{fig:plots_scale_category_all}, each category exhibits a different level of robustness to adversarial noise -- ``Human'' and ``Traffic'' categories are the hardest to fool, ``Construction'', ``Vehicle'' and ``Flat'' are more susceptible, and ``Sky'' and ``Nature'' are the easiest to attack. Plots are cropped at the maximum error across different categories to enable comparison of difficulty in fooling different categories. We note that attacking localized regions in the scene is considerably harder than attacking the entire scene. \figref{fig:plots_large_scales} shows that perturbations can attack the entire scene with small errors across various norms while \figref{fig:plots_scale_category_all} shows that, even with large norms, there are still errors ($\approx 2\%$ to $6\%$ ARE). We show visualizations for the ``Construction'', ``Nature'', and ``Vehicle'' categories in \figref{fig:scale_construction}, \ref{fig:scale_nature}, and \ref{fig:scale_vehicle} respectively.


\clearpage

\subsection{Instance Conditioned Targeted Attacks}
\begin{figure}[t]
    \centering
    \includegraphics[width=1.00\textwidth]{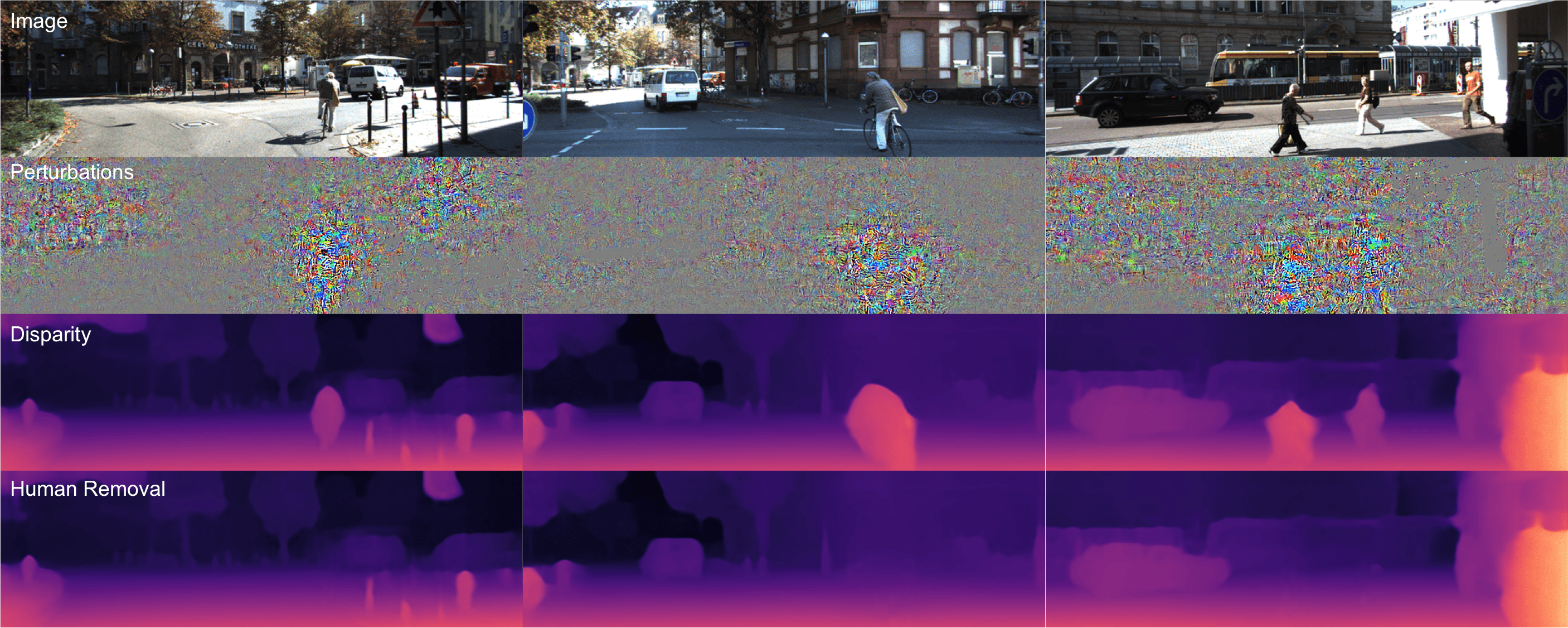}
    \caption{\textbf{Additional examples of human removal.} Targeted adversarial perturbations can remove humans from the predicted scene. Rightmost panel shows that we can target multiple humans and remove them from the scene without affecting the remaining pedestrian on the right.}
    \label{fig:human_removal}
\end{figure}

\begin{figure}[t]
    \centering
    \includegraphics[width=1.00\textwidth]{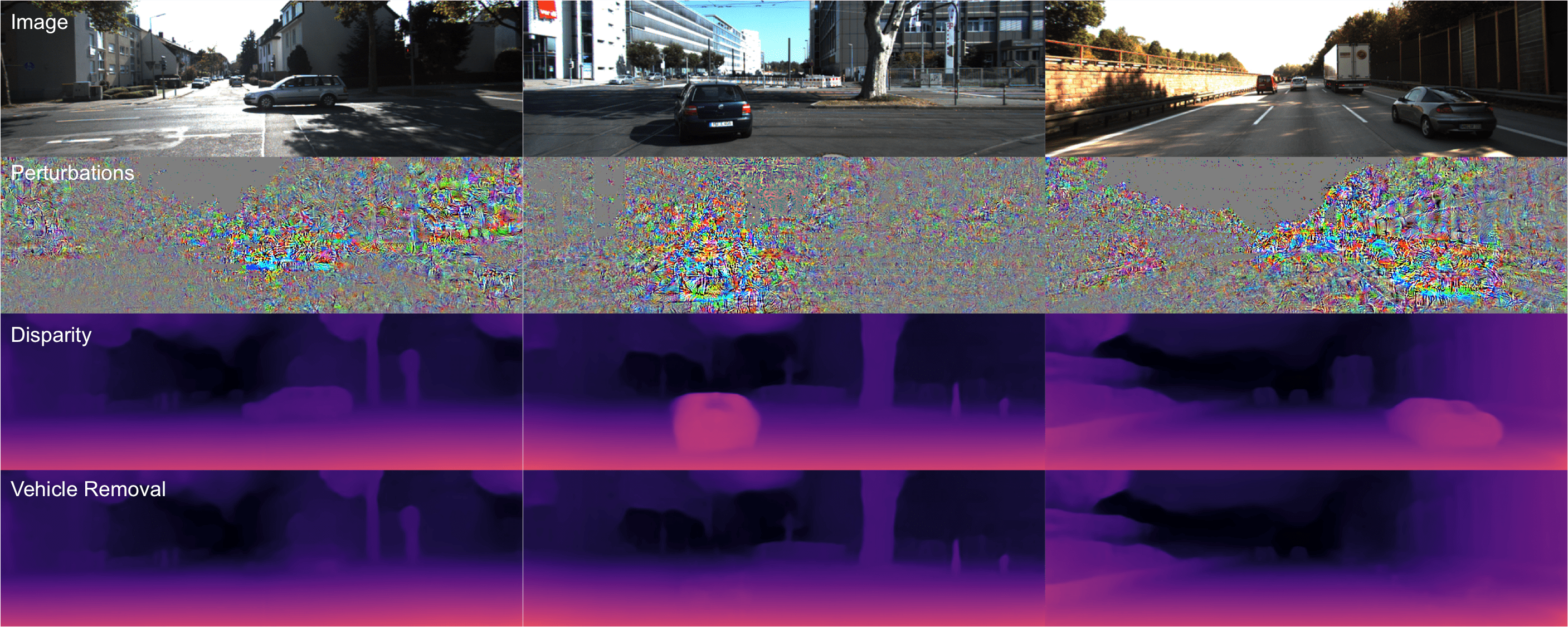}
    \caption{\textbf{Examples of vehicle removal.} Targeted adversarial perturbations can remove vehicles from the predicted scene. Rightmost panel shows that we can target a truck and a car on the right side and remove them. Note that the cars in the middle of the road still remain.}
    \label{fig:vehicle_removal}
\end{figure}

\begin{figure}[h]
    \centering
    \includegraphics[width=1.00\textwidth]{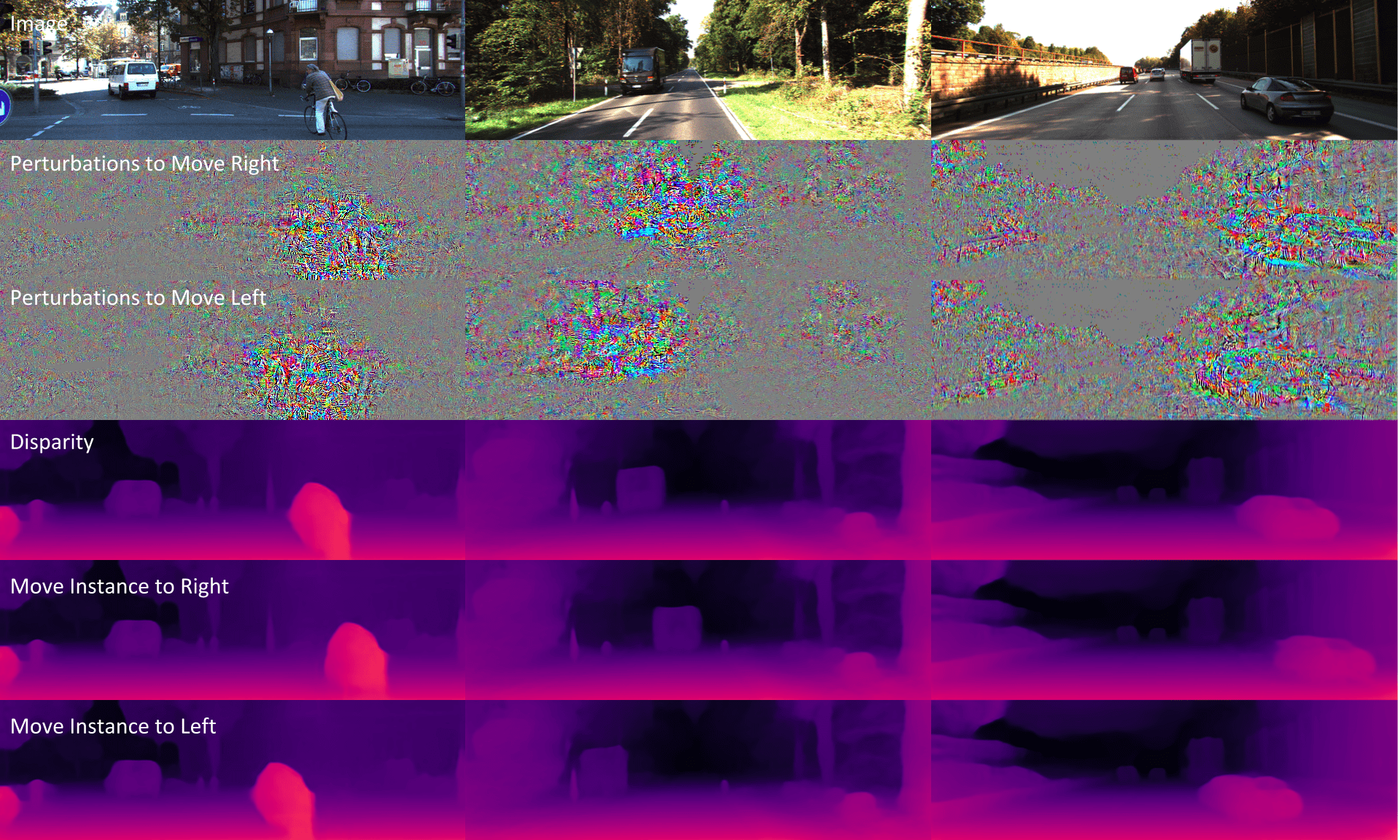}
    \caption{\textbf{Move instance horizontally.} Selected instance is moved by $\approx 8\%$ in left and right directions while rest of the scene is preserved. Noise and disparity for both directions are given. We note that the noise is concentrated around the targeted instance and the region to which the instance is moved.}
    \label{fig:horizontal_move}
\end{figure}
\begin{figure}[h]
    \centering
    \includegraphics[width=1.00\textwidth]{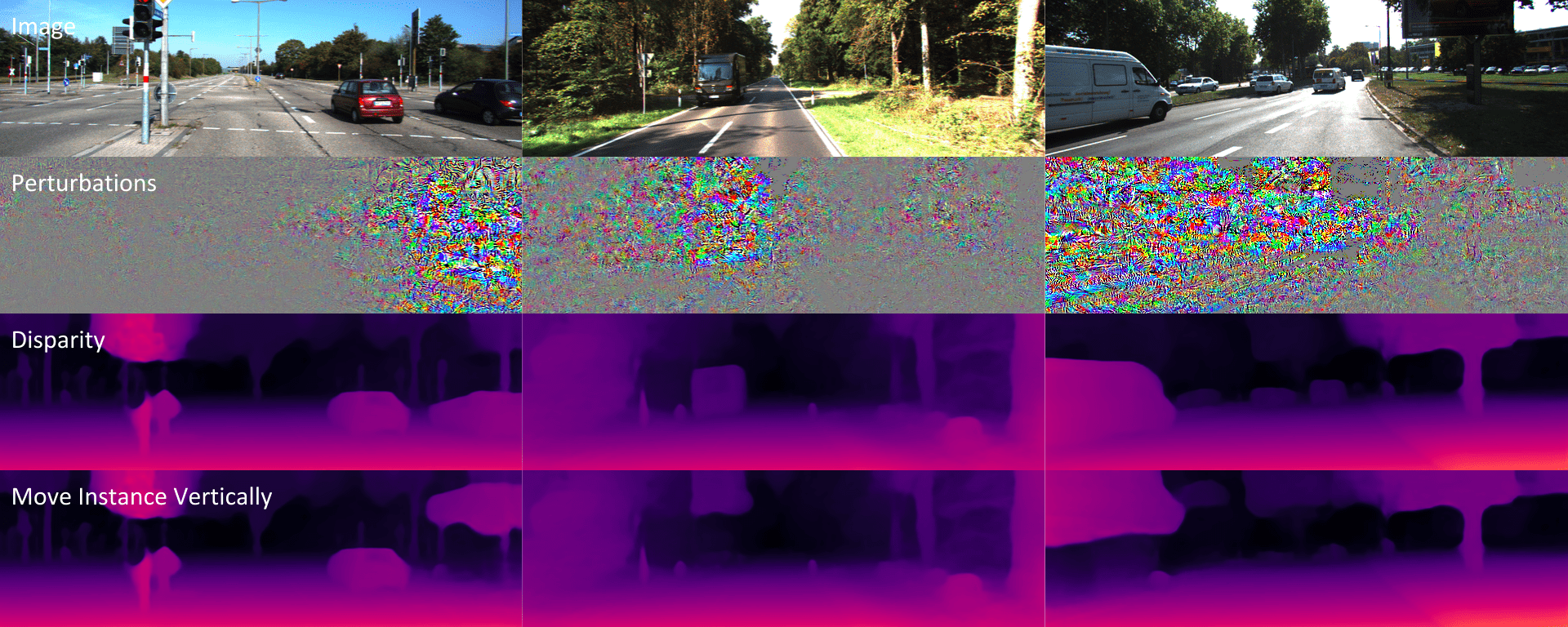}
    \caption{\textbf{Flying vehicles.} Selected vehicle is moved by $\approx 42\%$ in the vertical direction while rest of the scene is preserved. We note that the noise is generally concentrated around the targeted instance and the region to which the instance is moved.}
    \label{fig:vertical_move}
\end{figure}
In Sec. 5.2 and Fig. 6 in the main text, we show that, when given instance segmentation, adversarial perturbations can target specific instances and remove them from the scene and thus causing unforseen consequences. \figref{fig:human_removal} shows additional examples of removing humans from the scene and \figref{fig:vehicle_removal} demonstrates that it is possible to remove vehicles from the scene as well. In the rightmost panel of \figref{fig:human_removal}, we show that it is possible to remove \textit{some} pedestrians from the scene without affecting others. Similarly, in the rightmost panel of \figref{fig:vehicle_removal}, we removed a truck and a car on the right side and left the cars in the center untouched -- leaving this as \textit{still} a plausible highway driving scenario.

In Sec. 5.4 and Fig. 7 in the main text, we show that perturbations can move an instance to another location in the scene (requires removing the instance from its original location and creating it in the new location). In this section, we give more visuals for the perturbations used for moving an instance (e.g. vehicle, pedestrian) horizontally or vertically in the image space while keeping the rest of the scene unchanged.

\figref{fig:horizontal_move} shows that perturbations can fool a network to move the target instance by $\approx 8\%$ across the image in the left and right directions. Furthermore, \figref{fig:vertical_move} shows that perturbations can move select instances by $\approx 42\%$ in the upward direction, creating the illusion that there are ``flying vehicles'' in the scene. We note that in both cases, the perturbations are concentrated on the instance and the region to which the instance is moved. For example, when moving a vehicle right or left, the corresponding perturbations also move right or left.
\end{appendices}

\end{document}